\documentclass[sigconf,nonacm]{acmart}

\usepackage{multirow}
\usepackage{algorithm}
\usepackage{algpseudocode} 

\usepackage{longtable}
\usepackage{multirow}
\usepackage{pdflscape}  
\usepackage{booktabs}
\usepackage[T1]{fontenc}
\usepackage{tabularx}
\usepackage{subcaption}
\usepackage{booktabs}
\usepackage{amsmath}
\usepackage{graphicx}
\usepackage{standalone}
\usepackage{tikz}
\usepackage{pgfplots}
\pgfplotsset{compat=1.18}
\usetikzlibrary{patterns, decorations.pathreplacing, arrows.meta, 
  calc, positioning, backgrounds, fit}

\definecolor{gtcolor}{RGB}{0,0,0}
\definecolor{saitscolor}{RGB}{31,119,180}
\definecolor{lerpcolor}{RGB}{255,127,14}
\definecolor{maskstat}{RGB}{200,220,255}
\definecolor{masktrans}{RGB}{255,200,200}
\definecolor{stationaryzone}{RGB}{235,245,235}
\definecolor{transientzone}{RGB}{255,240,240}
\definecolor{hypoline}{RGB}{180,0,0}

\renewcommand\footnotetextcopyrightpermission[1]{} 
\begin{document}

\title{The Stationarity Bias: Stratified Stress-Testing for Time-Series Imputation in Regulated Dynamical Systems}
\author{Amirreza Dolatpour Fathkouhi}
\email{aww9gh@virginia.adu}
\affiliation{%
  \institution{Department of Computer Science Center for Diabetes Technology University of Virginia}
  \city{Charlottesville}
  \state{Virginia}
  \country{USA}
}

\author{Alireza Namazi}
\email{mez4em@virginia.edu}
\affiliation{%
  \institution{Department of Computer Science Center for Diabetes Technology University of Virginia}
  \city{Charlottesville}
  \state{Virginia}
  \country{USA}
}

\author{Heman Shakeri}
\email{hs9hd@virginia.edu}
\affiliation{%
  \institution{School of Data Science \\ Center for Diabetes Technology University of Virginia}
  \city{Charlottesville}
  \state{Virginia}
  \country{USA}
}
\renewcommand{\shortauthors}{Dolatpour Fathkouhi et al.}

\begin{abstract}
Time-series imputation benchmarks employ uniform random masking and 
shape-agnostic metrics (MSE, RMSE), implicitly weighting evaluation by 
regime prevalence. In systems with a dominant attractor---homeostatic 
physiology, nominal industrial operation, stable network 
traffic---this creates a systematic \emph{Stationarity Bias}: simple 
methods appear superior because the benchmark predominantly samples 
the easy, low-entropy regime where they trivially succeed.
We formalize this bias and propose a \emph{Stratified Stress-Test} 
that partitions evaluation into Stationary and Transient regimes. 
Using Continuous Glucose Monitoring (CGM) as a testbed---chosen for 
its rigorous ground-truth forcing functions (meals, insulin) that 
enable precise regime identification---we establish three findings 
with broad implications:
(i)~Stationary Efficiency: Linear interpolation achieves 
state-of-the-art reconstruction during stable intervals, confirming 
that complex architectures are computationally wasteful in 
low-entropy regimes.
(ii)~Transient Fidelity: During critical transients 
(post-prandial peaks, hypoglycemic events), linear methods exhibit 
drastically degraded morphological fidelity (DTW), disproportionate 
to their RMSE---a phenomenon we term the \emph{RMSE Mirage}, where 
low pointwise error masks the destruction of signal shape.
(iii)~Regime-Conditional Model Selection: Deep learning 
models preserve both pointwise accuracy and morphological integrity 
during transients, making them essential for safety-critical 
downstream tasks. We further derive empirical missingness distributions from clinical trials and impose them on complete training data, preventing models from exploiting unrealistically clean observations and encouraging robustness under real-world missingness. This framework generalizes to any regulated system where 
routine stationarity dominates critical transients. The source code and datasets for this project are available at \url{https://github.com/Shakeri-Lab/Glucose-Imputation}.
\end{abstract}

\maketitle

\section{Introduction}
Many real-world dynamical systems are predominantly stationary: a server's CPU load is flat until a DDoS attack~\cite{barford2002signal}; a turbine vibrates steadily until a bearing fault~\cite{randall2011rolling}; a patient's blood glucose drifts within a narrow band until a meal disrupts homeostasis~\cite{cobelli2011artificial}. When time-series from such systems contain missing data, imputation benchmarks that employ uniform random masking~\cite{khayati2020mind, yoon2018gain} will, by construction, sample the dominant stationary regime far more often than the rare but critical transients. We argue that this creates a systematic \emph{Stationarity Bias} that inflates the apparent performance of simple methods and obscures the superiority of learned models precisely where it matters most.

We ground our analysis in Continuous Glucose Monitoring (CGM), where data loss is endemic due to battery depletion~\cite{rehman2024impact} and connectivity failures~\cite{cichosz2025assessing}. CGM offers an ideal testbed: meals and insulin doses provide precisely timed, ground-truth forcing functions that enable unambiguous regime identification. However, the bias we expose---and the stratified evaluation framework we propose---applies to any regulated system where routine stability dominates critical transients. Figure~\ref{fig:concept} illustrates both phenomena schematically.

The effectiveness of imputation strategies is heavily dependent on the underlying signal dynamics. Glucose regulation is fundamentally homeostatic, characterized by prolonged periods of stability in the absence of external perturbations (e.g., meals or exercise). Consequently, many data gaps occur during these low-volatility regimes, where simple methods like linear interpolation are often sufficient to approximate the signal with high accuracy.

However, simple methods are mechanistically limited; they inherently fail to capture the non-linear dynamics and complex morphology that characterize periods of high volatility. In these scenarios, advanced architectures—specifically Transformer-based \cite{du2023saits} and generative models \cite{tashiro2021csdi}—offer a theoretical advantage, as they are designed to learn and reconstruct the intricate temporal dependencies that linear models cannot resolve.

Despite these advantages, prior time-series imputation benchmarks have relied heavily on standard statistical metrics such as Mean Squared Error (MSE) and Mean Absolute Error (MAE) \cite{du2024tsi}. While effective for quantifying pointwise accuracy, these metrics often favor models that regress to the mean, thereby encouraging the "over-smoothing" of vital signal variations. This represents a critical oversight in physiological modeling, where preserving morphological fidelity is paramount. Downstream applications, such as closed-loop control algorithms \cite{man2014uva} and automated insulin delivery systems, depend not merely on average values, but on the precise trajectory and dynamics of the signal to make safety-critical decisions.

This tension is concrete in safety-critical control. For instance, the Extremum Seeking Control (ESC) framework for Zone-MPC~\cite{cao2017extremum} uses glucose-derived risk gradients to optimize controller parameters. Linear interpolation systematically dampens post-prandial peaks and hypoglycemic nadirs, distorting the risk landscape and biasing the optimizer toward suboptimal parameters. In regulated dynamical systems, \emph{safety is defined by the extremes, not the averages}. As illustrated in Figure~\ref{fig:concept}b, linear interpolation draws a chord through the missing interval, achieving low RMSE while destroying the peak amplitude and derivative that downstream controllers depend on.




Consequently, a critical gap remains in defining the boundaries of utility: specifically, distinguishing when simple linear models are sufficient from when the superior shape-preservation capabilities of deep learning models are necessary. To address these challenges, this study makes the following contributions:

\begin{enumerate}
    \item \textbf{The Stationarity Bias:} We provide the first formal 
    characterization of how uniform random masking on signals with a 
    dominant stationary regime systematically inflates the apparent 
    performance of linear imputation methods.
   
    \item \textbf{Regime-Conditional Model Selection:} Through stratified 
    evaluation across three regimes---Stationary, Post-Prandial Transient, 
    and Hypoglycemic Transient---we establish rigorous boundaries defining 
    \emph{when} deep learning is necessary versus when linear baselines 
    suffice.
    
    \item \textbf{The RMSE Mirage:} We demonstrate that RMSE rewards ``corner-cutting'' through transients, masking substantial morphological divergence. We introduce morphological evaluation via Dynamic Time Warping (DTW) to quantify the divergence between imputed and ground-truth trajectories, complemented by a calibration analysis that investigates distributional shifts between ground truth and imputed outputs---phenomena inherently invisible to RMSE's pointwise formulation.
    
    \item \textbf{Ecologically Valid Missingness:} We derive empirical 
    missingness distributions from real clinical trials (DCLP3, DCLP5) and 
    apply them to complete datasets, ensuring that training and evaluation 
    reflect actual dropout patterns rather than synthetic random masking.
\end{enumerate}

\begin{figure*}[t]
\centering
\includestandalone[width = 0.9\textwidth]{Images/concept}
\caption{Conceptual illustration. \textbf{(a)}~The Stationarity Bias: random masking disproportionately samples the dominant stationary regime, inflating baseline performance. \textbf{(b)}~The RMSE Mirage: linear interpolation ``cuts the corner'' of a transient peak, achieving comparable RMSE to a deep model while destroying signal morphology (DTW).}
\label{fig:concept}
\end{figure*}

\begin{figure}[htpb]
  \centering
  \begin{subfigure}{0.5\textwidth}
    \centering
    \includegraphics[width=\textwidth]{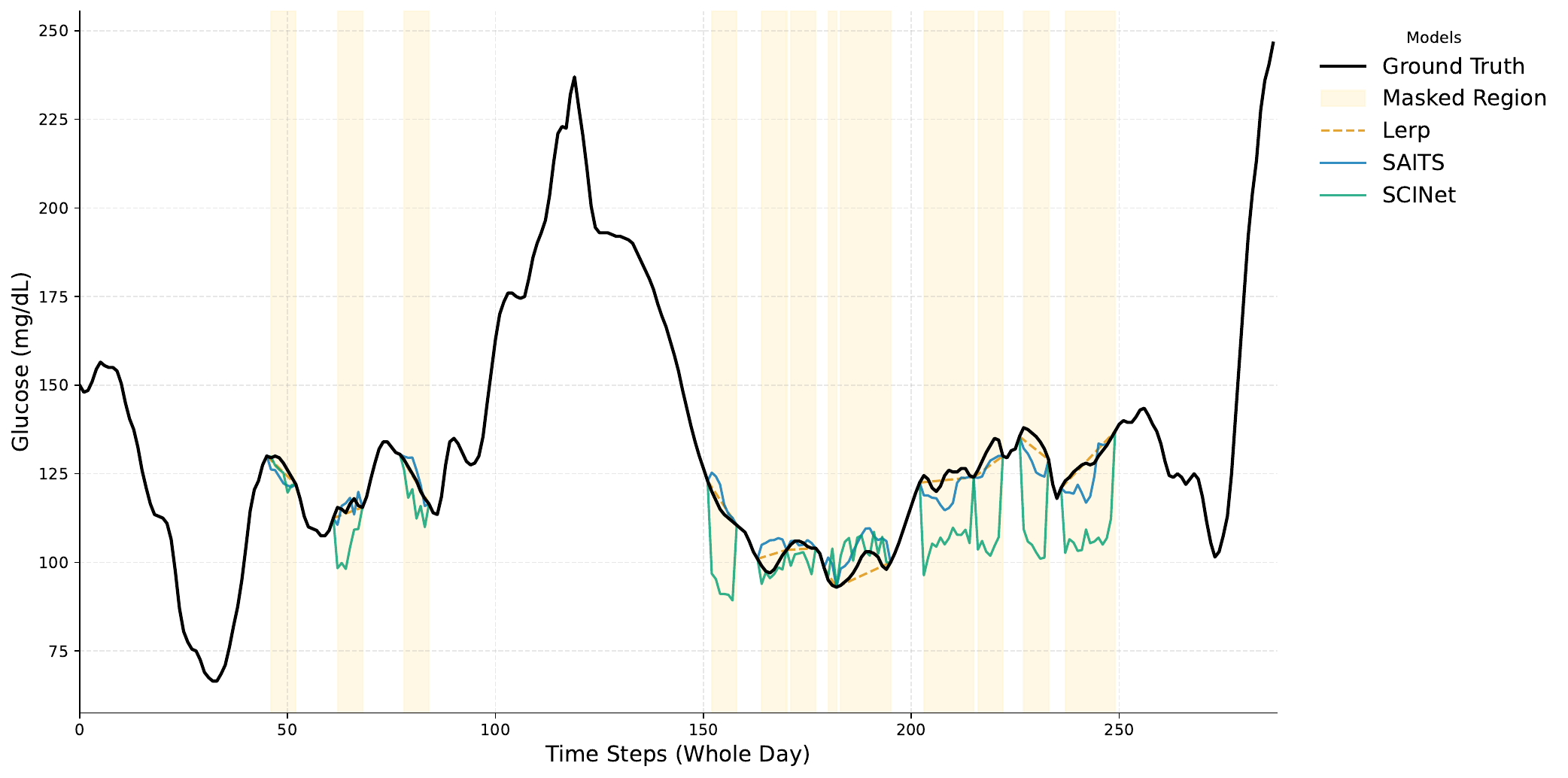}
    \caption{} 
    \label{fig:scenario_a_intro}
  \end{subfigure}
  \hfill 
  \begin{subfigure}{0.5\textwidth}
    \centering
    \includegraphics[width=\textwidth]{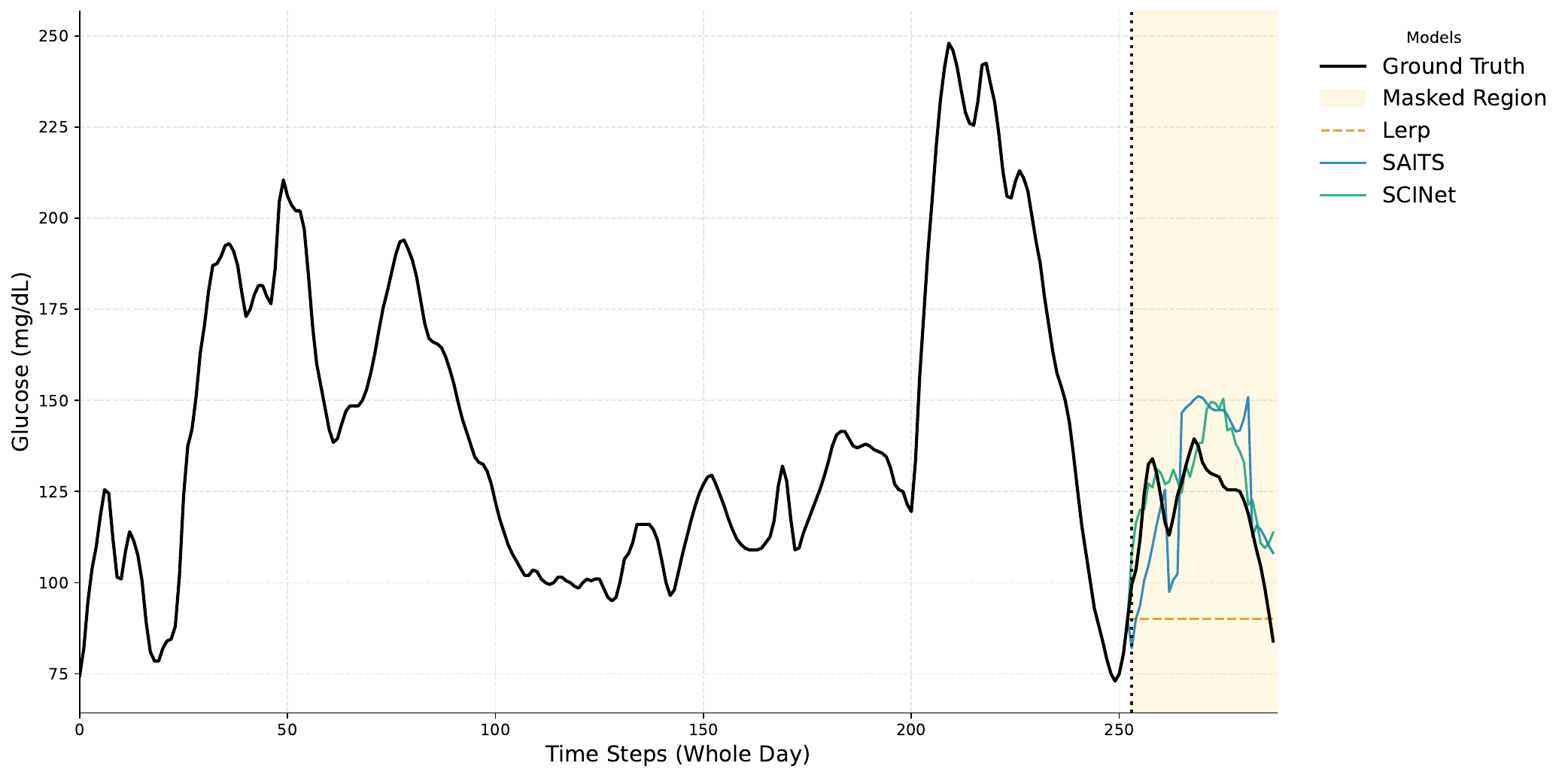}
    \caption{} 
    \label{fig:scenario_b_intro}
  \end{subfigure}
  \hfill
  \begin{subfigure}{0.5\textwidth}
    \centering
    \includegraphics[width=\textwidth]{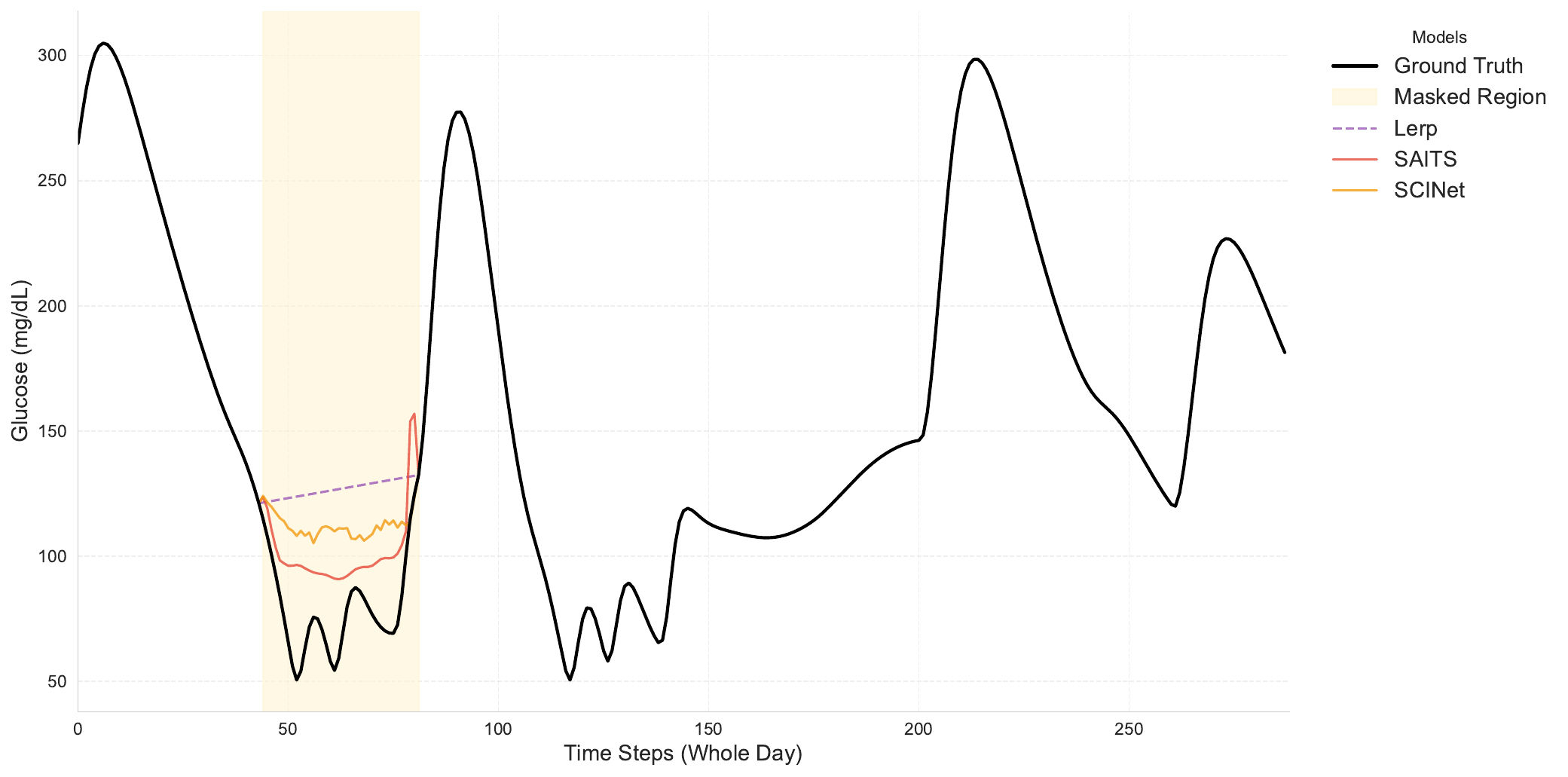}
    \caption{} 
    \label{fig:scenario_c_intro}
  \end{subfigure}

\caption{
    Imputation performance during \textbf{(a)} homeostatic periods, \textbf{(b)} post-prandial excursions, and \textbf{(c)} hypoglycemia during temporal controller resets. In \textbf{(a)}, linear interpolation is superior due to signal stability. In contrast, for \textbf{(b)} and \textbf{(c)}, deep learning models demonstrate superior morphological fidelity.
}
  \label{fig:intro_motivation}
\end{figure}

\section{Related Works}
Missing data mechanisms are classically categorized into three types based on the dependency structure of the missingness~\cite{rubin1976inference}. Missing Completely at Random (MCAR) describes purely stochastic data loss, independent of both observed and unobserved variables; in healthcare, this frequently manifests as random technical failures, such as sensor connectivity dropouts~\cite{rehman2024impact}. In contrast, Missing at Random (MAR) occurs when the probability of missingness is fully conditioned on the observed data, whereas Missing Not at Random (MNAR) arises when the missingness depends on the unobserved values themselves.

Despite the theoretical potential of advanced machine learning for addressing these mechanisms, recent benchmarks in Continuous Glucose Monitoring (CGM) have yielded counterintuitive results. Kuang et al.~\cite{kuang2024imputation} evaluated a broad spectrum of architectures, ranging from deep attention-based models like SAITS \cite{du2023saits} to traditional statistical techniques. Contrary to the anticipated superiority of deep learning, they found that simple Hot-Deck Imputation consistently outperformed complex alternatives across various missingness scenarios. This finding is corroborated by Toye et al.~\cite{toye2025benchmarking}, who benchmarked methods ranging from GP-VAE \cite{fortuin2020gp} to linear interpolation, and Hot-Deck Imputation on glucose and blood pressure datasets. By explicitly stratifying performance across MCAR, MAR, and MNAR regimes, they concluded that simple linear interpolation consistently demonstrated superior performance compared to sophisticated generative approaches.

In the domain of classical techniques, Rehman et al.~\cite{rehman2024impact} extended this analysis by benchmarking methods from linear interpolation to Random Forest (RF) on simulated missingness, identifying RF as the most robust classical reconstructor. However, the fundamental utility of imputation itself remains contested. A contrasting study by \cite{cichosz2025assessing} analyzed the impact of random and segmental missingness on clinical outcomes, ultimately concluding that excluding missing intervals often yields more accurate glycemic metrics than attempting to reconstruct them.

This section reviews the state-of-the-art methods evaluated in this study, categorizing them according to their underlying architectural paradigms:
\subsection{Generative \& Transformer Methods}
SAITS \cite{du2023saits} employs Diagonally-Masked Self-Attention within a dual-stage Transformer, cascading generation and refinement blocks that are fused via learnable weights. GPT4TS \cite{zhou2023one} bridges the modality gap by patchifying time series for a frozen GPT-2 \cite{radford2019language}. It freezes core attention blocks, fine-tuning only embeddings and normalization layers for efficient adaptation.

\subsection{Frequency-Domain Approaches}

FreTS \cite{yi2023frequency} leverages the frequency domain using a two-stage FreMLP architecture. It captures inter-variable and temporal dependencies by processing FFT-transformed data with complex-valued MLPs before inverting to the time domain. TSLANet \cite{eldele2024tslanet} integrates Adaptive Spectral Blocks (ASB) for frequency-domain filtering and Interactive Convolution Blocks (ICB) for multi-scale temporal modeling. The model is optimized by reconstructing masked patches.

\subsection{Multi-Scale Methods}

TimeMixer \cite{wang2024timemixer} handles multiscale variations via Past-Decomposable-Mixing (PDM) for trend/seasonality and Future-Multipredictor-Mixing (FMM) to aggregate predictions. 
SCINet \cite{liu2022scinet} employs a recursive downsampling framework where SCI-Blocks split inputs into even and odd components. An interactive learning mechanism updates these branches to prevent information loss, arranging them in a binary tree before realignment and decoding.

\subsection{Evidence \& Tokenization Frameworks}

TEFN \cite{zhan2025time} models uncertainty by integrating evidence theory with fuzzy mass functions. A Probability Assignment (BPA) module constructs these functions across time and channel dimensions, fusing them via an expectation-based linear transformation for the final forecast. 

TOTEM \cite{talukder2024totem} uses a VQ-VAE to tokenize time series into a shared vocabulary. A strided 1D convolutional encoder compresses inputs by a factor of $F$ before quantization. The frozen encoder and codebook subsequently serve as a deterministic, domain-agnostic tokenizer.


Recent work in time-series benchmarking has highlighted related concerns. TSI-Bench~\cite{du2024tsi} provides a comprehensive imputation benchmark but relies on uniform random masking with no regime stratification. In anomaly detection, similar critiques have emerged: benchmarks dominated by normal operation inflate precision scores and mask detector failures during rare events. Our work addresses the analogous problem for imputation, demonstrating that regime-agnostic evaluation creates a systematic bias favoring simplicity.

To the best of our knowledge, this is the first study to propose a regime-stratified stress test that prioritizes morphological analysis, formally characterizing the Stationarity Bias inherent in random masking.

\section{Methods}

\subsection{Problem Definition}
Let $\mathbf{G} = [g_1, g_2, \dots, g_T]^\top \in \mathbb{R}^T$ denote the fully observed Continuous Glucose Monitoring (CGM) time series of length $T$, where $g_t$ represents the ground truth glucose measurement at time step $t$.

Additionally, let $\mathbf{C} = [\mathbf{c}_1, \mathbf{c}_2, \dots, \mathbf{c}_T]^\top \in \mathbb{R}^{T \times 3}$ represent the concurrent clinical inputs for each time step, including carbohydrate intake flags, as well as bolus and basal insulin dosages. We treat $\mathbf{C}$ as an exogenous forcing variable to better capture the dynamics of glucose excursions.

To capture temporal dependencies and periodic patterns, each time step $t$ is encoded into a continuous feature vector $\phi(t) \in \mathbb{R}^2$ using a sinusoidal time-of-day embedding. Given a sampling interval of 5 minutes, a full 24-hour cycle consists of $T_{\text{day}} = 288$ steps. Let $i = t / T_{\text{day}}$ denote the normalized time index within the daily cycle. The encoding is defined as:

\begin{equation}
    \phi(t) = [\sin(2\pi \cdot i), \, \cos(2\pi \cdot i)]
\end{equation}

\textbf{Simulated Missingness and Training Masks:}
We employ a two-stage masking process to ensure ecological validity. First, we simulate real-world data loss by applying empirical missingness patterns (detailed in Section \ref{sec:missingness_simulation}) to the fully observed ground truth $\mathbf{G}$, yielding a sparse set of "observed" data.

Second, during model training, a binary mask $\mathbf{M} \in \{0, 1\}^T$ is applied exclusively to the observed subset, where $m_t=1$ signifies retention and $m_t=0$ signifies masking. The generation protocol for $\mathbf{M}$ is architecture-dependent: the diffusion-based CSDI~\cite{tashiro2021csdi} model utilizes structural masking of the input data, whereas the attention-based SAITS~\cite{du2023saits} model employs a random (MCAR) masking strategy.

The final model input $\mathbf{x}_t$ is formed by concatenating the masked glucose value (zero-imputed) with concurrent clinical inputs and the time encoding:

\begin{equation}
    \mathbf{x}_t = [ g_t \cdot m_t, \, c_t, \, \phi(t) ]
\end{equation}

The model parameters are learned by minimizing the reconstruction error exclusively at the masked time steps:

\begin{equation}
    \min_\theta \sum_{t=1}^{T} (1 - m_t) \cdot \mathcal{L}(g_t, \hat{g}_t)
\end{equation}

\subsection{Real-World Missingness Pattern Modeling}
\label{sec:missingness_simulation}

We model real-world CGM data loss as a stochastic process estimated from the DCLP3~\cite{brown2019six} and DCLP5~\cite{breton2020randomized} clinical datasets. The process has three components: (1)~an hourly onset probability $P_{\text{start}}(h)$, estimated as the fraction of valid monitoring days on which a gap begins at hour $h$; (2)~a binary classification of each gap as either a transient single-point dropout (5 min, probability $\pi_{\text{short}}$) or a sustained interruption; and (3)~a duration model for sustained gaps, parameterized as a mixture of an Exponential, a Gaussian (centered near the 120-min sensor warm-up period), and a uniform component, with parameters estimated separately for Day ($h \in [6,24)$) and Night ($h \in [0,6)$) regimes. Full derivations, equations, and the generative masking algorithm are provided in Appendix~\ref{sec:missingness_appendix}.

\subsection{Stress Evaluation Protocols}
\subsubsection{Protocol A: Homeostatic State Evaluation}
This protocol evaluates model performance under homeostatic conditions, defined as periods of glycemic stability devoid of exogenous clinical inputs (meal and bolus insulin). Stable $30$-minute time window candidates are identified by satisfying the following five criteria:

\begin{enumerate}
    \item \textbf{Euglycemia:} All glucose points within the window fall within the euglycemic range ($70 \leq g_t \leq 140$ \text{mg/dL}).
    \item \textbf{Gradient Stability:} At least $85\%$ of the time points adhere to a gradient threshold of $|\nabla g_t| < 0.6$ \text{mg/dL/min}.
    \item \textbf{Absence of Exogenous Inputs:} The $30$-minute window is free from meal and bolus events (i.e., $\sum c_t = 0$).
    \item \textbf{Washout Period:} The $60$ minutes immediately preceding the window are free from meal and bolus events, isolating the window from the delayed effects of prior interventions.
    \item \textbf{Low Variability:} The glucose range within the window ($g_{max} - g_{min}$) is $< 25$ \text{mg/dL}, ensuring minimal glycemic excursion.
\end{enumerate}

To simulate specific target durations of missing data, the total required masking length was allocated primarily into full $30$-minute segments. Any residual duration was masked by applying a corresponding partial segment to an available stable window. To evaluate the impact of missing data density, we conducted experiments with masking ratios of $10\%$, $20\%$, and $30\%$ of the total sequence length.

\subsubsection{Protocol B: Post-Prandial Peak Evaluation}
This protocol evaluates the model's ability to reconstruct critical glycemic excursions by specifically targeting post-prandial peaks. First, meals separated by intervals of less than one hour were aggregated into single meal events, as their associated glucose excursions significantly overlap. Subsequently, a four-hour post-prandial window was scanned to identify the peak value for each event. A masking window with a random duration of $3.5$--$4$ hours was then generated and centered around each identified peak. Two constraints were applied to this process: the masking window was restricted from covering pre-meal time steps and from exceeding the total daily duration. As the primary objective was to evaluate the capability of the methods regarding event reconstruction rather than the total ratio of masked time points, the investigation focused on the number of masked post-prandial peaks. Consequently, we tested scenarios where the number of missing post-prandial excursions ranged from $1$ to $3$.

\subsubsection{Protocol C: Temporal Control Reset for Hypoglycemia}
\label{prtcl:pc}

During clinical trials and routine daily management, clinicians and patients frequently adjust insulin pump settings—such as basal rates or insulin sensitivity factors—to mitigate adverse glycemic outcomes \cite{sherr2022automated}. These interventions are often precipitated by impending hypoglycemia (glucose $< 70$ mg/dL). Robust imputation during these critical windows is essential for accurately calibrating the pump's aggressiveness to prevent hypoglycemic events on subsequent days. To evaluate this capability, we utilize the TCR-Simulation dataset (Section~\ref{sec:datasets}), in which we mask a 1-hour window centered on the hypoglycemic event (glucose $< 70$ mg/dL) observed during TCR activation.

\section{Experiments}
    \subsection{Datasets}
    \label{sec:datasets}
    The utilized datasets comprise Continuous Glucose Monitoring (CGM) measurements, timestamps, meal, basal, and bolus insulin, sampled at 5-minute intervals. To ensure evaluation robustness and prevent data leakage, the data is partitioned into training, validation, and testing sets with a 70\%, 10\%, and 20\% ratio, respectively. Crucially, this split is implemented at the patient level, ensuring that the models are evaluated on unseen subjects to assess generalization capability. 
    
    \subsubsection{PEDAP}
    The Pediatric Artificial Pancreas (PEDAP) \cite{wadwa2023trial} study comprises data from 102 children aged 2 to younger than 6 years collected over a 13-week clinical trial. Participants were randomized to receive either hybrid closed-loop insulin delivery or standard care, including multiple daily insulin injections or insulin pump therapy, with continuous glucose monitoring used throughout the study. In this study, we utilize two versions of the PEDAP dataset:
    \begin{itemize}
        \item \textbf{Raw PEDAP:} The data is partitioned into episodes whenever the gap between glucose records exceeds 240 minutes. Missing glucose values are retained as missing (unimputed); however, the remaining parameters—including meal, bolus, and basal insulin—are zero-filled. We utilize Raw PEDAP exclusively for training; notably, missingness simulation is not applied to this dataset. Models trained with this data are tested on the Processed PEDAP dataset.
        \item \textbf{Processed PEDAP:} The data is partitioned into episodes based on gaps larger than 30 minutes. Similar to the Raw PEDAP configuration, auxiliary variables are zero-filled; however, missing CGM values are linearly interpolated. 
    \end{itemize}
    
    \subsubsection{UVA/Padova simulation}
    \label{sec:sim_dataset}
    This dataset was generated using the UVA/Padova simulator \cite{man2014uva}, the only simulator accepted by the FDA for the preclinical testing of artificial pancreas technologies. 
    The cohort consists of 100 adults who consume three meals per day at nominal times varying between 06:00--11:00 (Breakfast), 11:00--13:00 (Lunch), and 18:00--20:00 (Dinner). 
    Actual meal times are subject to random variability ($\sigma = 20$ min). 
    Meal sizes are defined relative to body weight, ranging from 0.84 to 1.44 g/kg, with an added size variability of 15\%. 
    Exercise is not included, and hypoglycemia treatment is triggered when glucose levels drop below 70 mg/dL.

    \subsubsection{TCR-Simulation Dataset}
    This dataset is generated specifically to simulate instances of Temporal Control Reset (TCR). The simulation configuration follows the UVA/Padova Simulation protocol detailed in Section \ref{sec:sim_dataset}, with the exception that temporal variability is removed. To induce hypoglycemia, we introduce a meal size estimation error of +20\% to +30\%, modeling scenarios where patients announce a meal size larger than their actual intake. This overestimation causes the controller to deliver excessive insulin. To simulate the TCR intervention, one meal is selected daily; 2.5 hours post-meal, TCR is activated for a duration of 4 hours. During this window, the basal rate is reduced to 5\% of its nominal value (multiplied by 0.05), representing a 95\% reduction in basal insulin delivery. The TCR-Simulation dataset is utilized exclusively for evaluating Protocol C(Section\ref{prtcl:pc}); no models are trained on this data, and we designated 20\% of the patient cohort as the test set.

\subsection{Metrics}

\textbf{Dynamic Time Warping (DTW):}
Unlike pointwise metrics such as RMSE, Dynamic Time Warping \cite{Vintsyuk1968SpeechDB} assesses the temporal similarity between the predicted sequence $\hat{\mathbf{y}}$ and the ground truth $\mathbf{y}$. It computes the optimal alignment path between the two sequences by minimizing the cumulative distance between aligned points. A lower DTW score indicates a higher degree of similarity between the reconstructed signal shape and the actual physiological trajectory.

\textbf{MARD:}
Mean Absolute Relative Difference (MARD) \cite{freckmann2019measures} is the gold-standard metric for assessing the clinical accuracy of Continuous Glucose Monitoring (CGM) systems. It quantifies the average relative error between the predicted values $\hat{y}_i$ and the ground truth measurements $y_i$. A lower MARD score indicates higher accuracy. The metric is defined as:

\begin{equation}
    \text{MARD} = \frac{1}{N} \sum_{i=1}^{N} \left| \frac{y_i - \hat{y}_i}{y_i} \right| \times 100\%
\end{equation}

Beyond these specialized metrics, we also report standard evaluation measures, including Root Mean Squared Error (RMSE), Empirical Standard Error (EmpSE), and Bias.

\subsection{Results and Discussion}
\subsubsection{\textbf{Scenario A: Validating the Stationarity Hypothesis}}

Table \ref{tab:results_A_all_datasets} in Appendix ~\ref{apdx_full_results} assesses the reconstruction fidelity of imputation methods during homeostatic intervals (Scenario A). The results yield two critical insights:

\begin{itemize}
    \item \textbf{Linearity as the Effective Ground Truth:} Linear Interpolation (Lerp) consistently outperforms deep learning counterparts across all evaluated metrics (RMSE, MARD, DTW, Bias, and Emp-SE). This provides \emph{empirical validation} of our central hypothesis: during periods of physiological stability, the signal manifold is intrinsically low-dimensional. In this regime, linear interpolation is not merely a heuristic baseline; it effectively recovers the ground truth.
    
    \item \textbf{Risks of Over-parameterization:} Conversely, complex deep learning architectures are ill-suited for these low-complexity regimes, often overfitting noise rather than capturing the underlying trend. A clear example appears in the Processed PEDAP dataset (0.3 ratio), where SAITS—despite being the top-performing deep model—lags significantly behind Lerp (RMSE: 6.10 vs. 2.75; DTW: 9.04 vs. 4.48). These results demonstrate that deep models struggle to reconstruct simple linear segments, instead introducing spurious fluctuations (``hallucinations'') that inflate error metrics by over two times. See Figure~\ref{fig:scenario_a_intro} and Appendix~\ref{apdx_c} for qualitative illustrations of these artifacts.
\end{itemize}

\subsubsection{Scenario B: Post-Prandial Excursion} Table \ref{tab:results_B_all_datasets} focuses on the reconstruction of post-prandial excursions. The results provide three key perspectives:

\begin{itemize}
    \item \textbf{The RMSE Mirage:} The mechanism illustrated schematically in Figure~\ref{fig:concept}b manifests empirically in Table~\ref{tab:results_B_all_datasets}. RMSE serves as a poor proxy for the true divergence between an imputation and the ground truth, largely because it ignores the morphological relationship between time sequences. A striking example appears in the Raw PEDAP dataset (two peaks): while the RMSE values for SAITS (49.05), SCINET (57.12), and Lerp (59.21) differ by a narrow margin of only 2--10, their structural differences are vast. In terms of DTW, SAITS achieves a score of 165.60, whereas SCINET and Lerp record 258.61 and 261.82, respectively. This represents a morphological divergence of over 90 between the models, a distinction completely obscured by the RMSE metric. In the Processed PEDAP (one peak) experiment, although Lerp achieves a slightly lower RMSE (57.56 vs. 58.26), SCINet demonstrates superior performance in DTW (225.96 vs. 240.59).
    
    \item \textbf{Morphological Fidelity over Point-wise Error:} The DTW metric reveals that the deep learning model (SAITS) is significantly more successful at reconstructing post-prandial morphology, outperforming baselines across all Scenario B experiments. Unlike linear methods, SAITS exhibits robust morphological consistency even as signal complexity increases. Notably, as the number of peaks rises in the Simulation dataset, SAITS maintains stable performance (DTW: $182.89 \to 197.11 \to 214.51$), whereas the shape disparity in Lerp worsens drastically(DTW: $228.89 \to 249.98 \to 278.54$).
    
    \item \textbf{Clinical Safety Implications:} MARD analysis confirms that the morphological fidelity of deep learning models translates directly into clinically safer imputation. SAITS achieves the lowest MARD across nearly all Scenario B experiments, validating the critical link between accurate shape reconstruction and patient safety. For a closed-loop controller, a ``low-RMSE'' linear imputation that erases physiological peaks is more hazardous than a ``higher-RMSE'' deep learning reconstruction that successfully preserves them.
\end{itemize}

\begin{table*}[h]
\centering
\caption{Performance in Scenario B. \textbf{Bold} indicates best; \underline{underline} indicates second best.}
\label{tab:results_B_all_datasets}
\small
\setlength{\tabcolsep}{2.7pt}
\begin{tabular}{c l ccccc ccccc ccccc}
\toprule
\textbf{Peaks} & \textbf{Model} & \multicolumn{5}{c}{\textbf{Processed PEDAP}} & \multicolumn{5}{c}{\textbf{Raw PEDAP}} & \multicolumn{5}{c}{\textbf{Simulation}} \\
\cmidrule(lr){3-7}\cmidrule(lr){8-12}\cmidrule(lr){13-17}
 &  & \textbf{RMSE} & \textbf{Bias} & \textbf{Emp\_SE} & \textbf{MARD} & \textbf{DTW} & \textbf{RMSE} & \textbf{Bias} & \textbf{Emp\_SE} & \textbf{MARD} & \textbf{DTW} & \textbf{RMSE} & \textbf{Bias} & \textbf{Emp\_SE} & \textbf{MARD} & \textbf{DTW} \\
\midrule
 & FreTS & 61.64 & \textbf{-11.29} & 60.60 & 30.75 & 234.22 & 66.33 & -27.80 & 60.23 & 28.96 & 251.27 & 59.52 & \textbf{-26.75} & 53.17 & 22.96 & \underline{221.40} \\
 & SCINet & 58.26 & -24.21 & \underline{53.00} & \underline{23.32} & \underline{225.96} & \underline{57.38} & \underline{-11.50} & 56.22 & 28.31 & 246.77 & 60.43 & -34.63 & 49.53 & 21.27 & 223.08 \\
 & TimeMixer & 138.97 & -123.36 & 63.98 & 68.21 & 836.12 & 144.71 & -129.48 & 64.63 & 72.18 & 876.02 & 148.70 & -135.36 & 61.55 & 73.41 & 897.82 \\
 & TSLANet & 128.35 & -104.38 & 74.69 & 60.32 & 563.02 & 129.09 & -107.05 & 72.15 & 60.47 & 593.91 & 135.68 & -115.80 & 70.71 & 63.20 & 612.62 \\
 & TEFN & 67.48 & -26.41 & 62.10 & 31.42 & 382.07 & 67.53 & -26.74 & 62.01 & 31.33 & 381.20 & 83.56 & -53.06 & 64.56 & 33.23 & 451.91 \\
1 & TOTEM & 142.86 & -127.35 & 64.72 & 70.84 & 771.04 & 138.72 & -120.50 & 68.72 & 65.62 & 657.29 & 148.81 & -134.63 & 63.40 & 72.65 & 826.37 \\
 & GPT4TS & 63.81 & -25.19 & 58.63 & 28.62 & 251.43 & 60.88 & \textbf{-11.12} & 59.85 & 31.04 & \underline{239.84} & 72.12 & -37.55 & 61.58 & 29.57 & 285.80 \\
 & SAITS & \textbf{52.17} & -15.45 & \textbf{49.83} & \textbf{22.67} & \textbf{186.48} & \textbf{48.60} & -15.97 & \textbf{45.90} & \textbf{19.00} & \textbf{155.42} & \underline{55.23} & \underline{-32.46} & \underline{44.68} & \textbf{19.08} & \textbf{182.89} \\
 & Mean & 67.15 & -18.07 & 64.67 & 32.95 & 380.26 & 67.15 & -18.07 & 64.67 & 32.95 & 380.26 & 73.75 & -39.45 & 62.32 & 28.65 & 389.89 \\
 & Median & 71.76 & -31.10 & 64.67 & 31.83 & 398.36 & 71.76 & -31.10 & 64.67 & 31.83 & 398.36 & 81.55 & -52.60 & 62.32 & 30.91 & 433.69 \\
 & LOCF & 74.66 & -16.75 & 72.76 & 35.90 & 413.15 & 74.66 & -16.75 & 72.76 & 35.90 & 413.15 & 67.00 & -44.24 & 50.32 & 27.46 & 374.41 \\
 & Lerp & \underline{57.56} & \underline{-14.17} & 55.79 & 26.67 & 240.59 & 57.56 & -14.17 & \underline{55.79} & \underline{26.67} & 240.59 & \textbf{53.08} & -36.06 & \textbf{38.94} & \underline{20.57} & 228.89 \\
\midrule
 & FreTS & 61.16 & \textbf{-8.02} & 60.63 & 31.58 & 246.59 & 66.78 & -27.72 & 60.75 & 29.43 & 263.90 & 60.08 & \textbf{-26.47} & 53.93 & 23.60 & 235.11 \\
 & SCINet & \underline{57.86} & -23.73 & \underline{52.77} & \underline{23.28} & \underline{237.43} & \underline{57.12} & \textbf{-11.42} & \underline{55.96} & 28.23 & 258.61 & 60.88 & -34.84 & 49.92 & \underline{21.59} & \underline{235.10} \\
 & TimeMixer & 139.01 & -123.74 & 63.34 & 69.00 & 874.68 & 143.50 & -128.50 & 63.87 & 72.12 & 907.44 & 148.18 & -134.92 & 61.27 & 73.34 & 904.71 \\
 & TSLANet & 127.13 & -103.44 & 73.91 & 60.18 & 568.40 & 127.92 & -106.09 & 71.47 & 60.35 & 600.61 & 135.06 & -115.35 & 70.26 & 63.14 & 605.53 \\
 & TEFN & 68.37 & -28.50 & 62.14 & 31.49 & 404.55 & 68.42 & -28.82 & 62.05 & 31.41 & 403.71 & 85.69 & -58.26 & 62.84 & 33.89 & 470.31 \\
2 & TOTEM & 141.84 & -126.59 & 63.98 & 70.89 & 798.62 & 137.95 & -120.18 & 67.73 & 66.00 & 682.75 & 148.22 & -134.18 & 62.98 & 72.60 & 831.67 \\
 & GPT4TS & 64.85 & -27.26 & 58.85 & 28.80 & 269.86 & 61.84 & \underline{-13.98} & 60.24 & 30.89 & \underline{256.31} & 74.84 & -44.49 & 60.18 & 29.99 & 310.95 \\
 & SAITS & \textbf{53.71} & -18.46 & \textbf{50.44} & \textbf{22.91} & \textbf{201.20} & \textbf{49.05} & -16.61 & \textbf{46.15} & \textbf{19.04} & \textbf{165.60} & \textbf{56.50} & \underline{-33.57} & \underline{45.45} & \textbf{19.68} & \textbf{197.11} \\
 & Mean & 66.92 & -19.87 & 63.91 & 32.34 & 394.74 & 66.92 & -19.87 & 63.91 & 32.34 & 394.74 & 77.40 & -46.60 & 61.80 & 29.80 & 415.78 \\
 & Median & 71.96 & -33.07 & 63.91 & 31.51 & 417.11 & 71.96 & -33.07 & 63.91 & 31.51 & 417.11 & 83.69 & -56.44 & 61.80 & 32.01 & 453.32 \\
 & LOCF & 75.72 & -18.14 & 73.51 & 36.34 & 435.87 & 75.72 & -18.14 & 73.51 & 36.34 & 435.87 & 70.00 & -46.75 & 52.11 & 27.83 & 388.22 \\
 & Lerp & 59.21 & \underline{-15.01} & 57.27 & 27.62 & 261.82 & 59.21 & -15.01 & 57.27 & \underline{27.62} & 261.82 & \underline{56.81} & -38.75 & \textbf{41.54} & 21.60 & 249.98 \\
\midrule
 & FreTS & 61.07 & \textbf{-6.43} & 60.73 & 32.23 & 264.74 & 66.63 & -25.24 & 61.67 & 30.14 & \underline{274.65} & \underline{61.73} & \textbf{-27.62} & 55.20 & 23.99 & 250.58 \\
 & SCINet & \underline{57.22} & -22.63 & \underline{52.56} & \textbf{23.29} & \underline{252.92} & \underline{57.09} & \textbf{-10.93} & \underline{56.03} & \underline{28.44} & 276.43 & 62.37 & -36.05 & 50.89 & \underline{21.88} & \underline{249.77} \\
 & TimeMixer & 138.55 & -123.48 & 62.83 & 69.60 & 918.60 & 141.88 & -127.00 & 63.26 & 71.92 & 944.54 & 148.81 & -135.50 & 61.50 & 73.47 & 916.59 \\
 & TSLANet & 125.51 & -101.82 & 73.38 & 59.91 & 567.87 & 126.29 & -104.54 & 70.86 & 60.06 & 603.43 & 135.51 & -115.90 & 70.21 & 63.29 & 603.56 \\
 & TEFN & 69.17 & -30.52 & 62.08 & 31.63 & 431.36 & 69.23 & -30.81 & 61.99 & 31.56 & 430.52 & 89.92 & -64.98 & 62.16 & 35.29 & 499.73 \\
3 & TOTEM & 140.49 & -125.36 & 63.43 & 70.83 & 831.48 & 136.95 & -119.55 & 66.81 & 66.38 & 712.14 & 148.72 & -134.70 & 63.04 & 72.71 & 841.66 \\
 & GPT4TS & 65.93 & -29.35 & 59.04 & 29.13 & 296.37 & 62.91 & \underline{-17.07} & 60.55 & 30.86 & 281.13 & 80.41 & -53.95 & 59.63 & 31.36 & 345.29 \\
 & SAITS & \textbf{55.51} & -21.39 & \textbf{51.22} & \underline{23.35} & \textbf{223.26} & \textbf{49.65} & -17.25 & \textbf{46.56} & \textbf{19.27} & \textbf{181.03} & \textbf{59.05} & \underline{-35.84} & \underline{46.94} & \textbf{20.39} & \textbf{214.51} \\
 & Mean & 66.50 & -20.43 & 63.28 & 32.06 & 413.02 & 66.50 & -20.43 & 63.28 & 32.06 & 413.02 & 81.29 & -52.87 & 61.75 & 31.02 & 441.78 \\
 & Median & 71.61 & -33.52 & 63.28 & 31.32 & 437.54 & 71.61 & -33.52 & 63.28 & 31.32 & 437.54 & 85.91 & -59.72 & 61.75 & 32.86 & 470.71 \\
 & LOCF & 76.25 & -21.41 & 73.18 & 36.41 & 464.48 & 76.25 & -21.41 & 73.18 & 36.41 & 464.48 & 74.82 & -51.48 & 54.29 & 29.01 & 413.80 \\
 & Lerp & 61.31 & \underline{-17.21} & 58.84 & 28.55 & 294.14 & 61.31 & -17.21 & 58.84 & 28.55 & 294.14 & 61.81 & -42.66 & \textbf{44.73} & 23.08 & 278.54 \\
\bottomrule
\end{tabular}
\end{table*}

\subsubsection{\textbf{Scenario C: Temporal Control Reset During Hypoglycemia}}
As shown in Table~\ref{tab:results_C_all_datasets}, the deep learning model (SAITS) demonstrates superior performance in both shape reconstruction and clinical safety compared to Lerp. Beyond a $\approx 10.50$ improvement in RMSE over Lerp, SAITS yields a critical $11.29\%$ reduction in MARD and a substantial $26.14$ improvement in morphological fidelity (DTW). These gains are vital for maintaining the stability of closed-loop control systems, specifically by preventing controller miscalibration caused by inaccurate hypoglycemia imputation.

Regarding bias, Lerp exhibits a high positive bias ($+25.55$), indicating a systematic overestimation of glucose values. In a hypoglycemic context, this is hazardous as it may mask low glucose events, leading to missed interventions. In contrast, SAITS exhibits a lower positive bias ($+14.08$). Although both models pose a clinical risk by potentially masking hypoglycemia, SAITS represents a improvement.

\begin{table}[h]
    \centering
    \caption{Performance on the TCR simulation dataset (Scenario C). \textbf{Bold} indicates best, \underline{underline} indicates second best.}
    \label{tab:results_C_all_datasets}
    \small
    \setlength{\tabcolsep}{4pt}
    
    \begin{tabular}{l c c c c c}
    \toprule
    \textbf{Model} & \textbf{RMSE} & \textbf{Bias} & \textbf{Emp\_SE} & \textbf{MARD} & \textbf{DTW} \\
    \midrule
    FreTS & 58.41 & 52.63 & 25.35 & 72.46 & 203.35 \\
    SCINet & \underline{32.44} & 27.18 & \underline{17.71} & 39.52 & 101.07 \\
    TimeMixer & 43.81 & -38.92 & 20.10 & 43.93 & 179.61 \\
    TSLANet & 44.43 & -23.50 & 37.71 & 44.28 & 165.33 \\
    TEFN & 57.25 & 51.49 & 25.03 & 71.89 & 227.88 \\
    TOTEM & 41.65 & -37.98 & \textbf{17.11} & 44.01 & 125.45 \\
    GPT4TS & 69.87 & 63.64 & 28.84 & 87.45 & 249.86 \\
    SAITS & \textbf{23.26} & \textbf{14.08} & 18.51 & \textbf{27.00} & \textbf{66.38} \\
    Mean & 68.43 & 65.28 & 20.50 & 88.76 & 283.91 \\
    Median & 51.97 & 47.76 & 20.50 & 66.83 & 215.05 \\
    LOCF & 38.90 & \underline{22.63} & 31.64 & 43.24 & 140.24 \\
    Lerp & 33.76 & 25.55 & 22.07 & \underline{38.29} & \underline{92.52} \\
    \bottomrule
    \end{tabular}
\end{table}

\subsubsection{\textbf{Distributional Calibration Analysis}}
\begin{figure*}[t]
    \centering
    \begin{subfigure}[t]{\textwidth}
        \centering
        \includegraphics[width=\textwidth]{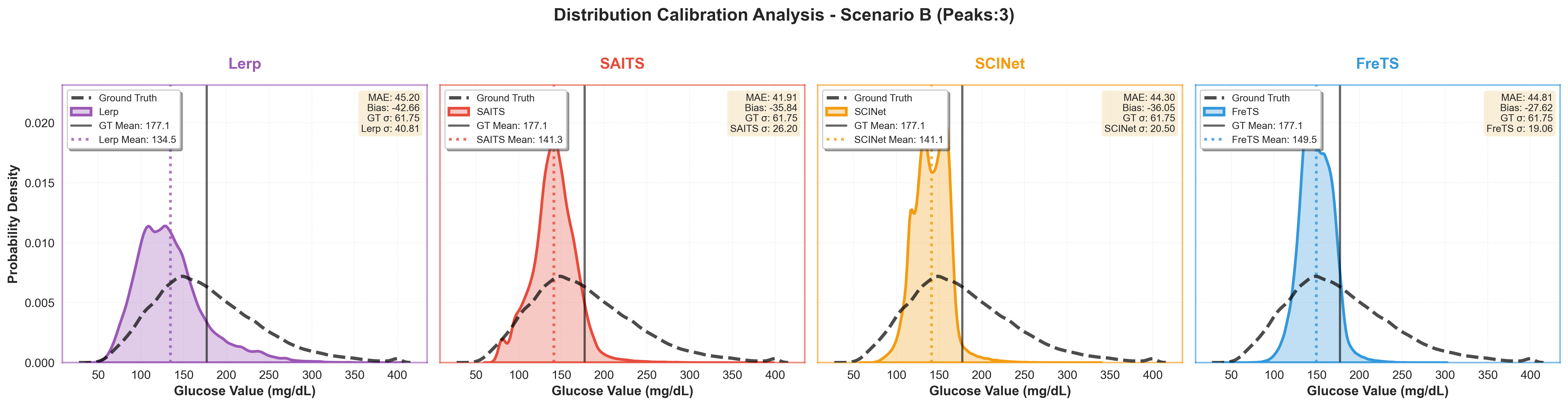}
        \caption{Scenario B (Post-prandial transients, 3 peaks, Simulation dataset): Lerp attenuates the distribution by $42.6$\,mg/dL; SAITS and FreTS partially recover the ground-truth distribution, reducing the offset to $35.8$\,mg/dL and $27.6$\,mg/dL, respectively.}
        \label{fig:calib_B}
    \end{subfigure}
    \vspace{0.3cm}
    \begin{subfigure}[t]{\textwidth}
        \centering
        \includegraphics[width=\textwidth]{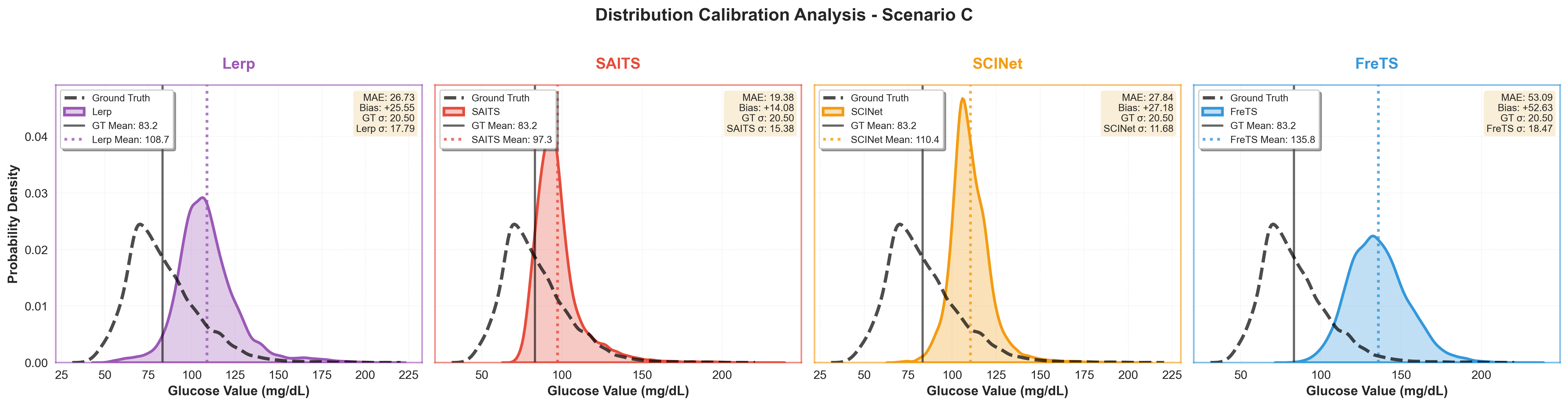}
        \caption{Scenario C (Hypoglycemic transients during TCR): Lerp overestimates by $+25.5$\,mg/dL, shifting imputations above the hypoglycemia threshold; SAITS exhibits improved but persistent miscalibration ($\Delta = +14.08$\,mg/dL).}
        \label{fig:calib_C}
    \end{subfigure}
    
    \caption{
    Distributional calibration analysis. Each panel shows the 
    conditional density of imputed values (shaded) against the 
    ground-truth density (dashed black) during transient regimes. 
    Vertical lines indicate distribution means.}
    \label{fig:calibration}
\end{figure*}
The preceding analyses assessed imputation quality through pointwise 
(RMSE, MARD), morphological (DTW), and bias metrics. We now examine 
a complementary perspective: \emph{distributional calibration}---whether 
the imputed values occupy the correct region of the measurement space. 
For each model, we estimate the conditional density $p(\hat{y} \mid y 
\in \mathcal{R})$, where $\mathcal{R}$ denotes the ground-truth 
regime (transient or hypoglycemic), and compare it against the 
ground-truth density $p(y \mid y \in \mathcal{R})$.

\paragraph{Post-Prandial Transients (Scenario B)}
Figure~\ref{fig:calib_B} reveals a striking distributional shift under both linear interpolation and deep learning models. While the ground-truth distribution during post-prandial peaks has a mean of $177.1$~mg/dL, Lerp produces imputations centered at $134.5$~mg/dL---a systematic attenuation of $42.6$~mg/dL that compresses the imputed distribution well below the true glycemic range. This is the distributional signature of corner-cutting: by drawing a chord through each peak, Lerp maps high-glucose transients into the euglycemic range. In contrast, SAITS partially recovers the distribution with a mean of $141.3$~mg/dL ($\Delta = 35.8$~mg/dL) and a standard deviation of $\sigma = 26.2$~mg/dL (vs. ground truth $\sigma = 61.75$~mg/dL). FreTS achieves better mean recovery ($149.5$~mg/dL) but exhibits an even narrower distribution ($\sigma = 19.06$~mg/dL). These results demonstrate that while deep learning models remarkably improve shape recovery compared to linear interpolation, capturing the full distributional spread of post-prandial excursions remains a significant challenge.


\paragraph{Hypoglycemic Transients (Scenario C)}
As illustrated in Figure~\ref{fig:calib_C}, the ground-truth distribution during hypoglycemic events centers at $83.2$~mg/dL. In contrast, Lerp shifts the imputed distribution to $108.7$~mg/dL---a substantial $+25.5$~mg/dL overestimation that systematically elevates predictions well above the clinical hypoglycemia threshold ($70$~mg/dL). While SAITS mitigates this error (mean $97.3$~mg/dL, bias $+14.08$~mg/dL), it still fails to align the distribution with the ground truth. For a closed-loop controller, this shift—present in both models—is hazardous: it creates a false impression of euglycemia, potentially causing the system to withhold necessary interventions during low-glucose events.

\paragraph{Implications.}
These results expose a critical blind spot in standard evaluation: \emph{distributional miscalibration}, whereby generated trajectories are drawn from a clinically invalid region of the measurement space.
Our distributional analysis confirms that while deep learning significantly improves shape recovery during transient states (Tables~\ref{tab:results_B_all_datasets} and~\ref{tab:results_C_all_datasets}), reliable reconstruction of physiological excursions remains an open challenge. Figures~\ref{fig:calib_B} and~\ref{fig:calib_C} demonstrate that although our models achieve lower variance and better mean alignment than linear baselines, they remain far from the ground-truth center. This systematic offset suggests that current architectures fundamentally misinterpret essential exogenous variables as noise rather than as deterministic signal drivers—a limitation whereby models fail to leverage the presence of exogenous inputs to appropriately drive glucose signal reconstruction, as detailed in~\cite{shakeri2025driver}.

In summary, although deep learning does not fully resolve the challenge of excursion reconstruction, it offers a definitive improvement over linear baselines in calibration, clinical safety, and morphological fidelity. The persistence of distributional errors does not render these models ineffective; rather, it delineates a specific target for future architectural refinement. Crucially, the substantial reduction in clinical risk relative to Lerp establishes deep learning as the preferred approach for handling transient regions, providing a safer—if not yet perfect—approximation of ground-truth dynamics. Ablation studies validating and investigating the sensitivity of our scenario-based evaluation are available in Appendix \ref{ablation_appx}, which support the aforementioned findings.

\section{Practical Implications: Toward Adaptive Inference}

Our findings suggest a heterogeneous deployment strategy; because homeostatic intervals dominate the signal and linear interpolation proves optimal for these stationary regimes (Table~\ref{tab:results_A_all_datasets}), deploying a Transformer globally is computationally inefficient---despite its safety advantages during transients. Conversely, relying exclusively on linear interpolation is clinically unsafe given the model's failure during critical excursions (Tables~\ref{tab:results_B_all_datasets}--\ref{tab:results_C_all_datasets}).

This tension motivates a \emph{regime-conditional} model selection strategy: a lightweight detection framework---utilizing the gradient stability criterion from Protocol~A ($|\nabla g_t| < 0.6$\,mg/dL/min)---routes each missing segment to the appropriate method. Stationary gaps are resolved via linear interpolation at negligible cost, while transient gaps automatically trigger deep learning inference.

This is not merely a computational convenience. In resource-constrained clinical devices (e.g., insulin pumps with embedded processors), full Transformer inference at every missing interval may be infeasible. An adaptive strategy reduces average inference cost to a fraction of always-on deep learning while preserving morphological fidelity precisely where it is safety-critical.

These findings provide the conceptual blueprint for an adaptive inference framework. While this section outlines the architectural logic derived from our stationarity analysis, it establishes a foundation for future engineering optimizations—specifically regarding threshold sensitivity and real-time latency—necessary for deployment in closed-loop control systems.

\subsection{Implementation Details}
All models were implemented using the PyPOTS library~\cite{du2024tsi} with a fixed batch size of 32. For data preprocessing, we employed a sliding window approach with a sequence length of 288 timesteps (corresponding to 24 hours). The stride was set to 57 for the training set and increased to 128 for the test set. To mitigate overfitting, we applied early stopping based on validation loss with a patience of 10 epochs, capping the maximum training duration at 100 epochs. Hyperparameter optimization was conducted using Optuna~\cite{akiba2019optuna} over 40 trials per model, and the best-performing configuration was selected for final evaluation. To ensure reproducibility, a fixed random seed of 7 was used across all experiments.

\section{Limitations and Ethical Considerations}

This study utilized the publicly available DCLP3~\cite{brown2019six}, DCLP5~\cite{breton2020randomized}, and PEDAP~\cite{wadwa2023trial} datasets for secondary analysis. Because the data are de-identified, open-source, and involved no direct human interaction, ethical approval was not required. Furthermore, synthetic data were generated using the UVA/Padova simulator~\cite{man2014uva}, a platform widely recognized and accepted by the FDA for preclinical evaluation.

Despite these contributions, this study is subject to certain limitations that suggest directions for future work. First, our experimental evaluation could be expanded to include comparisons with both classical machine learning baselines, such as Random Forest and k-Nearest Neighbors (k-NN), and advanced deep learning architectures, such as diffusion models~\cite{tashiro2021csdi}. In addition, a potential challenge in Scenario B involves the 'late-logging' phenomenon, where meal entries might be recorded mid-peak or retrospectively. However, our empirical evaluation demonstrates that this temporal misalignment has a negligible impact on performance, yielding trends and conclusions consistent with our simulation results. This alignment suggests that the proposed method is robust to the reporting lags inherent in real-world diabetes datasets. Finally, the practical utility of our method in downstream tasks, specifically within diabetes control systems, remains to be validated.

\section{Conclusion}

We have identified and formalized the Stationarity Bias in time-series imputation evaluation: when missing segments are sampled uniformly from signals with a dominant stationary regime, benchmarks systematically overestimate the performance of simple methods and underestimate the value of learned models. This bias creates an \emph{RMSE Mirage}---low pointwise error that masks the destruction of signal morphology during critical transients.

Through three regime-stratified stress tests on both real-world and simulated CGM datasets, we established a clear regime-conditional model selection principle: linear interpolation is preferred during stationary intervals, where it avoids the spurious artifacts of over-parameterized models; deep learning (SAITS) is essential during transients, where it significantly outperforms baselines in morphological fidelity, clinical accuracy, and safety-relevant bias.

While validated on physiological data, the Stationarity Bias is not domain-specific. Any system that spends the majority of its time near a stable operating point---server infrastructure, industrial processes, power grids, financial markets---will exhibit the same evaluation pathology under uniform random masking. We hope this work motivates the adoption of regime-stratified evaluation as standard practice in time-series imputation benchmarking.




\section{GenAI Disclosure}
Generative AI tools were used for minor editorial assistance, including grammar and wording refinement.

\bibliographystyle{ACM-Reference-Format}
\bibliography{Reference}

%
\appendix
\section{Missingness Pattern Modeling: Full Derivation}
\label{sec:missingness_appendix}
To accurately simulate realistic data loss, the missingness mechanism is modeled as a stochastic process derived from real-world CGM datasets. This process proceeds in two distinct phases: (1) Empirical Parameter Estimation and (2) Generative Masking, as detailed in Algorithm \ref{alg:missingness_short}.

\subsubsection{Data Preprocessing and Gap Identification}
Let $\mathcal{D} = \{(\mathbf{t}, \mathbf{g})\}$ denote the raw dataset. The time series is first resampled to a uniform grid with a sampling interval of $\Delta t = 5$ minutes. To ensure the reliability of distribution statistics, the dataset is filtered to retain only valid days. Let $O_{obs}^{(d)}$ be the number of observed points on day $d$, and $O_{max} = 288$ be the maximum expected points per day. The set of valid days $\mathcal{V}$ is defined as:

\begin{equation}
    \mathcal{V} = \left\{ d \mid \frac{O_{obs}^{(d)}}{O_{max}} \geq 0.50 \right\}
\end{equation}

Within $\mathcal{V}$, the set of all data gaps $\mathcal{G} = \{(\tau_i, \delta_i)\}_{i=1}^K$ is identified, where $\tau_i$ represents the start time (hour of day) and $\delta_i$ represents the duration of the $i$-th gap.

\subsubsection{Hourly Onset Probability}

\begin{figure}[h]
    \centering
    \includegraphics[width=\linewidth]{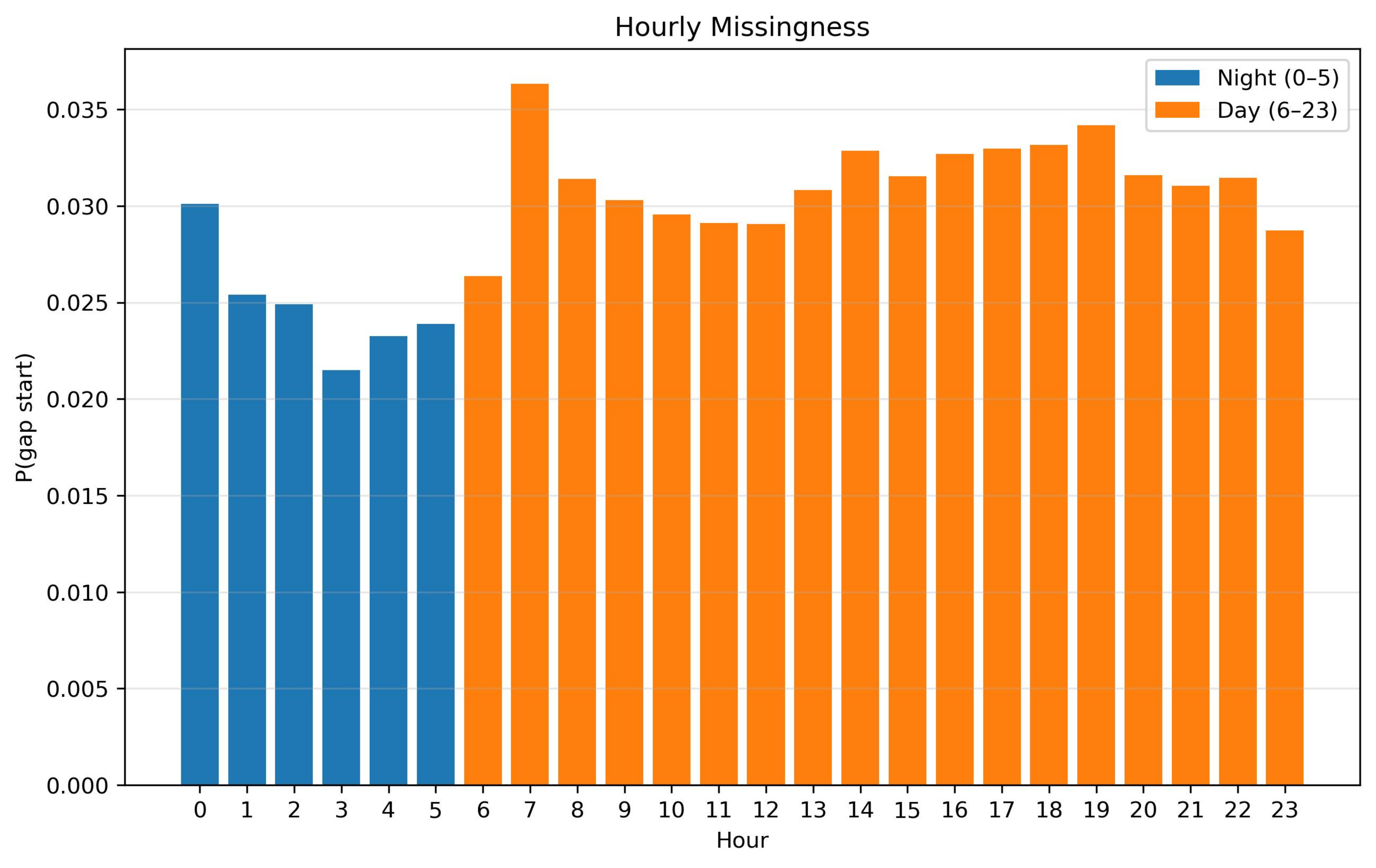}
    \caption{Hourly probability of a missingness gap beginning during the day.}
    \label{fig:hourly_missingness}
\end{figure}

We model the likelihood of a missing data gap beginning at hour $h \in \{0, \dots, 23\}$ as a Bernoulli probability. This parameter is estimated as the ratio of the number of days where a gap starts at hour $h$ (denoted as $N_h^{\text{day}}$) to the total number of valid monitoring days ($|\mathcal{V}|$). Figure \ref{fig:hourly_missingness} illustrates the hourly probability of missingness. Under this framework, each hour $h$ is treated as an independent Bernoulli distribution characterizing the binary risk of a gap originating at that specific time:
\begin{equation}
P_{\text{start}}(h) = \frac{N_h^{\text{day}}}{|\mathcal{V}|}
\end{equation}

\subsubsection{Duration Distribution Modeling}
The duration of missingness intervals, denoted as $\delta \in \{5, 10, \dots, 240\}$ minutes, is modeled using a two-stage distribution to distinguish between single-point dropouts (noise) and longer systemic interruptions. 

\begin{figure*}[t]
    \centering
    \begin{subfigure}{0.49\linewidth}
        \centering
        \includegraphics[width=\linewidth]{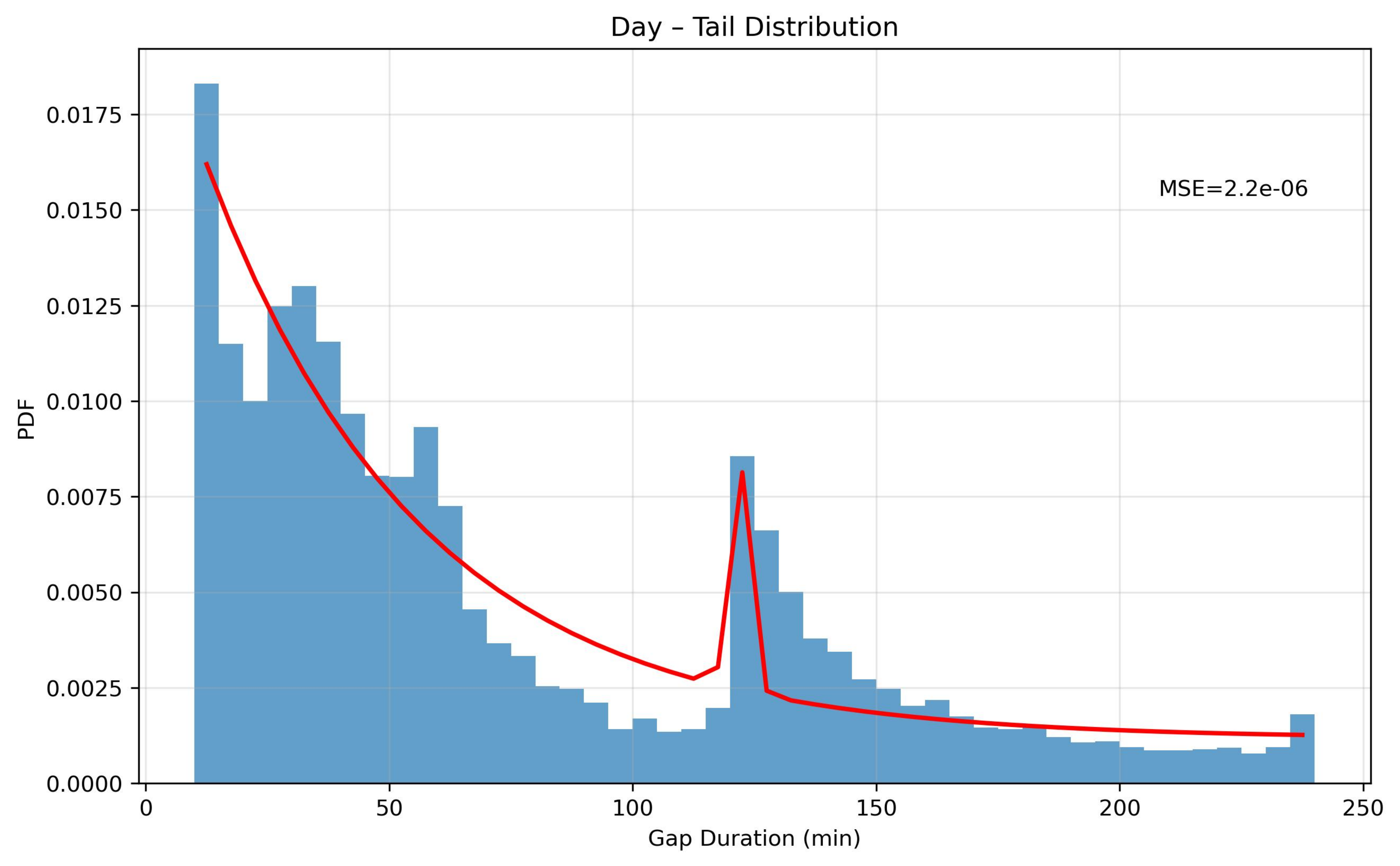}
        \caption{Day}
        \label{fig:day_tail}
    \end{subfigure}
    \hfill
    \begin{subfigure}{0.49\linewidth}
        \centering
        \includegraphics[width=\linewidth]{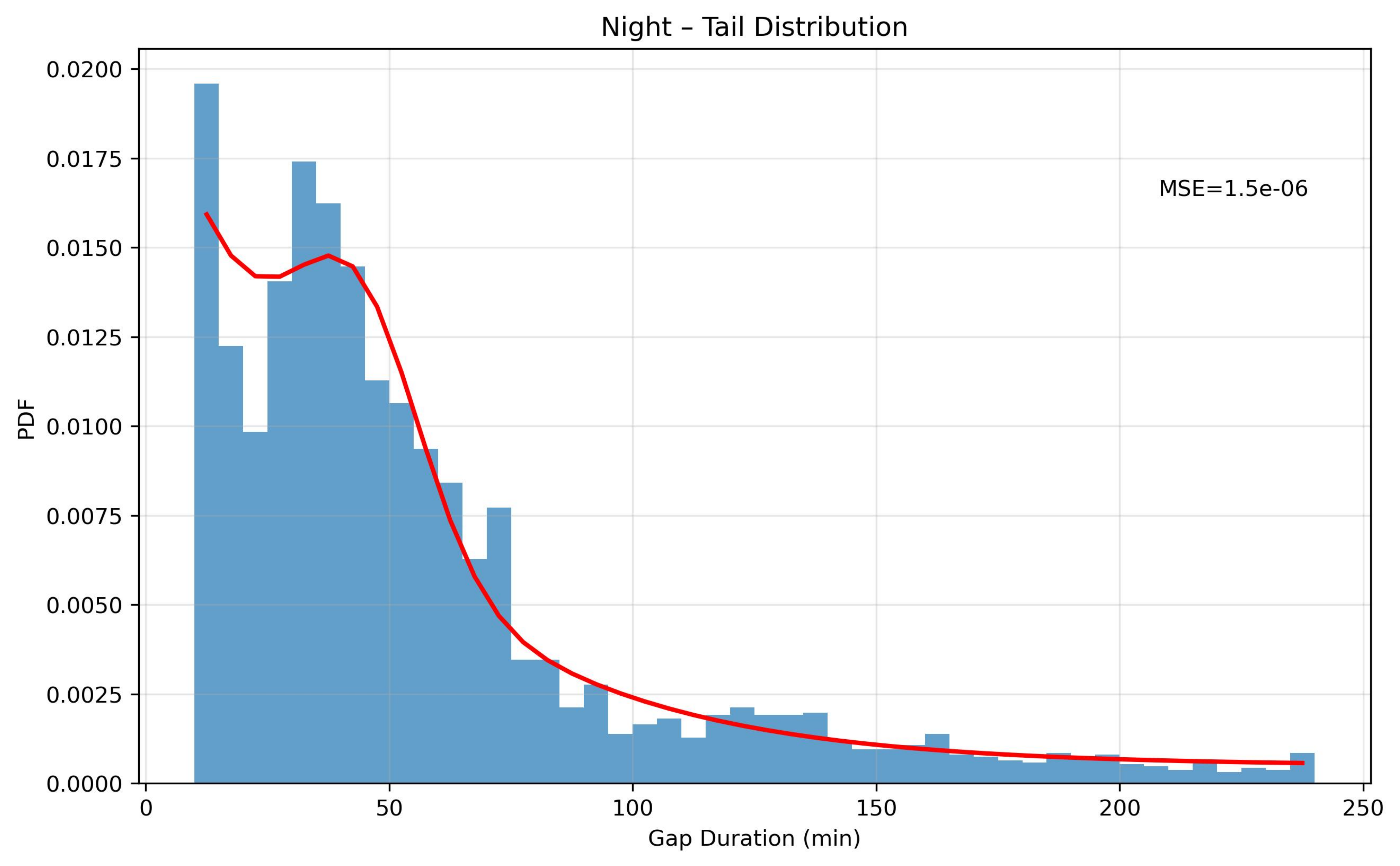}
        \caption{Night}
        \label{fig:night_tail}
    \end{subfigure}

    \caption{\textbf{Tail Density of Missingness Durations.}
    Empirical tails overlaid with the fitted mixture model (Eq.~\ref{eq:duration_dist}).}
    \label{fig:gap_tail}
\end{figure*}

Moreover, we stratify gaps based on their onset time into two distinct periods: \textit{Night} ($h \in [0, 6)$) and \textit{Day} ($h \in [6, 24)$). This distinction addresses behavioral patterns related to device maintenance and the underlying causes of missingness. Sensor replacements—which typically mandate a 120-minute warmup period (e.g., Dexcom 6 \cite{garg2022accuracy} sensors)—occur predominantly during waking hours. Consequently, as expected, Figure \ref{fig:day_tail} shows a distinct probability peak at $\delta \approx 120$ minutes in the daytime distribution. Conversely, gaps initiating during overnight hours (sleeping periods) are typically shorter, reflecting transient connectivity issues or Pressure-Induced Sensor Attenuation (PISA) \cite{facchinetti2016modeling} rather than planned sensor maintenance. The following two stages are applied independently to both the \textit{Day} and \textit{Night} periods.

\paragraph{Stage 1: Single Dropouts}
The probability of a ``noise'' dropout (duration exactly 5 minutes) is estimated as the proportion of such events within the short-term gap category:

\begin{equation}
    \pi_{\text{short}} = P(\delta = 5) = \frac{\sum_{i=1}^K \mathbb{I}(\delta_i = 5)}{|\delta|}
\end{equation}

where $\mathbb{I}(\cdot)$ is the indicator function.

\paragraph{Stage 2: Complex Gap Durations}
For gaps longer than 5 minutes ($\delta > 5$), the distribution is modeled using a mixture model combining an Exponential component, a Gaussian component, and a constant offset. The density function $f(\delta)$ is defined as:

\begin{equation}
f(\delta) = A e^{-k(\delta - 10)} + B e^{-\frac{(\delta-\mu)^2}{2\sigma^2}} + \gamma \cdot \mathbb{I}(10 \le \delta \le \delta_{max})
\label{eq:duration_dist}
\end{equation}

\begin{figure*}[t]
    \centering
    \begin{subfigure}{0.49\linewidth}
        \centering
        \includegraphics[width=\linewidth]{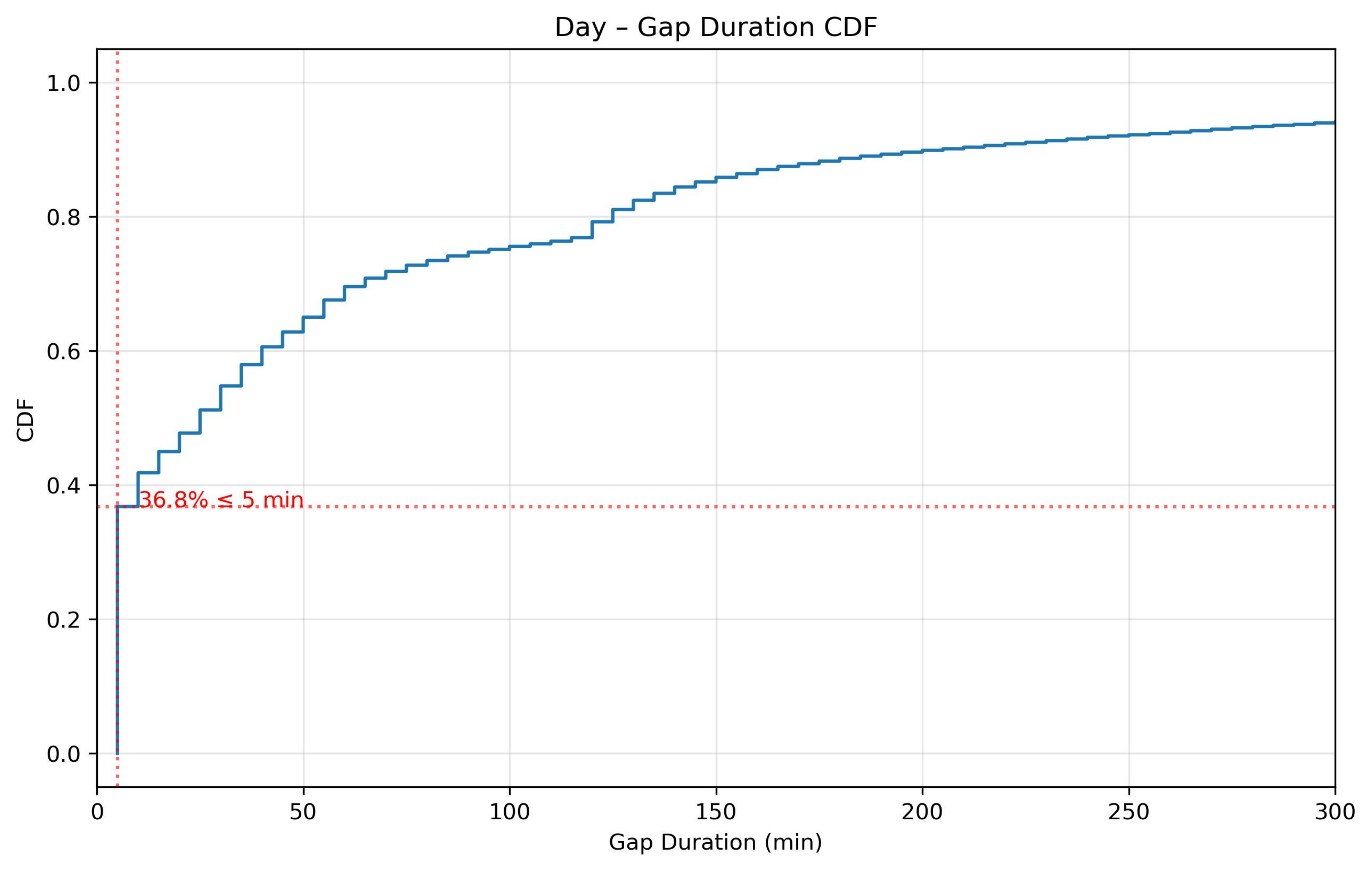}
        \caption{Day}
        \label{fig:day_cdf}
    \end{subfigure}
    \hfill
    \begin{subfigure}{0.49\linewidth}
        \centering
        \includegraphics[width=\linewidth]{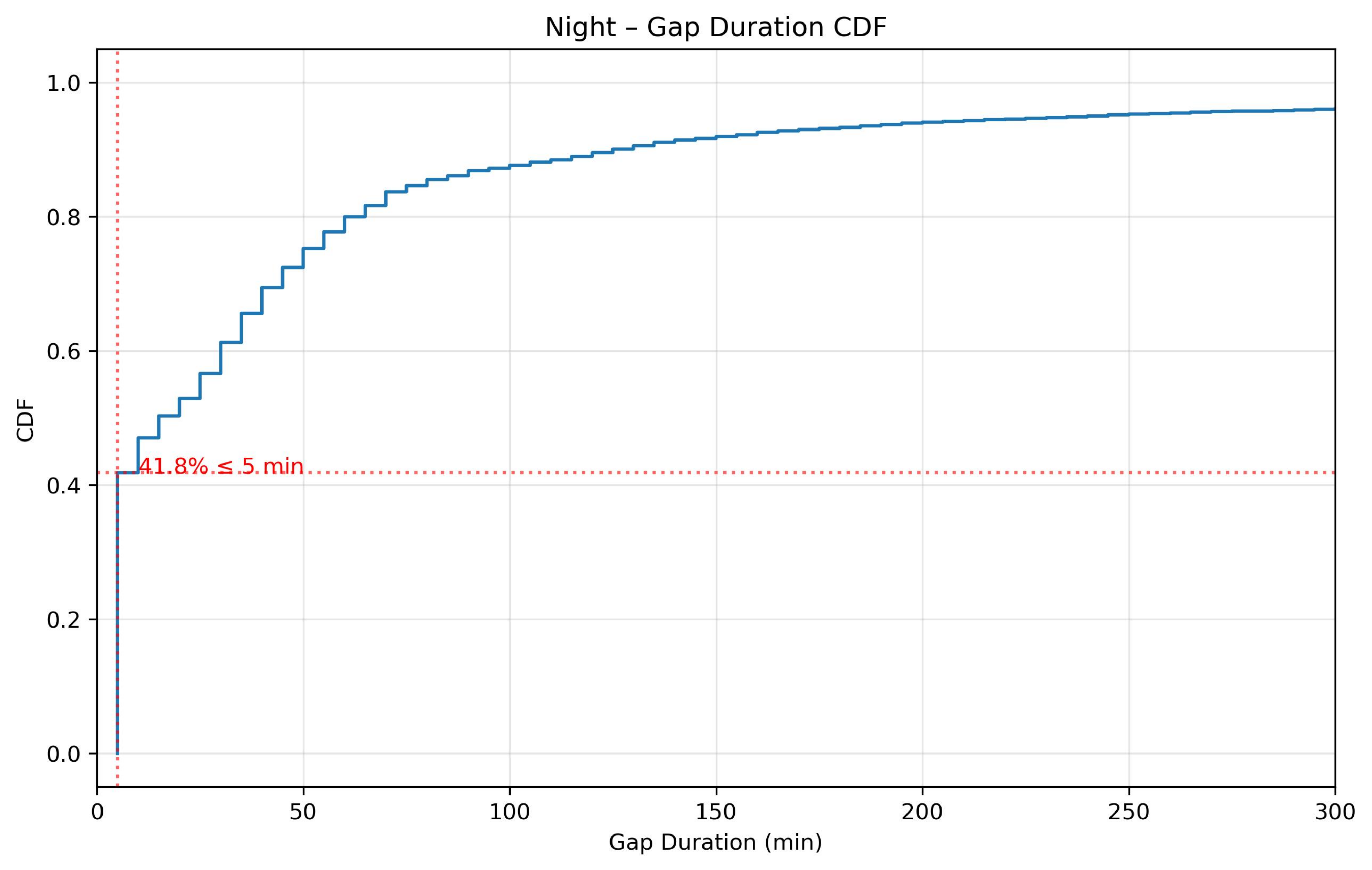}
        \caption{Night}
        \label{fig:night_cdf}
    \end{subfigure}

    \caption{\textbf{Empirical cumulative distribution function (CDF) of missingness durations.}
    Comparison between day and night regimes.}
    \label{fig:gap_cdf}
\end{figure*}

The parameters $\Theta = \{A, k, B, \mu, \sigma, \gamma\}$ are estimated via non-linear least squares optimization \cite{vugrin2007confidence} on the empirical histogram of durations within the range $10 \le \delta \le 240$ minutes. We impose an explicit maximum duration cap of $\delta_{max} = 240$ minutes; as illustrated in Figures \ref{fig:day_cdf}, \ref{fig:night_cdf}, gap durations exceeding this threshold effectively vanish in the empirical distribution. Consequently, these extreme outliers are considered statistically negligible and are excluded from our experimental simulation to ensure stability.

\subsubsection{Generative Masking Process}

To simulate missingness on a fully observed sequence $\mathbf{X}$, we generate a binary mask $\mathbf{B} \in \{0, 1\}^T$. For each hour $h$, a missingness event is triggered according to a Bernoulli process with probability $P_{\text{start}}(h)$.
The nature of each event is determined hierarchically. First, we classify the dropout as either a transient pulse or a sustained gap based on the calculated $\pi_{\text{short}}$. Given a random sample $u' \sim \mathcal{U}(0,1)$:

\begin{itemize}
    \item \textbf{Transient Noise:} If $u' < \pi_{\text{short}}$, a single-point dropout (5 minutes) occurs.
    \item \textbf{Sustained Gap:} Otherwise, a duration $\hat{\delta} \in [10, 240]$ minutes is drawn from a regime-specific (Day or Night) mixture model.
\end{itemize}

For sustained gaps, we derive the normalized weights $w_{\text{exp}}$ and $w_{\text{gauss}}$ by computing the relative probability masses of the Exponential and Gaussian components within the $[10, 240]$ interval. The specific sampling distribution for an interval of duration $\hat{\delta}$ is then selected via a second random draw $u \sim \mathcal{U}(0,1)$:

\begin{equation}
\hat{\delta} \sim 
\begin{cases} 
\text{Exp}(\lambda = 1/k) & \text{if } u < w_{\text{exp}} \\
\mathcal{N}(\mu, \sigma)_{[10, 240]} & \text{if } w_{\text{exp}} \le u < w_{\text{exp}} + w_{\text{gauss}} \\
\mathcal{U}(10, 240) & \text{otherwise (constant offset } \gamma)
\end{cases}
\end{equation}

Finally, a start time $\tau$ is sampled uniformly within the hour $h$. The mask $\mathbf{B}$ is set to 0 for the interval $\hat{\delta}$, effectively "zeroing out" the corresponding observations in $\mathbf{X}$.

\begin{algorithm}[H]
\caption{Realistic Missingness Simulation}
\label{alg:missingness_short}
\begin{algorithmic}[1]
\Require Sequence length $T$, Hourly risk $P_{\text{start}}(h)$, Mixture params $(\Theta_r,\mathbf{w}_r)$, Noise prob $\pi_{\text{short}}$
\Ensure Mask $\mathbf{B}\in\{0,1\}^T$
\State $\mathbf{B}\gets \mathbf{1}_T$, \ $t\gets 0$
\While{$t<T$}
    \State $h\gets \mathrm{Hour}(t)$; \ $t_{\text{next}}\gets \mathrm{NextHourIndex}(t)$
    \If{$\mathrm{Bernoulli}\!\big(P_{\text{start}}(h)\big)$}
        \State $r\gets \mathrm{Regime}(h)$ \Comment{Day/Night}
        \State Draw $u'\sim \mathcal{U}[0,1]$
        \If{$u'<\pi_{\text{short}}$}
            \State $\hat{\delta}\gets 5$ \Comment{Noise}
        \Else
            \State $\hat{\delta}\sim \mathrm{Mixture}(\Theta_r,\mathbf{w}_r)$ \Comment{Sustained gap}
        \EndIf
        \State $L\gets \lceil \hat{\delta}/5\rceil$; \ $t_{\text{start}}\gets t+\mathrm{Unif}\{0,\dots,t_{\text{next}}-t-1\}$
        \State $\mathbf{B}[t_{\text{start}}:\min(t_{\text{start}}+L,\,T)]\gets 0$
        \State $t\gets t_{\text{start}}+L$
    \Else
        \State $t\gets t_{\text{next}}$
    \EndIf
\EndWhile
\State \Return $\mathbf{B}$
\end{algorithmic}
\end{algorithm}

\section{Ablation Study on Scenarios}
\label{ablation_appx}
\begin{figure*}[t]
    \centering
    \begin{subfigure}[b]{1\linewidth}
        \centering
        \includegraphics[width=\linewidth]{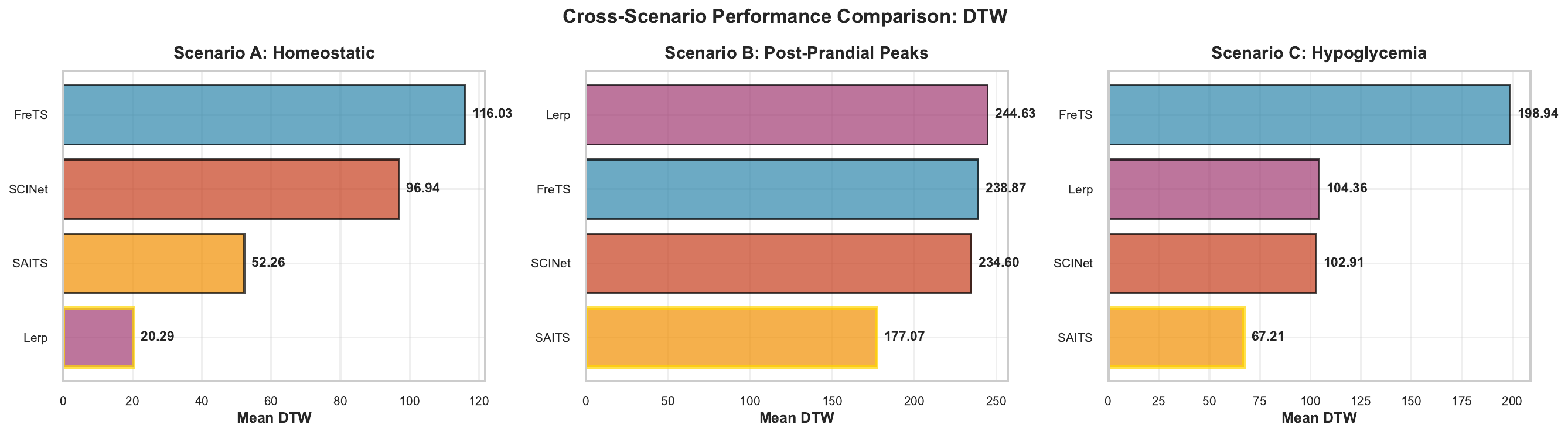}
        \caption{DTW comparison}
        \label{fig:cross_dtw}
    \end{subfigure}\hfill
    \begin{subfigure}[b]{1\linewidth}
        \centering
        \includegraphics[width=\linewidth]{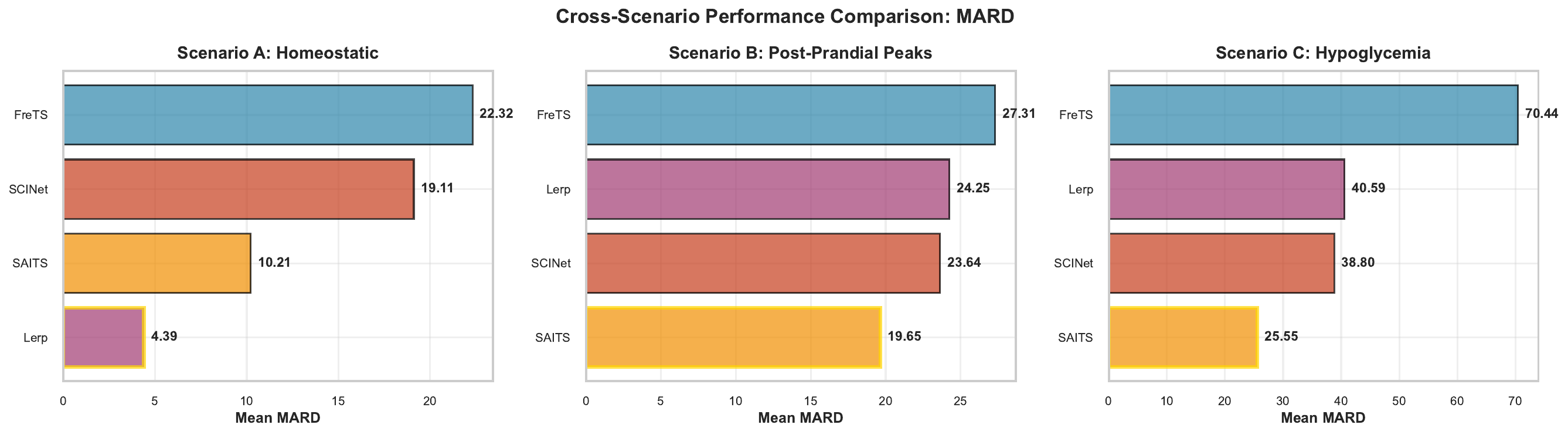}
        \caption{MARD comparison}
        \label{fig:cross_mard}
    \end{subfigure}\hfill
    \begin{subfigure}[b]{1\linewidth}
        \centering
        \includegraphics[width=\linewidth]{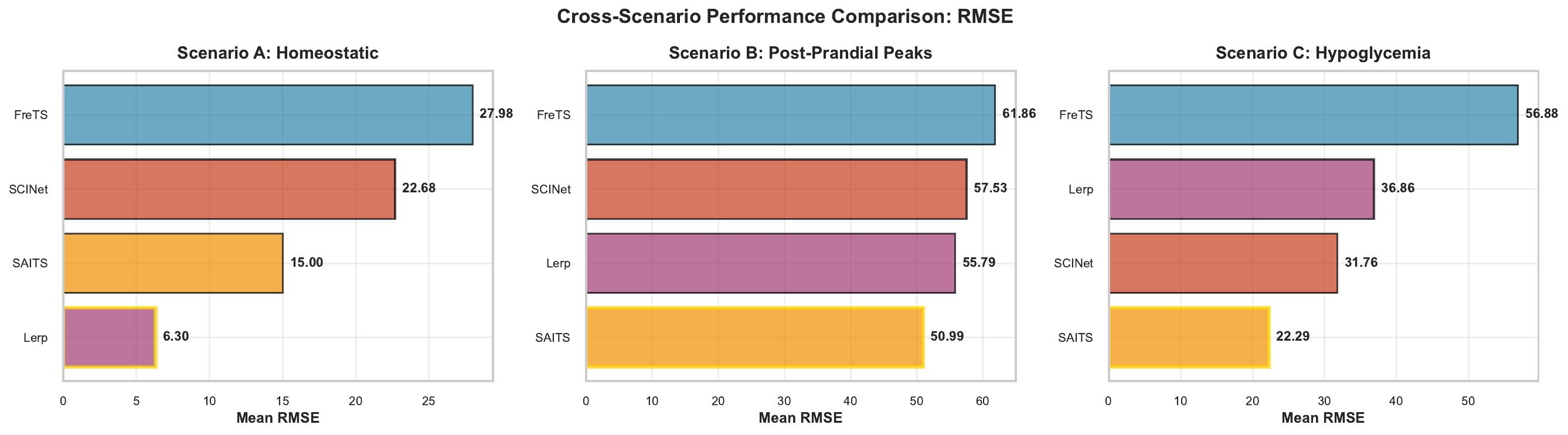}
        \caption{RMSE comparison}
        \label{fig:cross_rmse}
    \end{subfigure}
    \caption{\textbf{Cross-scenario performance comparison. Each panel displays the average performance of the models (Lerp, FreTS, SCINet, SAITS) in terms of (a) DTW, (b) MARD, and (c) RMSE. Lower values indicate better performance.}}
    \label{fig:ablation_cross_scenario}
\end{figure*}

\begin{figure*}[t]
    \centering
    \begin{subfigure}[b]{0.95\linewidth}
        \centering
        \includegraphics[width=\linewidth]{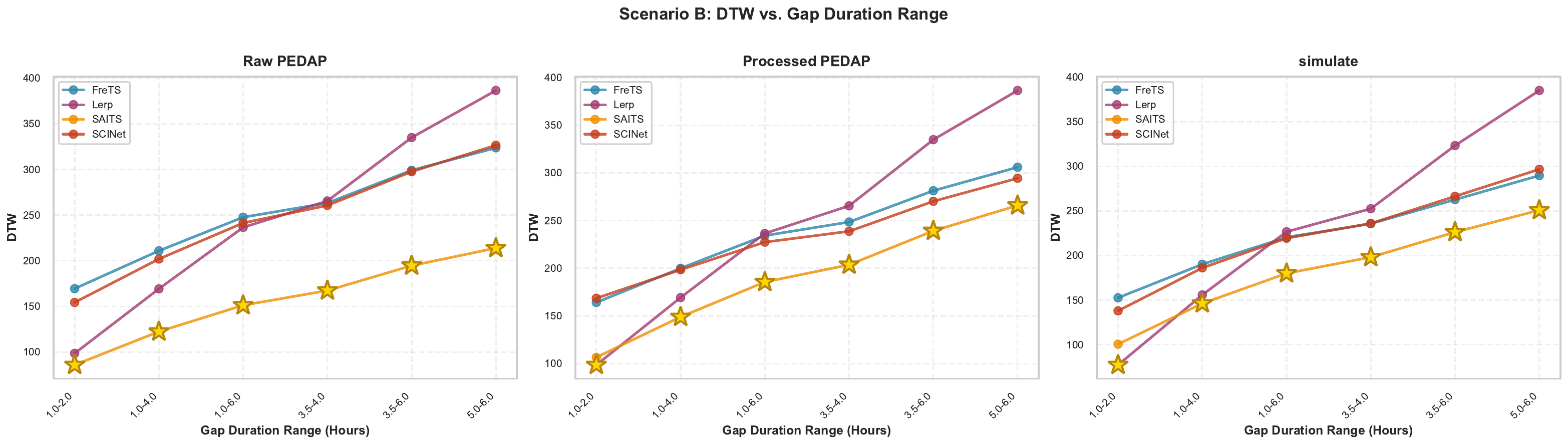}
        \caption{DTW vs. gap duration range}
        \label{fig:scenario_B_dtw_gap}
    \end{subfigure}
    
    \vspace{0.3cm}
    
    \begin{subfigure}[b]{0.95\linewidth}
        \centering
        \includegraphics[width=\linewidth]{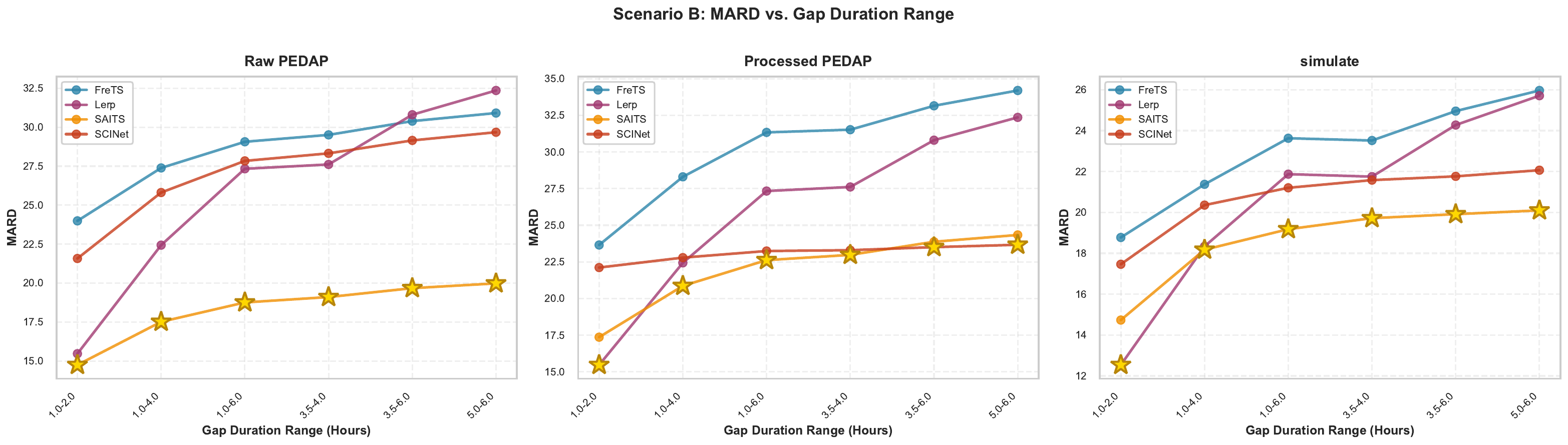}
        \caption{MARD vs. gap duration range}
        \label{fig:scenario_B_mard_gap}
    \end{subfigure}
    
    \vspace{0.3cm}
    
    \begin{subfigure}[b]{0.95\linewidth}
        \centering
        \includegraphics[width=\linewidth]{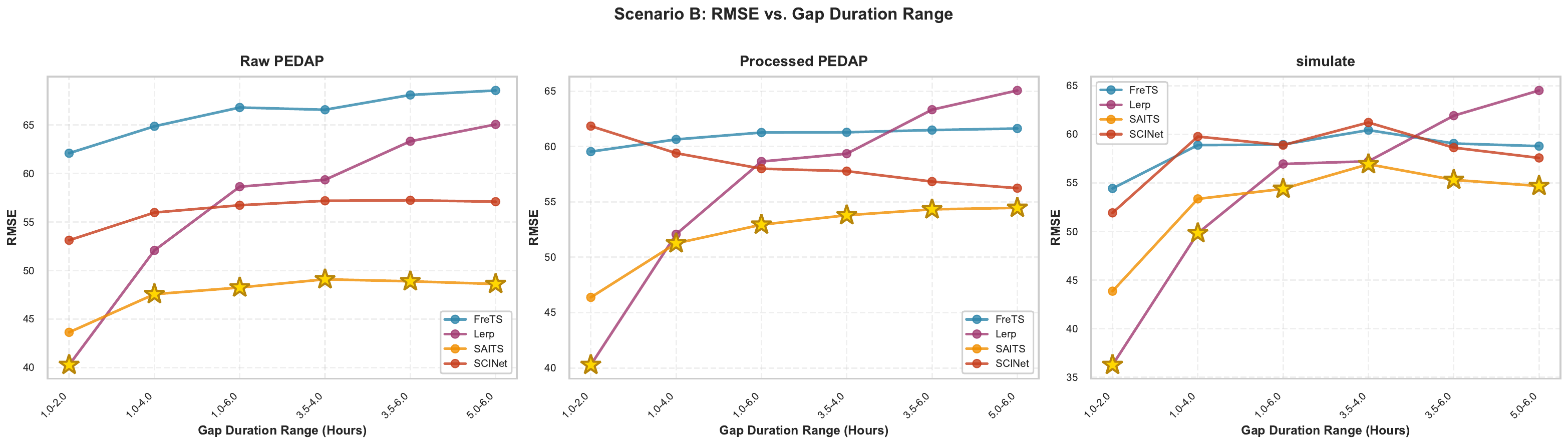}
        \caption{RMSE vs. gap duration range}
        \label{fig:scenario_B_rmse_gap}
    \end{subfigure}
    \caption{\textbf{Scenario B: Impact of gap duration on post-prandial peak reconstruction.} Performance comparison across three datasets with masked meal peaks of varying durations. Gold stars denote the best-performing model at each duration.}
    
    \label{fig:ablation_scenario_B_gap}
\end{figure*}

\begin{figure*}[t]
    \centering
    \begin{subfigure}[b]{0.95\linewidth}
        \centering
        \includegraphics[width=\linewidth]{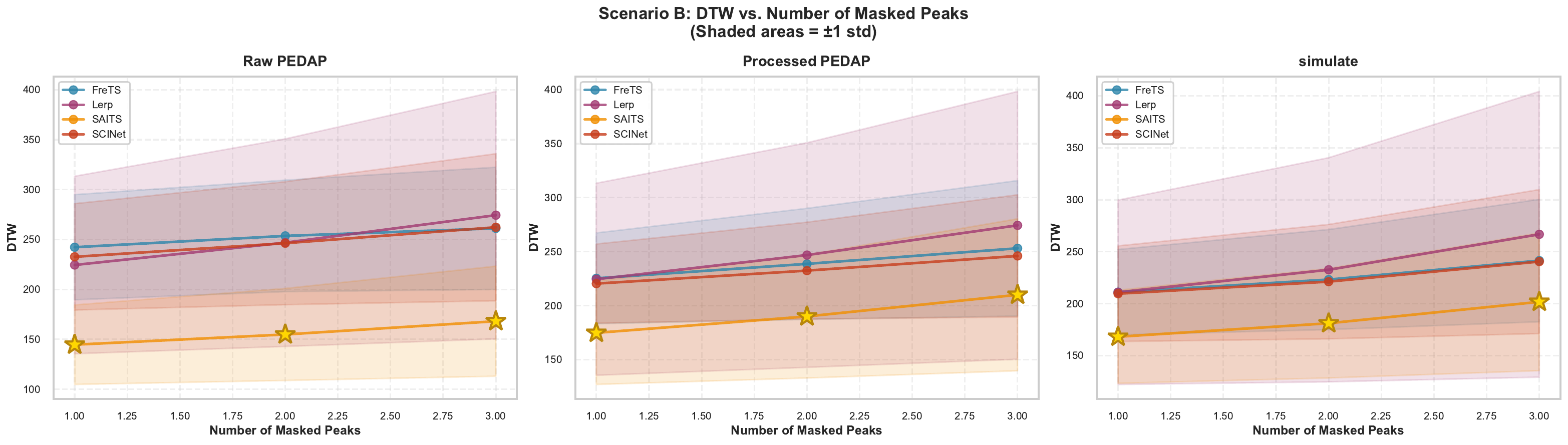}
        \caption{DTW vs. number of masked peaks}
        \label{fig:scenario_B_dtw_meals}
    \end{subfigure}
    
    \vspace{0.3cm}
    
    \begin{subfigure}[b]{0.95\linewidth}
        \centering
        \includegraphics[width=\linewidth]{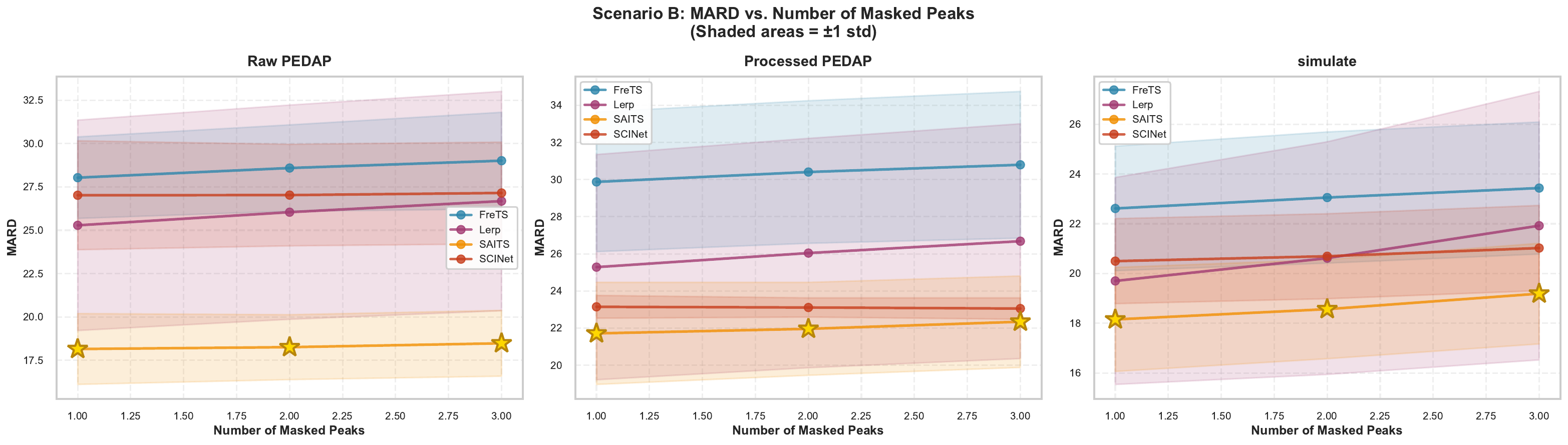}
        \caption{MARD vs. number of masked peaks }
        \label{fig:scenario_B_mard_meals}
    \end{subfigure}
    
    \vspace{0.3cm}
    
    \begin{subfigure}[b]{0.95\linewidth}
        \centering
        \includegraphics[width=\linewidth]{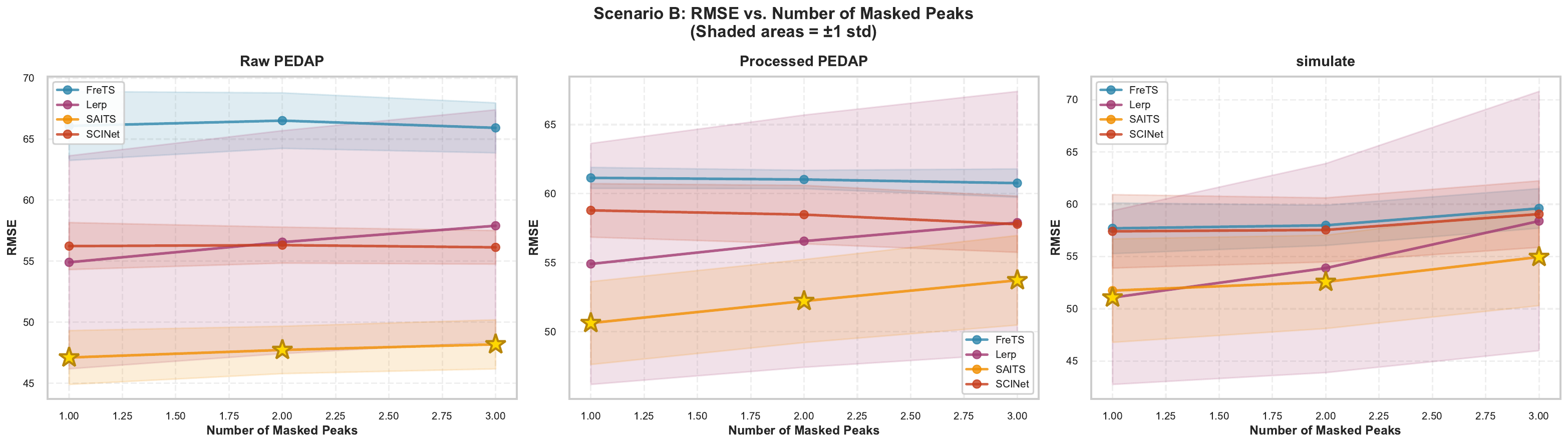}
        \caption{RMSE vs. number of masked peaks}
        \label{fig:scenario_B_rmse_meals}
    \end{subfigure}
    \caption{\textbf{Scenario B: Impact of post-prandial severity on model performance.} Performance comparison across three datasets with an increasing number of masked meal peaks across different gap durations. Shaded regions denote $\pm1$ standard deviation; gold stars mark the best-performing model.}
    \label{fig:ablation_scenario_B_meals}
\end{figure*}

Figures \ref{fig:ablation_cross_scenario}--\ref{fig:ablation_scenario_C} present ablation studies comparing the top-performing methods (Lerp, FreTS, SCINet, and SAITS) across all datasets. These comparisons evaluate the impact of masking length and missing ratios in Scenario A, while Scenarios B and C focus specifically on masking length and the number of physiological peaks.

Figure \ref{fig:ablation_cross_scenario} aggregates the results detailed in Figures \ref{fig:ablation_scenario_A}--\ref{fig:ablation_scenario_C}. As shown, linear interpolation (Lerp) outperforms deep learning methods in the homeostatic regions of Scenario A across all evaluated metrics. Specifically, Lerp demonstrates superior performance in morphology (DTW: 20.29 vs. 52.26 for SAITS), clinical safety (MARD: 4.39\% vs. 10.21\% for SAITS), and point-wise error (RMSE: 6.30 vs. 15.00 mg/dL for SAITS), establishing it as the preferred method for stable glycemic regimes.
In contrast, deep learning methods (SAITS, SCINet) significantly outperform linear interpolation in volatile regimes, particularly when reconstructing post-prandial peaks (Scenario B) and hypoglycemic events (Scenario C). In these physiologically complex scenarios, deep learning approaches achieve superior performance across all morphological, safety, and point-wise metrics.

These ablation studies support our model-selection framework: linear interpolation is preferred for homeostatic regions, while deep learning method (SAITS) is superior during physiological excursions—including post-prandial peaks and hypoglycemic events that occur during model temporal control reset—in terms of both morphology and clinical safety.

\subsection{Scenario A}
Figure \ref{fig:ablation_scenario_A} presents the performance of models in homeostatic regions (Scenario A) across varying missing window lengths (10 to 60 minutes) and masking ratios (10\% to 50\%). Notably, results for the 50\% masking ratio are reported exclusively for the simulation dataset, as other datasets lacked sufficient regions satisfying the Scenario A criteria at this level.

In all evaluated cases, linear interpolation outperforms deep learning models. Furthermore, it demonstrates superior stability as both the missing window length and masking ratio increase.

\subsection{Scenario B}
Figure \ref{fig:ablation_scenario_B_gap} illustrates the average RMSE, MARD, and DTW for each model across different numbers of peaks present, stratified by missing data duration.
Linear interpolation performs better on average only for small gaps of 1-2 hours, suggesting that within this limited missingness range, physiological peaks requiring deep learning reconstruction are typically absent. Beyond this range, deep learning models demonstrate superior performance. Figure \ref{fig:ablation_scenario_B_meals} further presents the average metrics stratified by the number of peaks across all gap durations.
The results demonstrate that deep learning models (SAITS, SCINet) consistently outperform linear interpolation across all three metrics—RMSE, MARD, and DTW—highlighting their superior capability to reconstruct post-prandial peaks.

\subsection{Scenario C}
\begin{figure*}[t]
    \centering   
    \begin{subfigure}[b]{0.32\linewidth}
        \centering
        \includegraphics[width=\linewidth]{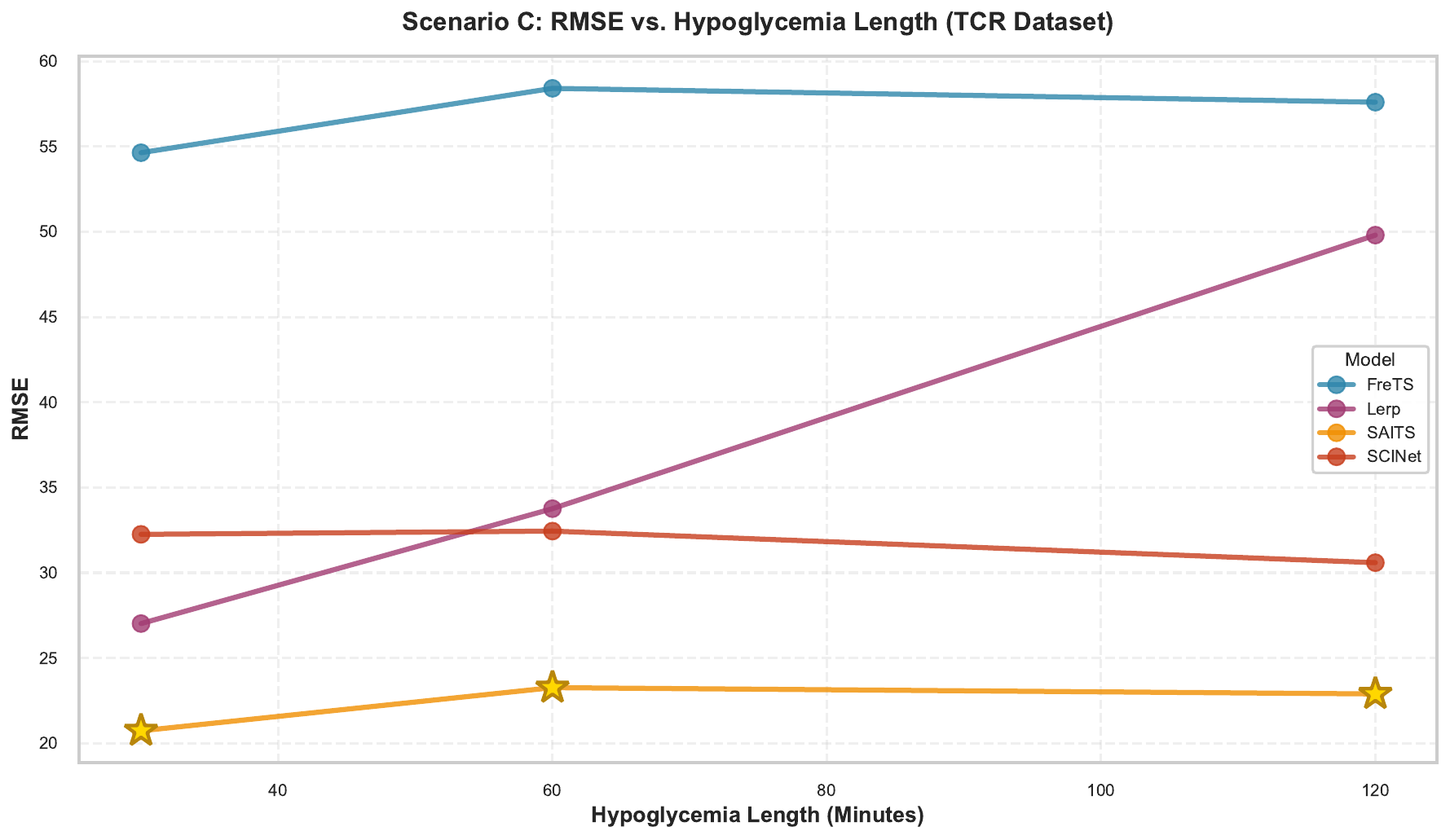}
        \caption{RMSE performance}
        \label{fig:scenario_C_rmse}
    \end{subfigure}
    \begin{subfigure}[b]{0.32\linewidth}
        \centering
        \includegraphics[width=\linewidth]{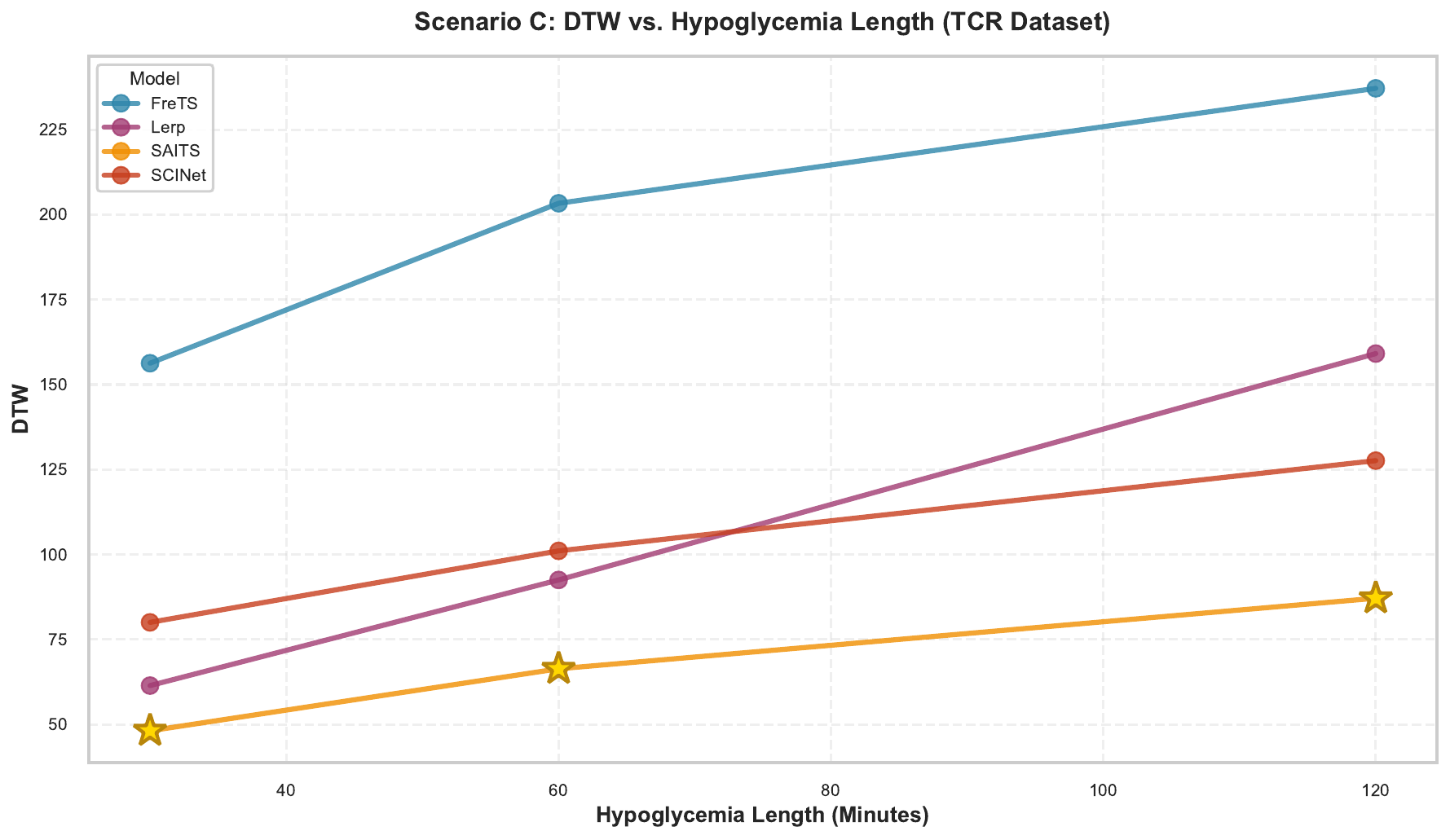}
        \caption{DTW performance}
        \label{fig:scenario_C_dtw}
    \end{subfigure}\hfill
    \begin{subfigure}[b]{0.32\linewidth}
        \centering
        \includegraphics[width=\linewidth]{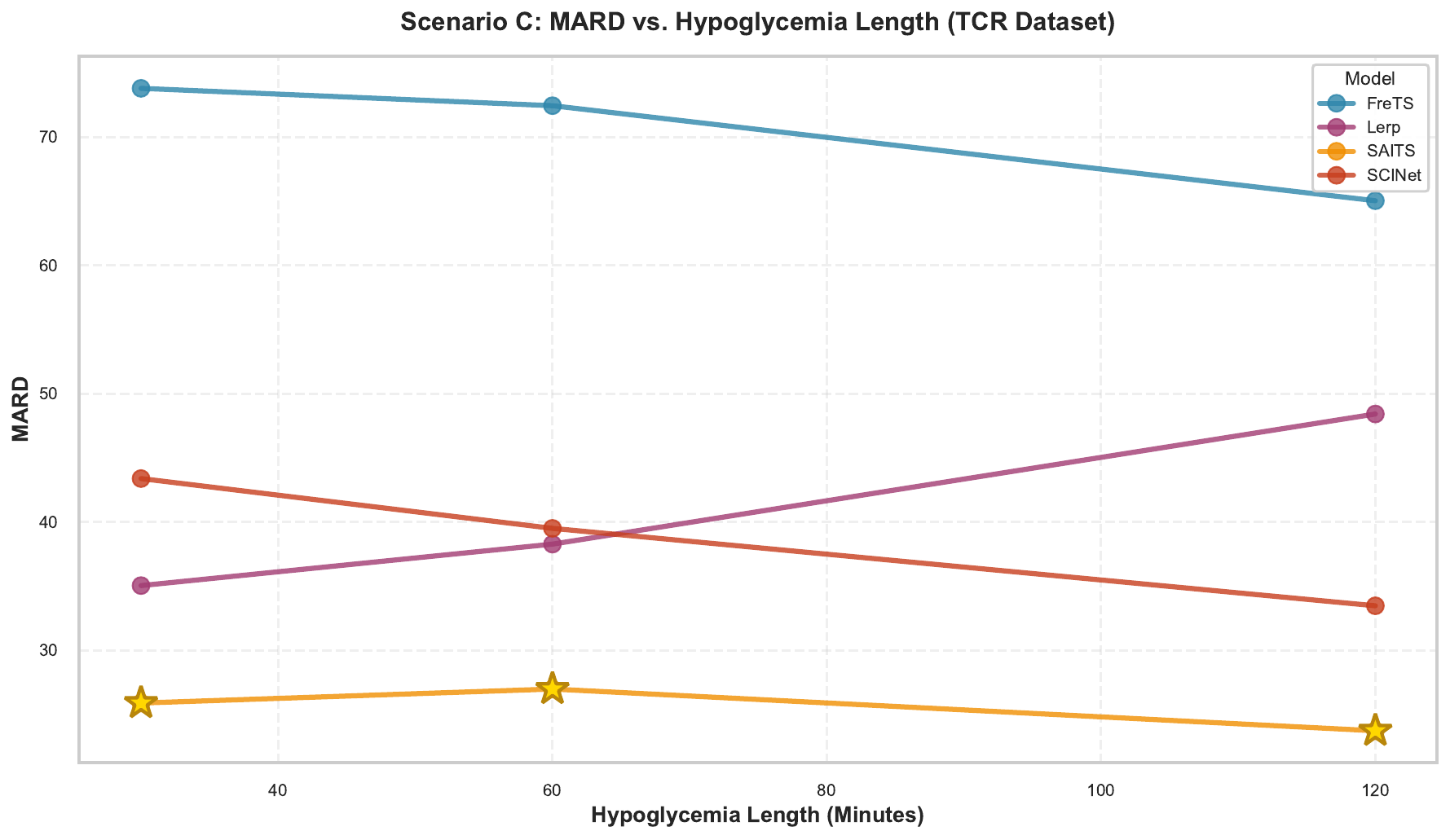}
        \caption{MARD performance}
        \label{fig:scenario_C_mard}
    \end{subfigure}\hfill
    \caption{\textbf{Scenario C: Hypoglycemia stress test (TCR dataset).} Impact of hypoglycemic event duration (30--120 min) on model performance. Gold stars denote the best-performing model at each duration.}
    \label{fig:ablation_scenario_C}
\end{figure*}

Figure \ref{fig:ablation_scenario_C} illustrates the results of experiments conducted with hypoglycemia window lengths ranging from 30 to 120 minutes. Across all evaluated durations, the deep learning model (SAITS) consistently outperformed linear interpolation.

\begin{figure*}[t]
    \centering
    \begin{subfigure}[b]{0.95\linewidth}
        \centering
        \includegraphics[width=\linewidth]{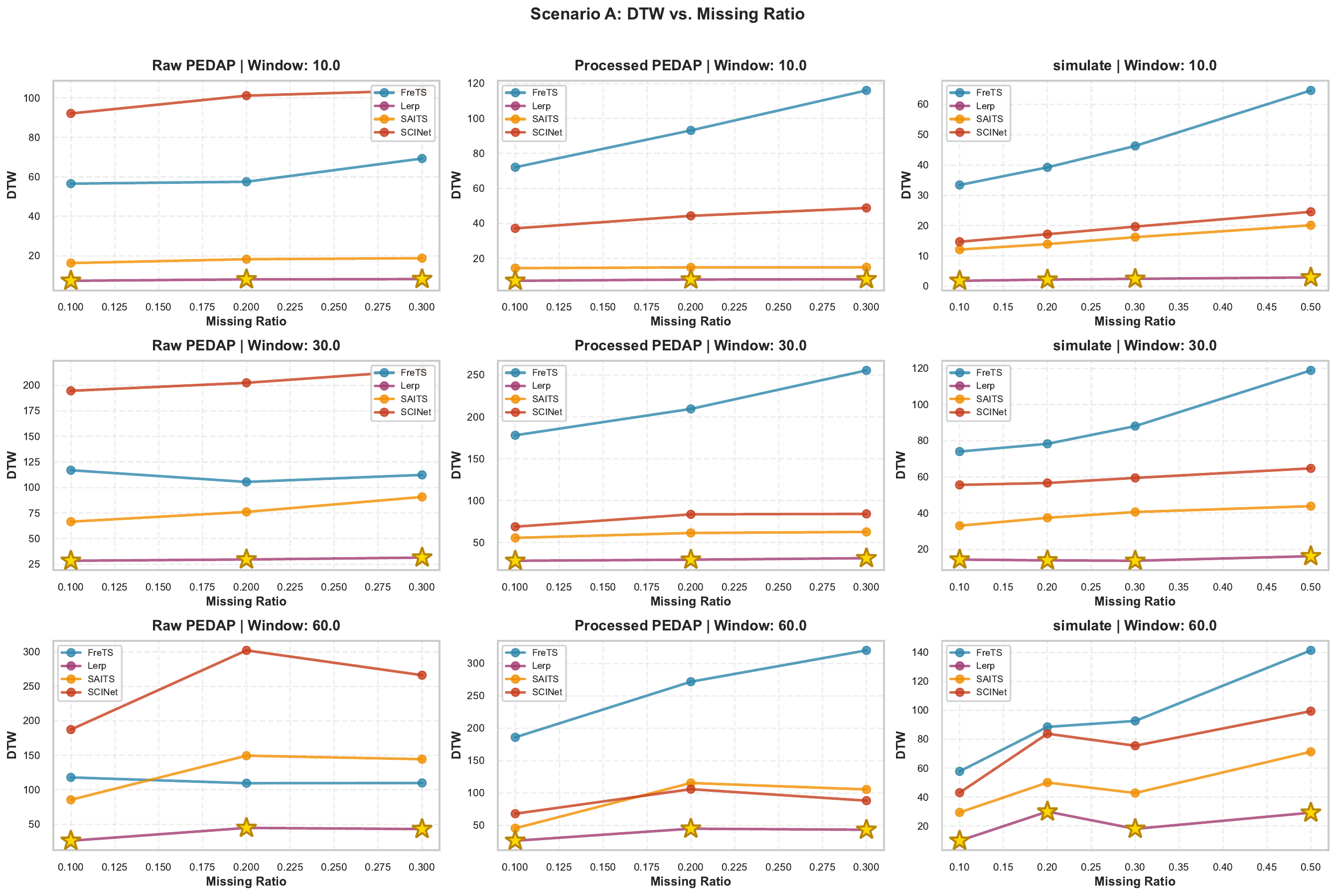}
        \caption{DTW performance across missing ratios}
        \label{fig:scenario_A_dtw}
    \end{subfigure}
    
    \vspace{0.3cm}
    
    \begin{subfigure}[b]{0.95\linewidth}
        \centering
        \includegraphics[width=\linewidth]{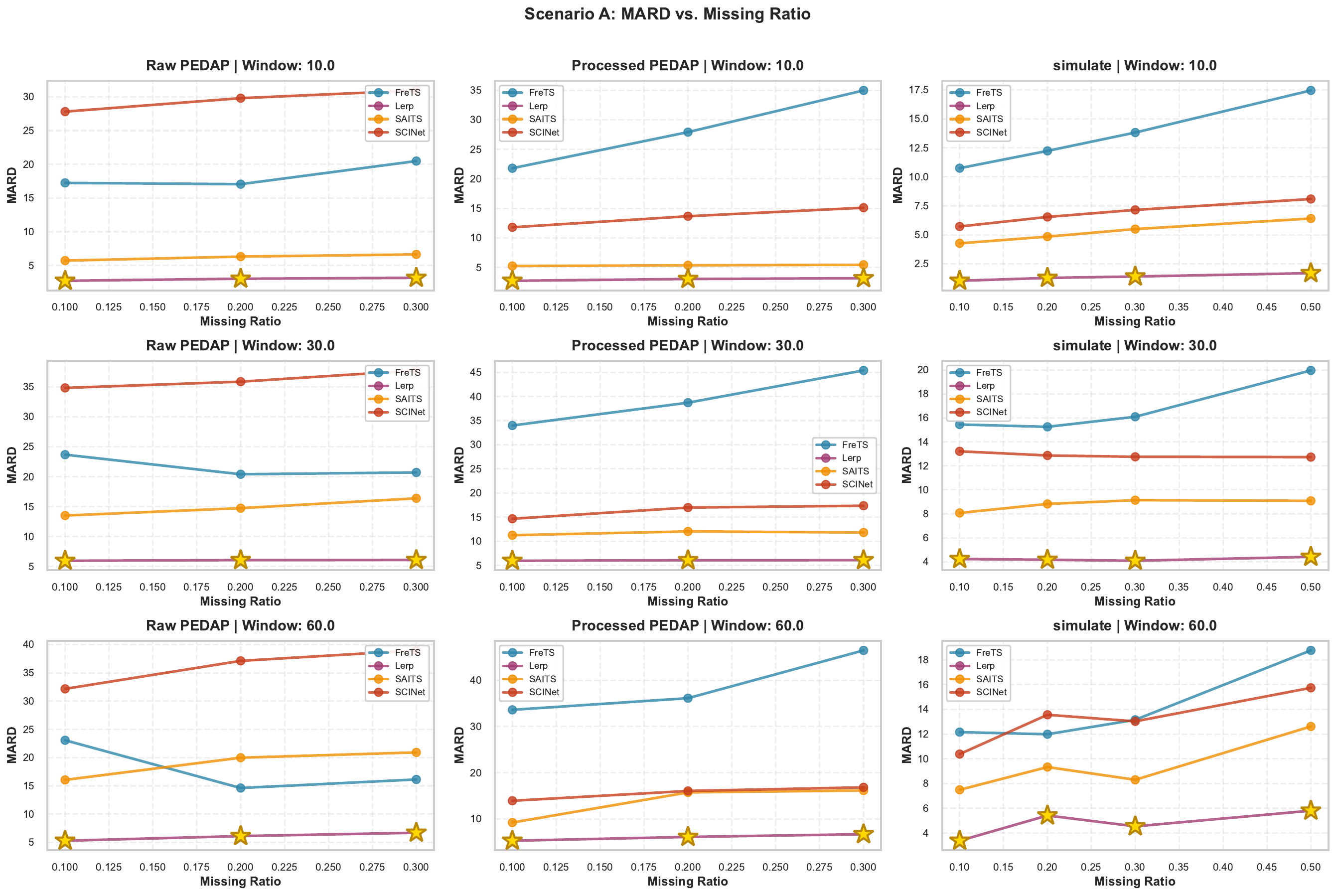}
        \caption{MARD performance across missing ratios}
        \label{fig:scenario_A_mard}
    \end{subfigure}
\end{figure*}

\begin{figure*}[t]
    \ContinuedFloat
    \centering
    \begin{subfigure}[b]{0.95\linewidth}
        \centering
        \includegraphics[width=\linewidth]{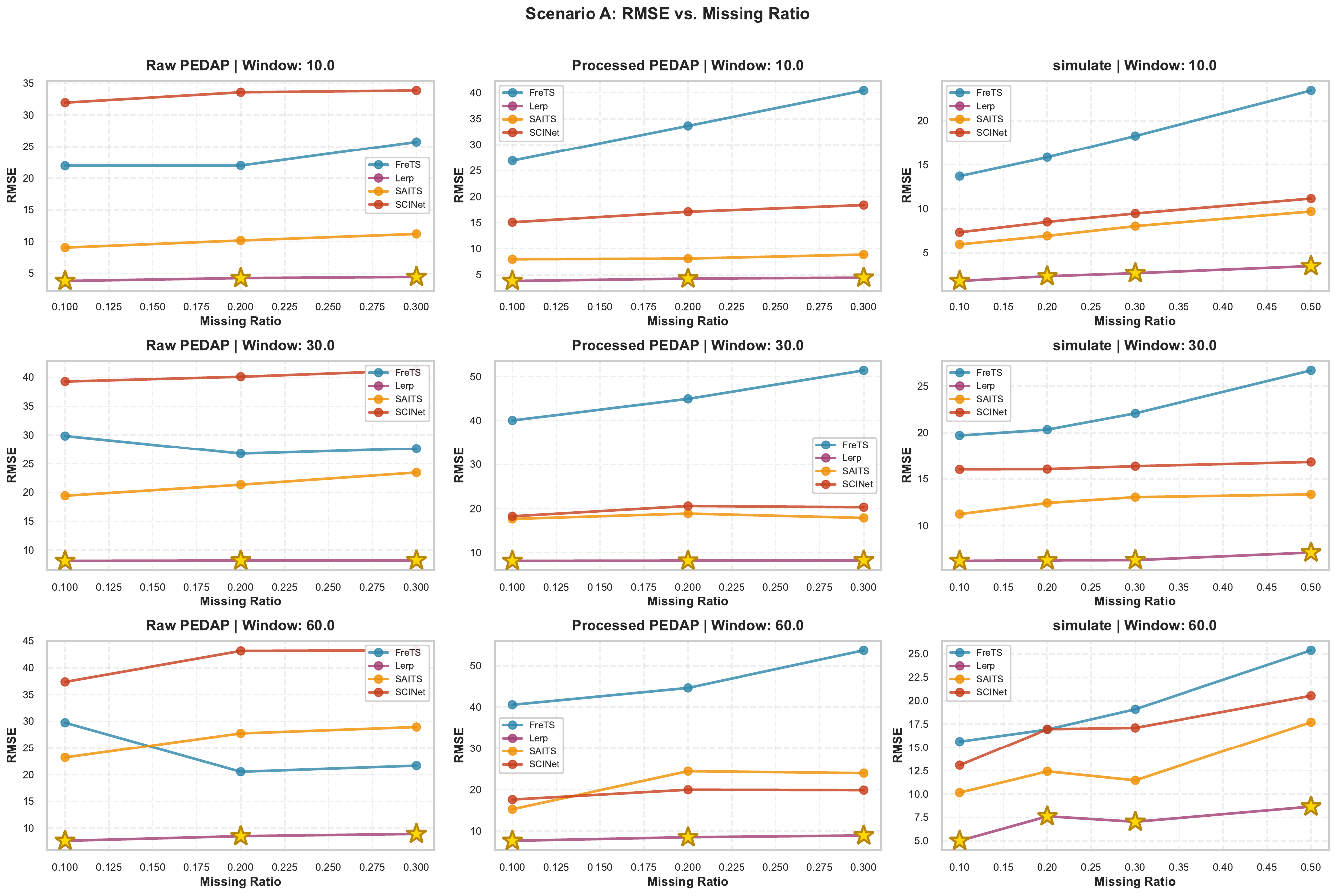}
        \caption{RMSE performance across missing ratios}
        \label{fig:scenario_A_rmse}
    \end{subfigure}
    
    \caption{\textbf{Scenario A: Impact of missing data ratio and gap duration on model performance.} Results are presented for three datasets (Raw PEDAP, Processed PEDAP, Simulation). Gold stars indicate the best-performing model for each condition.}
    \label{fig:ablation_scenario_A}
\end{figure*}

\section{Complete Results for Scenario A}
\label{apdx_full_results}
Table~\ref{tab:results_A_all_datasets} presents the complete experimental results for Scenario A across all datasets and conditions.

\begin{table*}[b]
\centering
\caption{Performance in Scenario A. \textbf{Bold} indicates best; \underline{underline} indicates second best.}
\label{tab:results_A_all_datasets}
\small
\setlength{\tabcolsep}{3pt}
\begin{tabular}{c l ccccc ccccc ccccc}
\toprule
\textbf{Ratio} & \textbf{Model} & \multicolumn{5}{c}{\textbf{Processed PEDAP}} & \multicolumn{5}{c}{\textbf{Raw PEDAP}} & \multicolumn{5}{c}{\textbf{Simulation}} \\
\cmidrule(lr){3-7}\cmidrule(lr){8-12}\cmidrule(lr){13-17}
 &  & \textbf{RMSE} & \textbf{Bias} & \textbf{Emp\_SE} & \textbf{MARD} & \textbf{DTW} & \textbf{RMSE} & \textbf{Bias} & \textbf{Emp\_SE} & \textbf{MARD} & \textbf{DTW} & \textbf{RMSE} & \textbf{Bias} & \textbf{Emp\_SE} & \textbf{MARD} & \textbf{DTW} \\
\midrule
 & FreTS & 23.18 & 15.59 & 17.15 & 18.18 & 48.61 & 19.15 & 10.71 & 15.88 & 14.68 & 39.26 & 12.24 & 6.76 & 10.20 & 9.35 & 23.65 \\
 & SCINet & 13.46 & 7.65 & 11.07 & 10.29 & 26.62 & 29.91 & 26.82 & 13.23 & 25.66 & 67.06 & 5.43 & 3.07 & \underline{4.48} & 4.12 & 8.92 \\
 & TimeMixer & 65.56 & -63.51 & 16.27 & 56.83 & 156.67 & 71.25 & -69.35 & 16.34 & 62.25 & 170.99 & 67.20 & -65.44 & 15.29 & 58.91 & 157.89 \\
 & TSLANet & 59.98 & -44.12 & 40.63 & 48.66 & 143.04 & 59.12 & -47.24 & 35.55 & 47.92 & 140.35 & 61.42 & -49.05 & 36.98 & 50.64 & 143.34 \\
 & TEFN & 36.54 & 28.63 & 22.71 & 29.85 & 74.00 & 36.08 & 28.17 & 22.54 & 29.44 & 73.00 & 30.89 & 22.31 & 21.36 & 25.06 & 58.60 \\
0.1 & TOTEM & 52.01 & -48.14 & 19.69 & 43.03 & 116.31 & 21.73 & -14.43 & 16.25 & 13.58 & 32.01 & 32.21 & -27.20 & 17.25 & 25.11 & 63.11 \\
 & GPT4TS & 30.63 & 23.36 & 19.81 & 23.88 & 60.43 & 35.45 & 28.77 & 20.71 & 28.32 & 72.60 & 33.78 & 27.07 & 20.21 & 27.68 & 65.99 \\
 & SAITS & 5.32 & 1.38 & 5.14 & 3.66 & \underline{8.45} & 5.91 & 1.71 & 5.66 & 3.64 & \underline{8.26} & 5.27 & -1.09 & 5.16 & 3.84 & 9.37 \\
 & Mean & 49.36 & 46.59 & 16.31 & 45.85 & 114.87 & 49.36 & 46.59 & 16.31 & 45.85 & 114.87 & 41.00 & 37.59 & 16.35 & 37.56 & 90.61 \\
 & Median & 38.32 & 34.68 & 16.31 & 34.74 & 85.60 & 38.32 & 34.68 & 16.31 & 34.74 & 85.60 & 28.45 & 23.28 & 16.35 & 24.52 & 57.38 \\
 & LOCF & \underline{4.78} & \underline{0.76} & \underline{4.72} & \underline{3.45} & 9.91 & \underline{4.78} & \underline{0.76} & \underline{4.72} & \underline{3.45} & 9.91 & \underline{4.96} & \underline{0.40} & 4.94 & \underline{3.23} & \underline{8.83} \\
 & Lerp & \textbf{2.47} & \textbf{0.37} & \textbf{2.44} & \textbf{1.79} & \textbf{4.02} & \textbf{2.47} & \textbf{0.37} & \textbf{2.44} & \textbf{1.79} & \textbf{4.02} & \textbf{0.99} & \textbf{0.17} & \textbf{0.97} & \textbf{0.53} & \textbf{0.89} \\
\midrule
 & FreTS & 29.35 & 22.77 & 18.52 & 23.84 & 64.09 & 19.43 & 9.07 & 17.18 & 14.85 & 41.02 & 14.71 & 9.13 & 11.53 & 11.22 & 29.19 \\
 & SCINet & 15.48 & 10.67 & 11.21 & 12.03 & 32.01 & 31.96 & 29.27 & 12.82 & 27.87 & 75.22 & 6.44 & 3.82 & \underline{5.18} & 4.82 & 10.62 \\
 & TimeMixer & 66.56 & -64.74 & 15.43 & 58.22 & 166.82 & 70.84 & -69.13 & 15.46 & 62.30 & 178.04 & 68.60 & -66.89 & 15.22 & 59.93 & 168.17 \\
 & TSLANet & 59.63 & -44.01 & 40.23 & 48.67 & 148.60 & 58.66 & -46.72 & 35.47 & 47.82 & 145.59 & 61.82 & -49.67 & 36.80 & 50.74 & 150.42 \\
 & TEFN & 29.74 & 22.01 & 20.01 & 24.06 & 62.23 & 29.38 & 21.64 & 19.87 & 23.75 & 61.42 & 31.50 & 23.54 & 20.94 & 25.58 & 62.65 \\
0.2 & TOTEM & 55.55 & -52.05 & 19.41 & 46.70 & 130.72 & 26.93 & -19.30 & 18.79 & 17.81 & 43.70 & 39.34 & -33.77 & 20.19 & 30.88 & 80.27 \\
 & GPT4TS & 25.83 & 18.92 & 17.58 & 20.04 & 51.46 & 30.31 & 24.20 & 18.24 & 24.17 & 63.01 & 35.05 & 28.91 & 19.82 & 29.02 & 72.33 \\
 & SAITS & 5.65 & 1.20 & 5.52 & 3.76 & \underline{8.79} & 7.06 & 2.27 & 6.68 & 4.20 & \underline{9.51} & 5.83 & \textbf{0.20} & 5.83 & 4.16 & 10.24 \\
 & Mean & 48.24 & 45.70 & 15.45 & 44.76 & 117.83 & 48.24 & 45.70 & 15.45 & 44.76 & 117.83 & 43.43 & 40.32 & 16.13 & 39.78 & 101.26 \\
 & Median & 36.83 & 33.43 & 15.45 & 33.32 & 86.36 & 36.83 & 33.43 & 15.45 & 33.32 & 86.36 & 32.15 & 27.81 & 16.13 & 28.18 & 70.06 \\
 & LOCF & \underline{5.28} & \underline{0.68} & \underline{5.23} & \underline{3.73} & 11.00 & \underline{5.28} & \underline{0.68} & \underline{5.23} & \underline{3.73} & 11.00 & \underline{5.62} & 0.54 & 5.59 & \underline{3.53} & \underline{9.85} \\
 & Lerp & \textbf{2.74} & \textbf{0.28} & \textbf{2.72} & \textbf{1.94} & \textbf{4.36} & \textbf{2.74} & \textbf{0.28} & \textbf{2.72} & \textbf{1.94} & \textbf{4.36} & \textbf{1.47} & \underline{0.21} & \textbf{1.46} & \textbf{0.70} & \textbf{1.08} \\
\midrule
 & FreTS & 37.31 & 30.97 & 20.81 & 31.71 & 85.70 & 23.08 & 12.56 & 19.37 & 17.93 & 50.09 & 17.49 & 11.53 & 13.15 & 13.26 & 35.73 \\
 & SCINet & 17.36 & 12.48 & 12.07 & 13.76 & 36.80 & 32.16 & 29.50 & 12.80 & 28.79 & 78.28 & 7.59 & 4.69 & \underline{5.97} & 5.57 & 12.50 \\
 & TimeMixer & 65.43 & -63.67 & 15.06 & 58.48 & 170.17 & 68.62 & -66.93 & 15.12 & 61.57 & 178.74 & 69.89 & -68.24 & 15.13 & 60.69 & 177.89 \\
 & TSLANet & 58.12 & -41.90 & 40.28 & 48.38 & 149.83 & 57.36 & -45.13 & 35.41 & 47.69 & 146.81 & 62.42 & -50.52 & 36.66 & 50.94 & 157.69 \\
 & TEFN & 25.41 & 18.72 & 17.19 & 21.09 & 55.16 & 25.08 & 18.39 & 17.06 & 20.79 & 54.41 & 29.74 & 22.62 & 19.32 & 24.16 & 62.05 \\
0.3 & TOTEM & 56.37 & -53.06 & 19.03 & 48.56 & 137.75 & 30.47 & -23.09 & 19.88 & 21.42 & 53.68 & 44.68 & -38.76 & 22.22 & 34.92 & 93.79 \\
 & GPT4TS & 22.51 & 16.33 & 15.50 & 17.93 & 44.99 & 27.07 & 21.98 & 15.79 & 22.37 & 57.35 & 34.23 & 28.74 & 18.59 & 28.44 & 74.01 \\
 & SAITS & 6.10 & 1.23 & 5.97 & \underline{3.95} & \underline{9.04} & 8.63 & 2.70 & 8.20 & 4.89 & \underline{11.13} & 6.82 & 1.43 & 6.67 & 4.72 & 11.66 \\
 & Mean & 49.78 & 47.43 & 15.12 & 47.22 & 127.08 & 49.78 & 47.43 & 15.12 & 47.22 & 127.08 & 42.70 & 39.67 & 15.79 & 38.73 & 103.27 \\
 & Median & 37.82 & 34.67 & 15.12 & 35.09 & 93.14 & 37.82 & 34.67 & 15.12 & 35.09 & 93.14 & 31.80 & 27.60 & 15.79 & 27.63 & 72.02 \\
 & LOCF & \underline{5.65} & \underline{0.70} & \underline{5.61} & 4.05 & 12.11 & \underline{5.65} & \underline{0.70} & \underline{5.61} & \underline{4.05} & 12.11 & \underline{6.09} & \underline{0.78} & 6.04 & \underline{3.70} & \underline{10.48} \\
 & Lerp & \textbf{2.75} & \textbf{0.01} & \textbf{2.75} & \textbf{1.97} & \textbf{4.48} & \textbf{2.75} & \textbf{0.01} & \textbf{2.75} & \textbf{1.97} & \textbf{4.48} & \textbf{1.83} & \textbf{0.24} & \textbf{1.81} & \textbf{0.85} & \textbf{1.23} \\
\bottomrule
\end{tabular}
\end{table*}

\section{Imputation Qualitative Examples}
\label{apdx_c}
\begin{figure*}[htbp]
    \centering
    \begin{subfigure}{0.32\linewidth}\centering \textbf{Raw Pedap}\end{subfigure}%
    \hfill
    \begin{subfigure}{0.32\linewidth}\centering \textbf{Processed Pedap}\end{subfigure}%
    \hfill
    \begin{subfigure}{0.32\linewidth}\centering \textbf{Simulate}\end{subfigure}
    \smallskip
    
    \begin{subfigure}[b]{0.32\linewidth}
        \includegraphics[width=\linewidth]{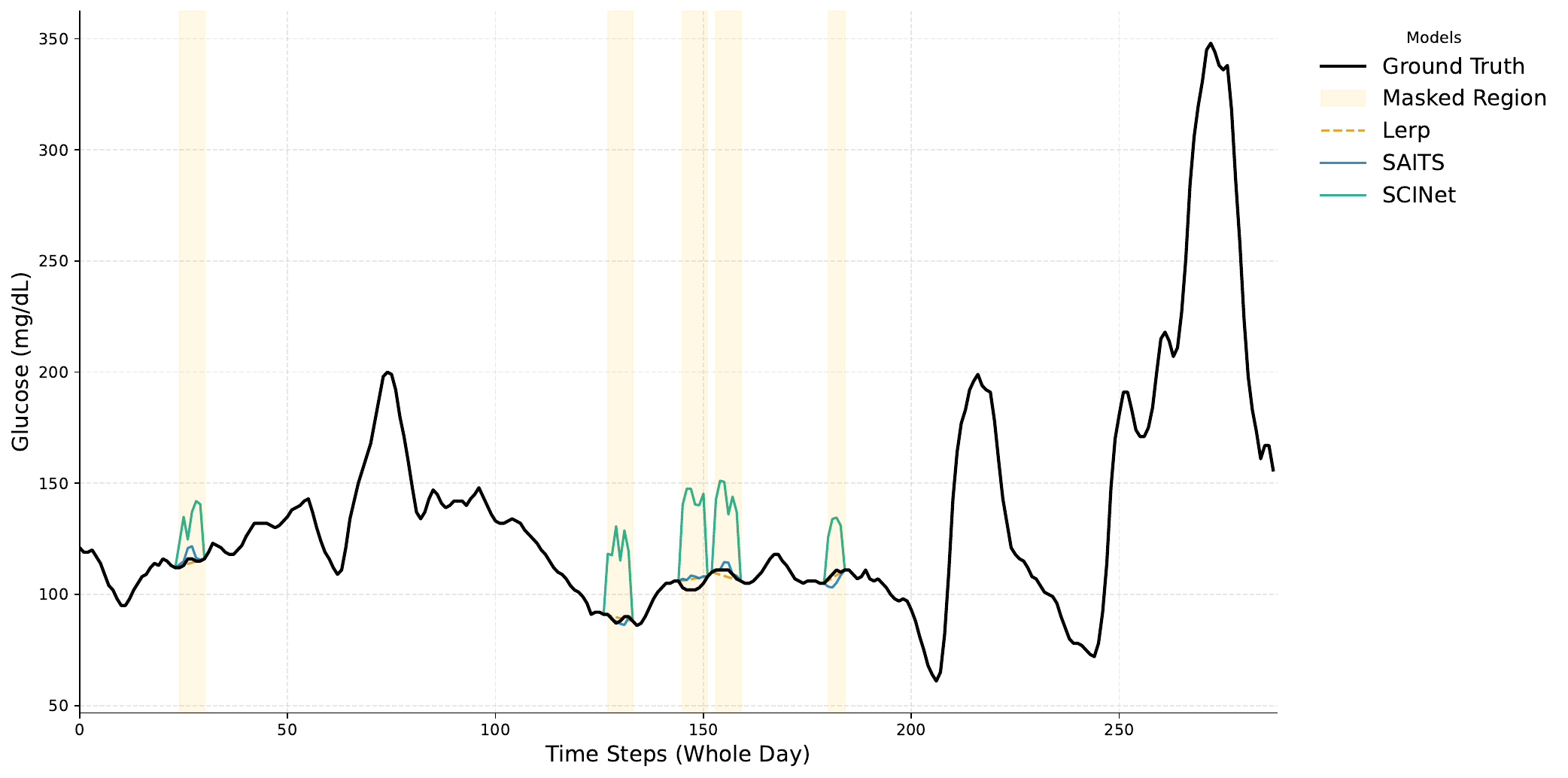}
    \end{subfigure}\hfill
    \begin{subfigure}[b]{0.32\linewidth}
        \includegraphics[width=\linewidth]{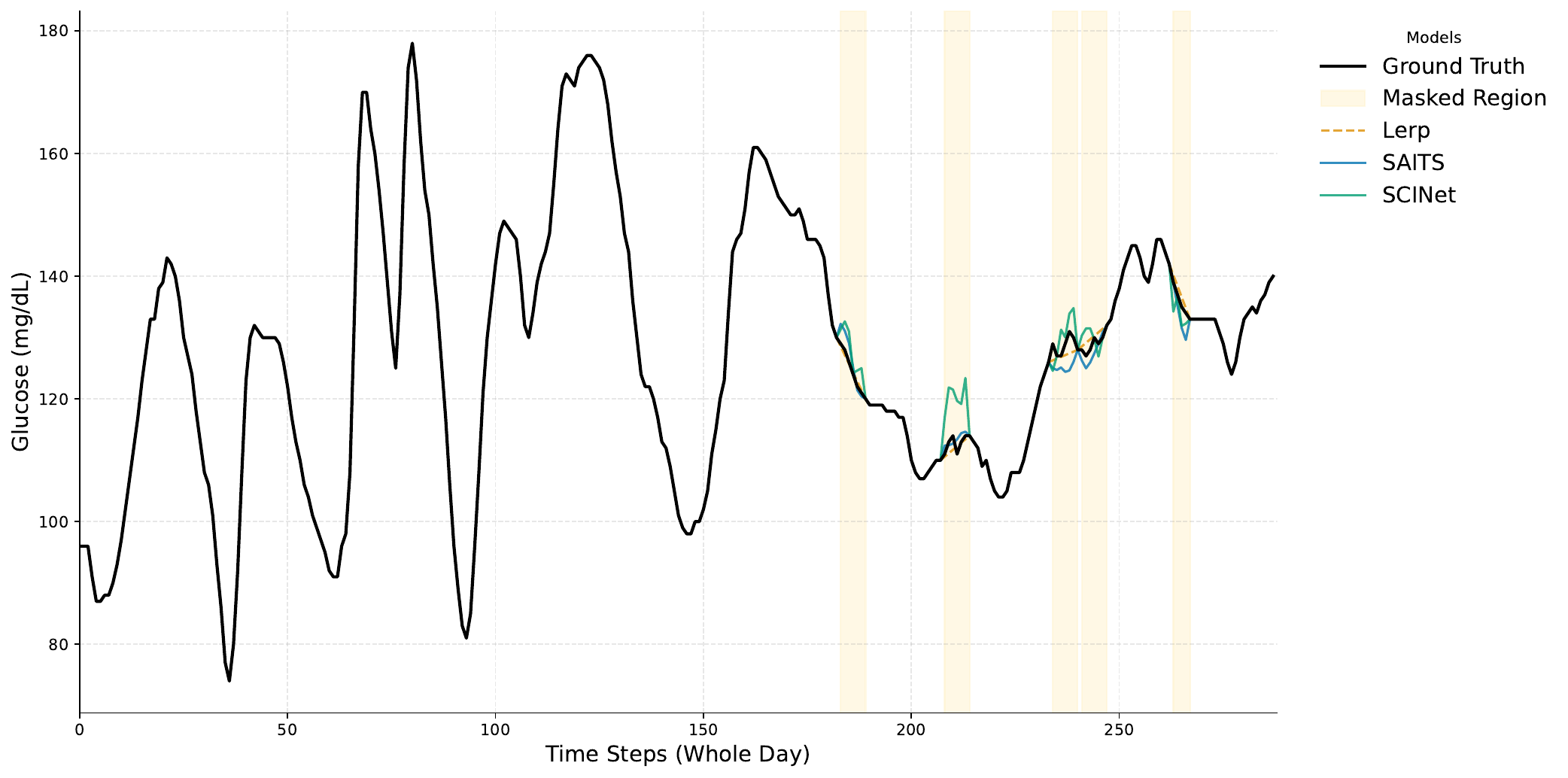}
    \end{subfigure}\hfill
    \begin{subfigure}[b]{0.32\linewidth}
        \includegraphics[width=\linewidth]{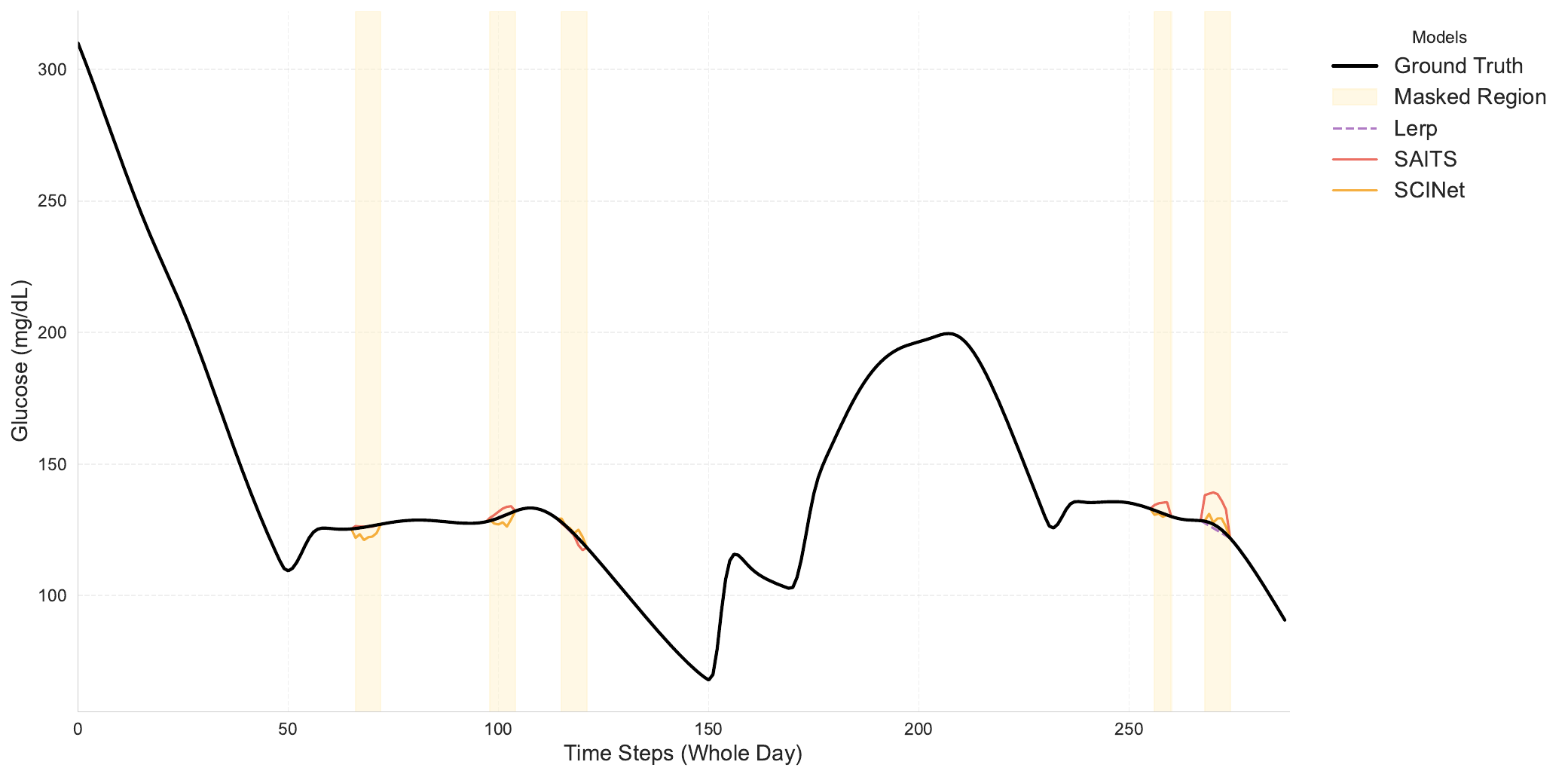}
    \end{subfigure}
    
    \smallskip
    
    \begin{subfigure}[b]{0.32\linewidth}
        \includegraphics[width=\linewidth]{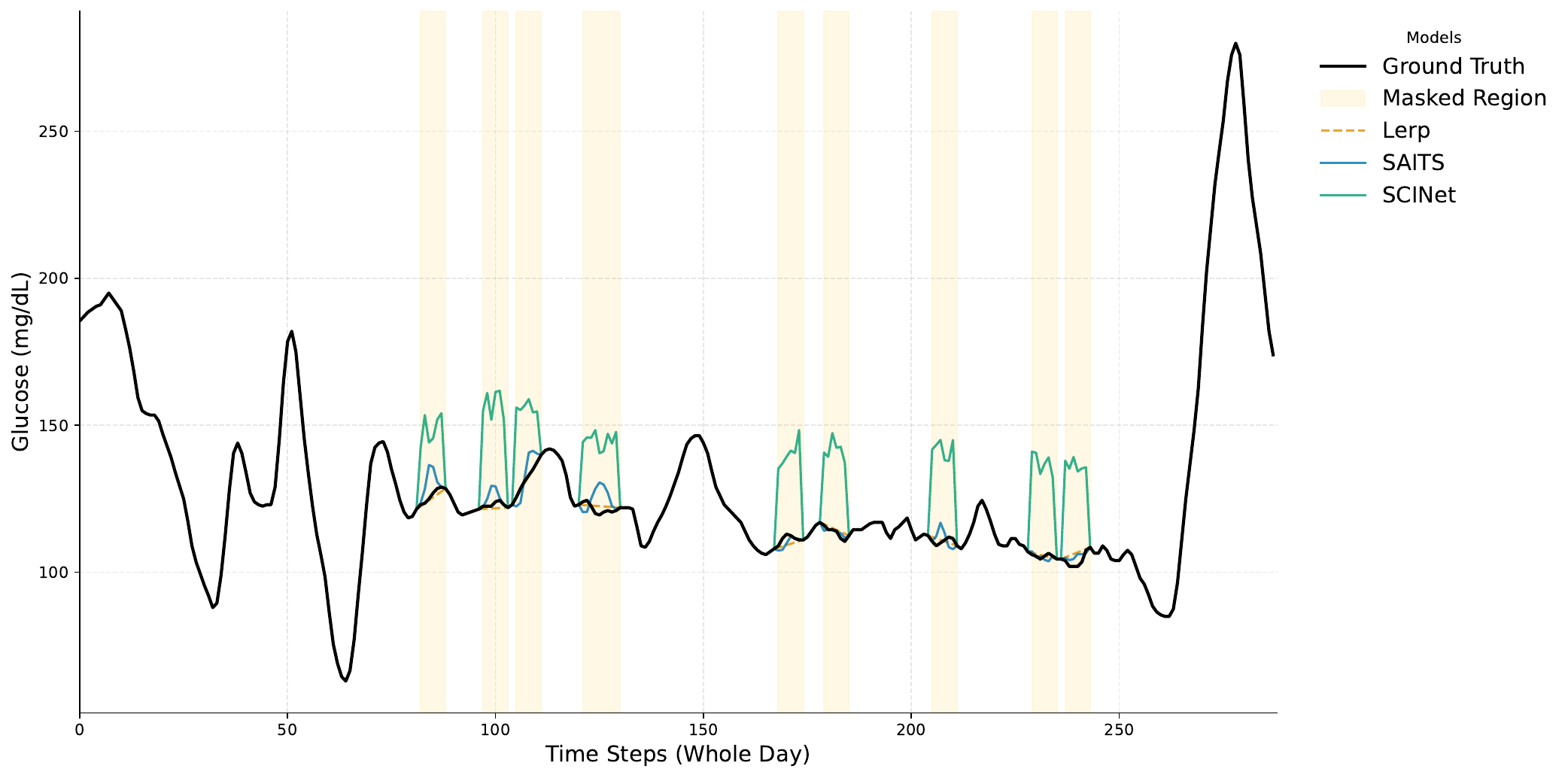}
    \end{subfigure}\hfill
    \begin{subfigure}[b]{0.32\linewidth}
        \includegraphics[width=\linewidth]{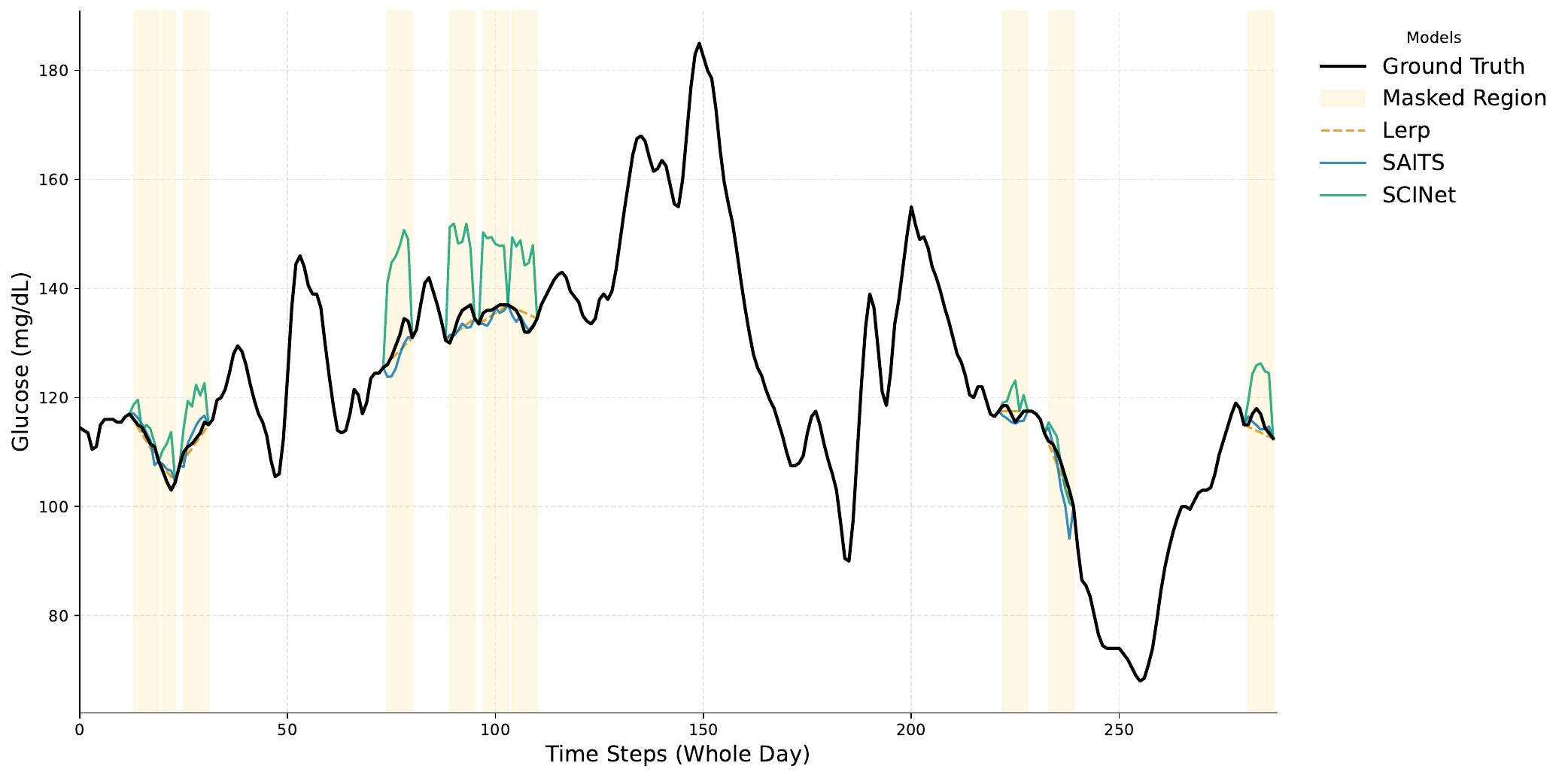}
    \end{subfigure}\hfill
    \begin{subfigure}[b]{0.32\linewidth}
        \includegraphics[width=\linewidth]{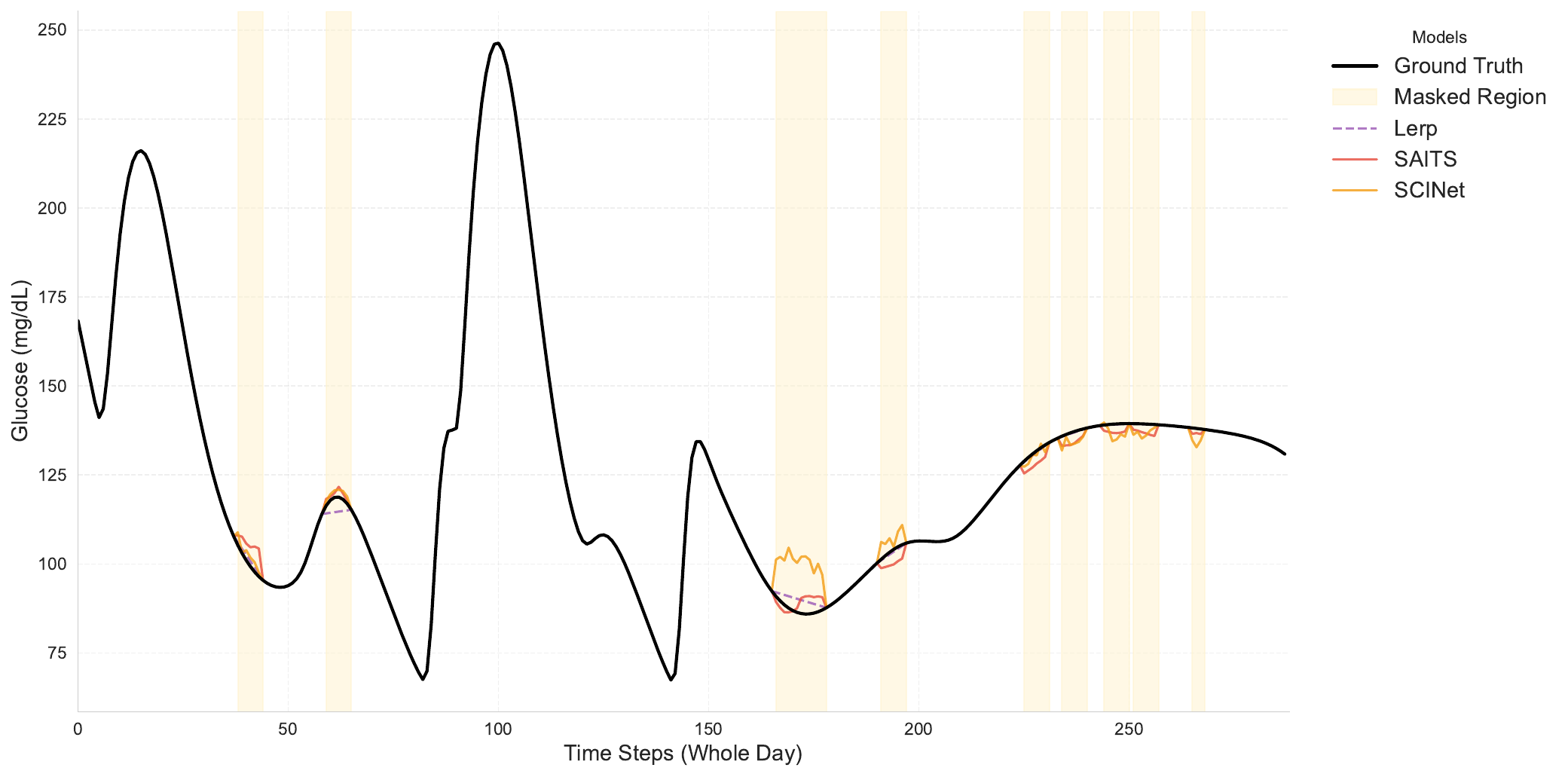}
    \end{subfigure}
    
    \smallskip
    
    \begin{subfigure}[b]{0.32\linewidth}
        \includegraphics[width=\linewidth]{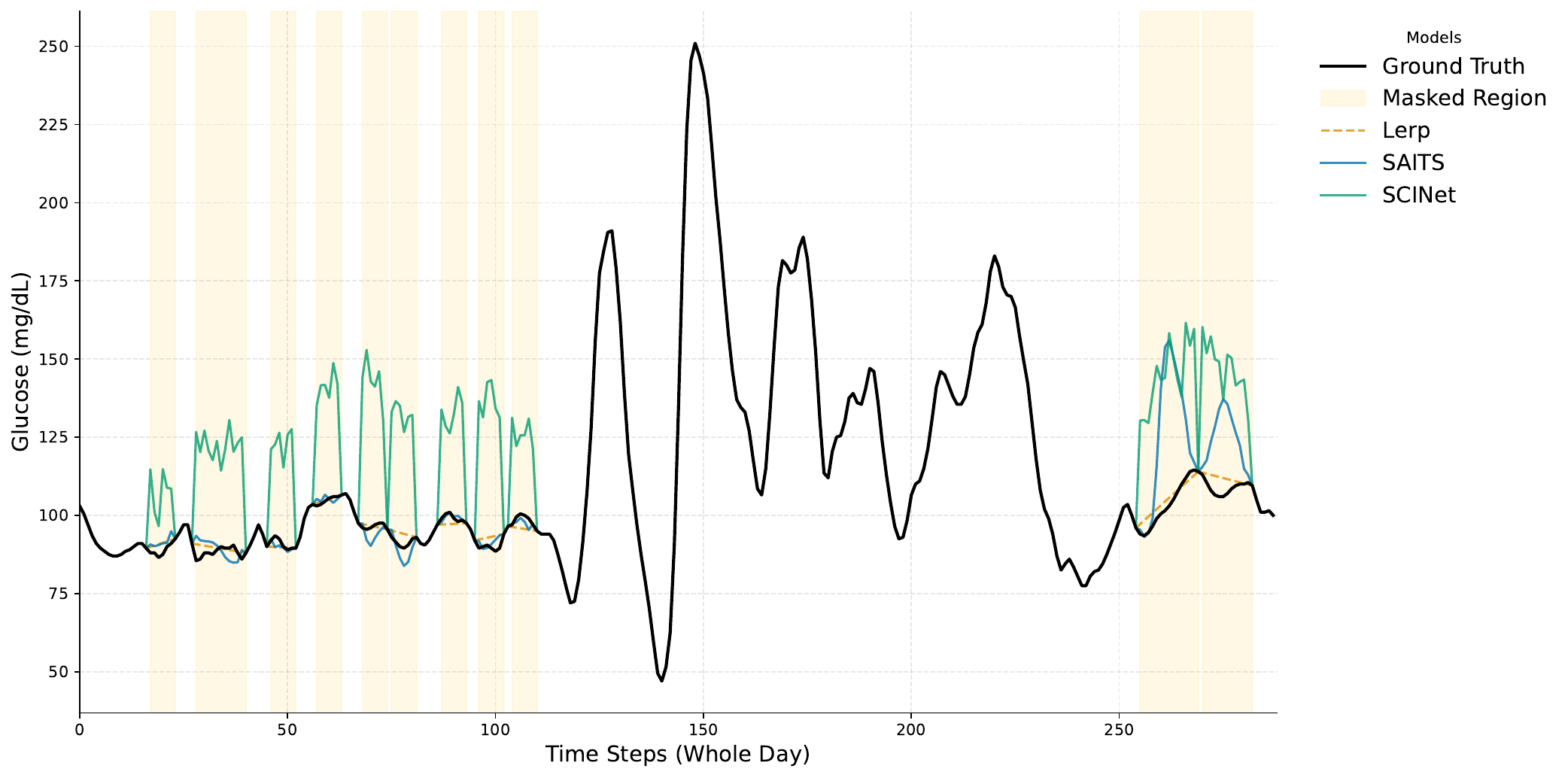}
    \end{subfigure}\hfill
    \begin{subfigure}[b]{0.32\linewidth}
        \includegraphics[width=\linewidth]{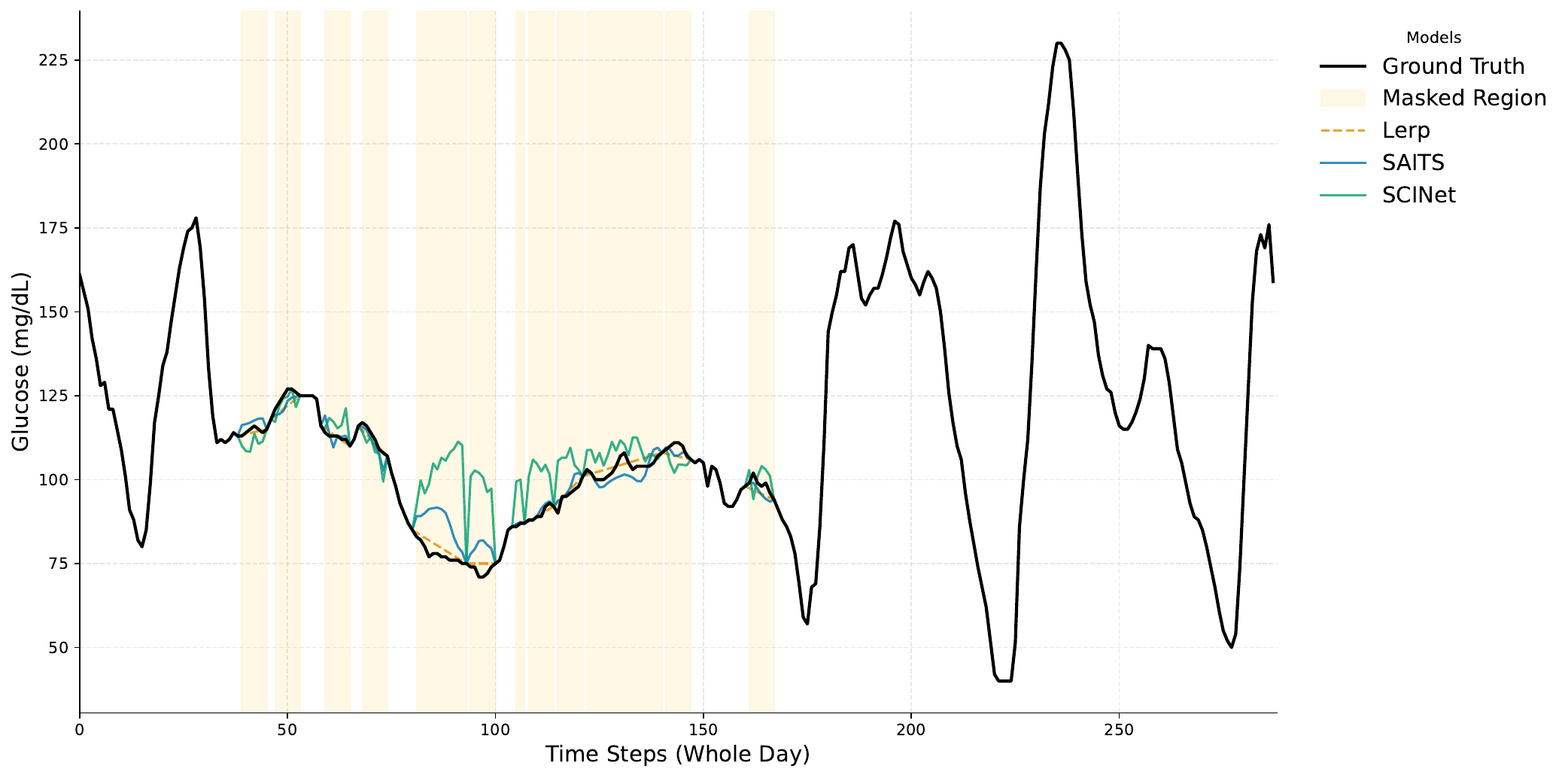}
    \end{subfigure}\hfill
    \begin{subfigure}[b]{0.32\linewidth}
        \includegraphics[width=\linewidth]{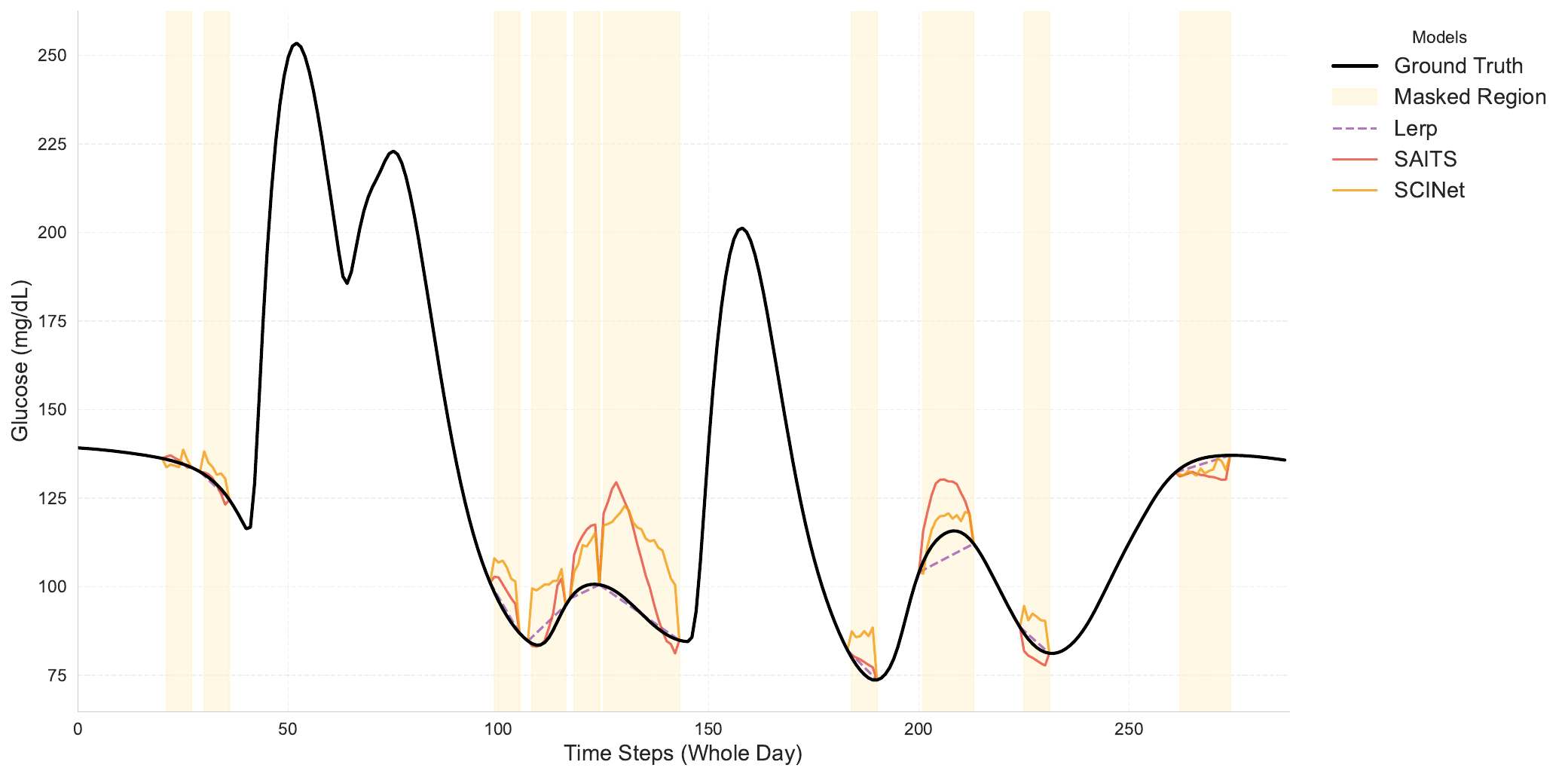}
    \end{subfigure}
    
\caption{Imputation examples for Scenario A. Rows correspond to masking ratios of 0.1, 0.2, and 0.3, while columns represent datasets.} 
\label{fig:scenario_A}
\end{figure*}

\begin{figure*}[htbp]
    \centering
    \begin{subfigure}{0.32\linewidth}\centering \textbf{Raw Pedap}\end{subfigure}%
    \hfill
    \begin{subfigure}{0.32\linewidth}\centering \textbf{Processed Pedap}\end{subfigure}%
    \hfill
    \begin{subfigure}{0.32\linewidth}\centering \textbf{Simulate}\end{subfigure}
    \smallskip

    \begin{subfigure}[b]{0.32\linewidth}
        \includegraphics[width=\linewidth]{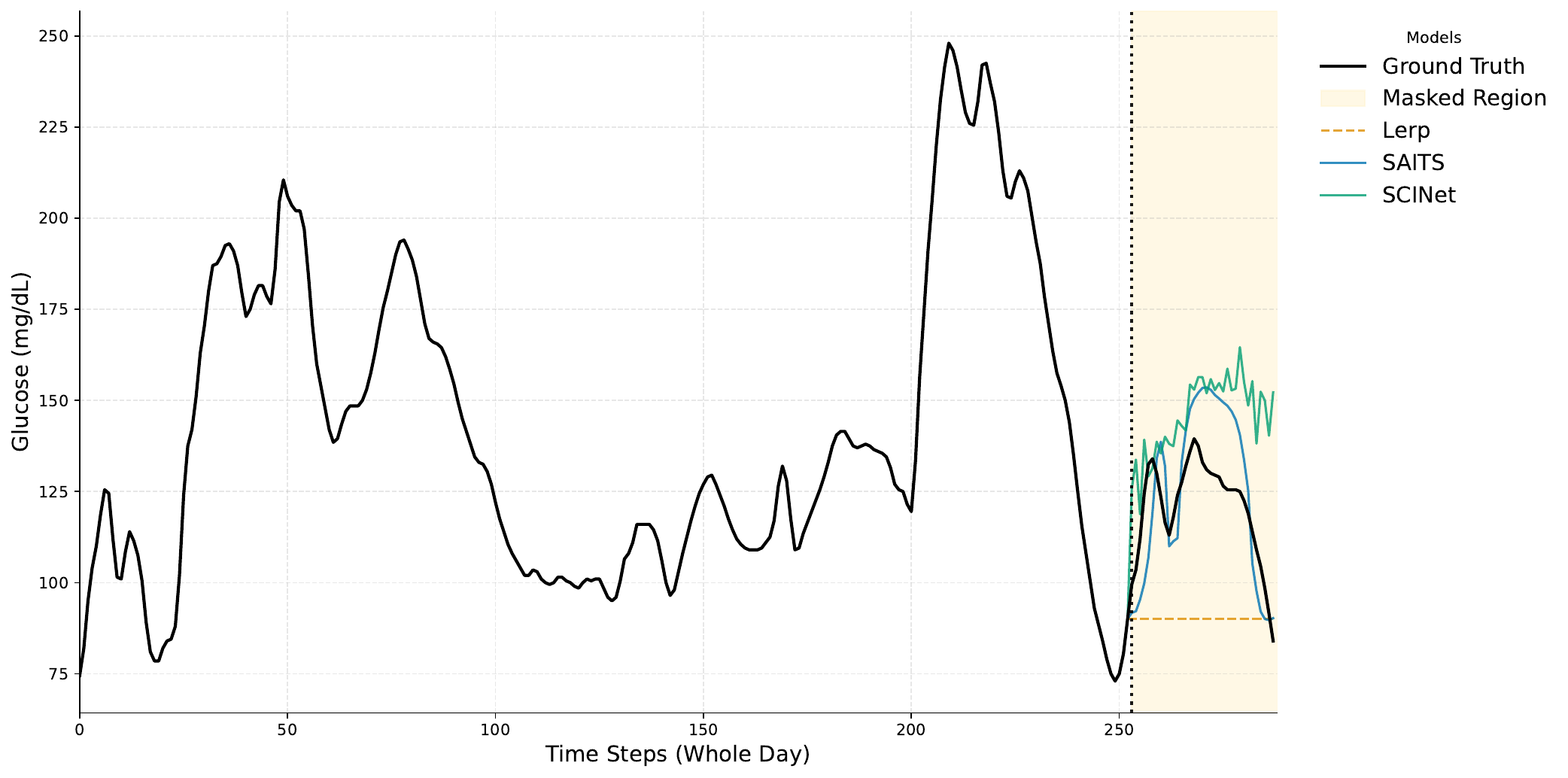}
    \end{subfigure}\hfill
    \begin{subfigure}[b]{0.32\linewidth}
        \includegraphics[width=\linewidth]{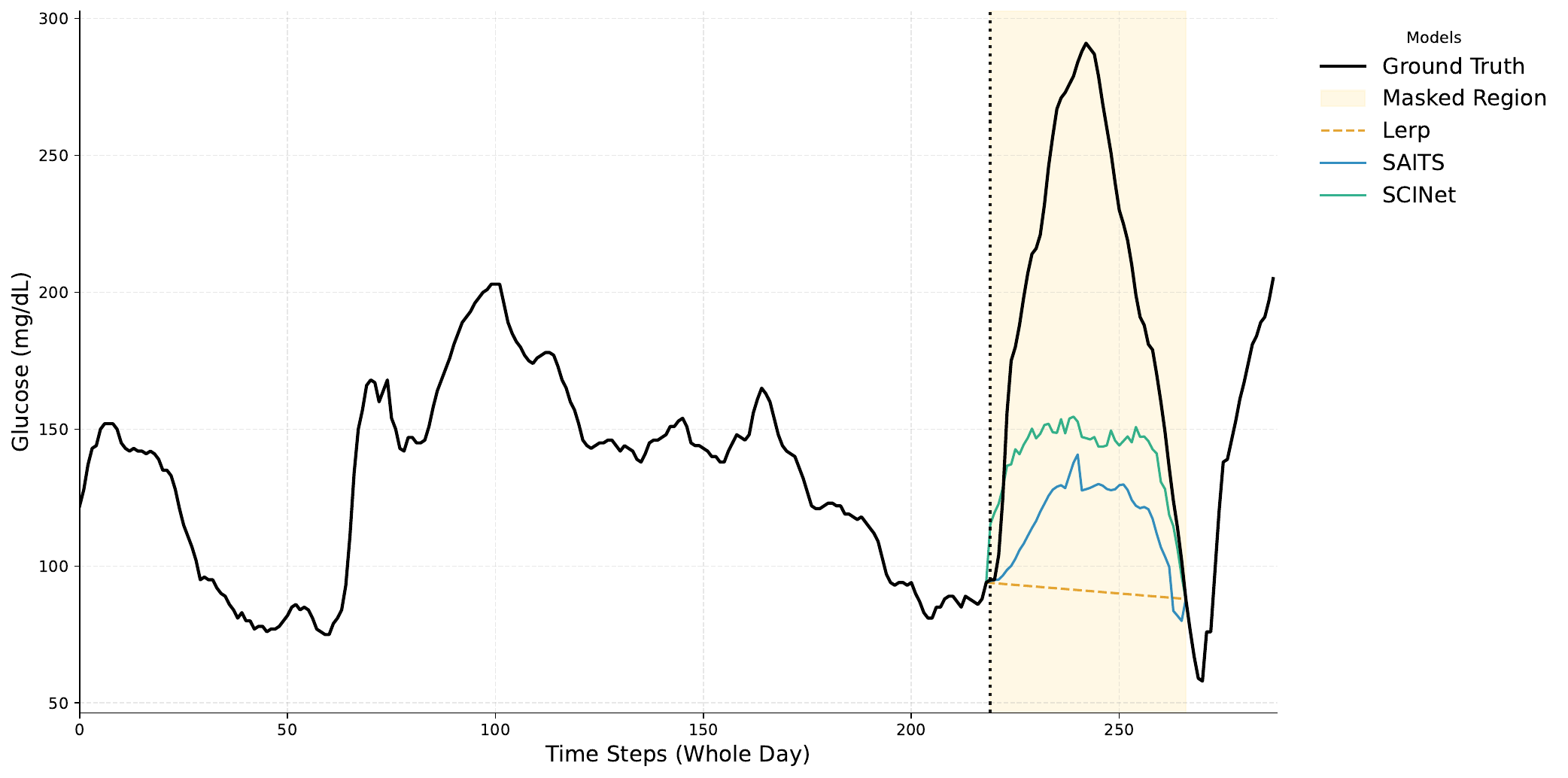}
    \end{subfigure}\hfill
    \begin{subfigure}[b]{0.32\linewidth}
        \includegraphics[width=\linewidth]{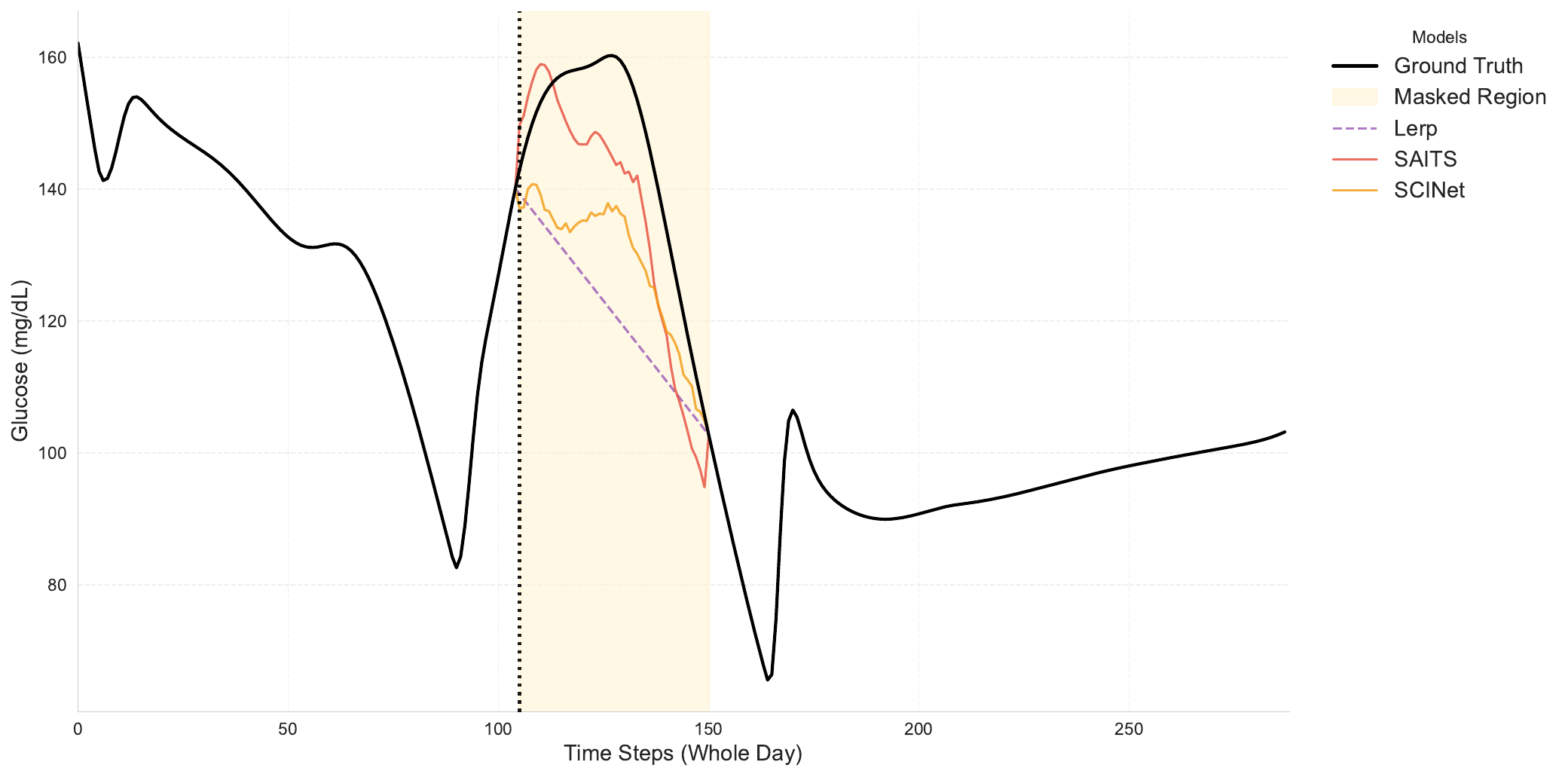}
    \end{subfigure}

    \smallskip

    \begin{subfigure}[b]{0.32\linewidth}
        \includegraphics[width=\linewidth]{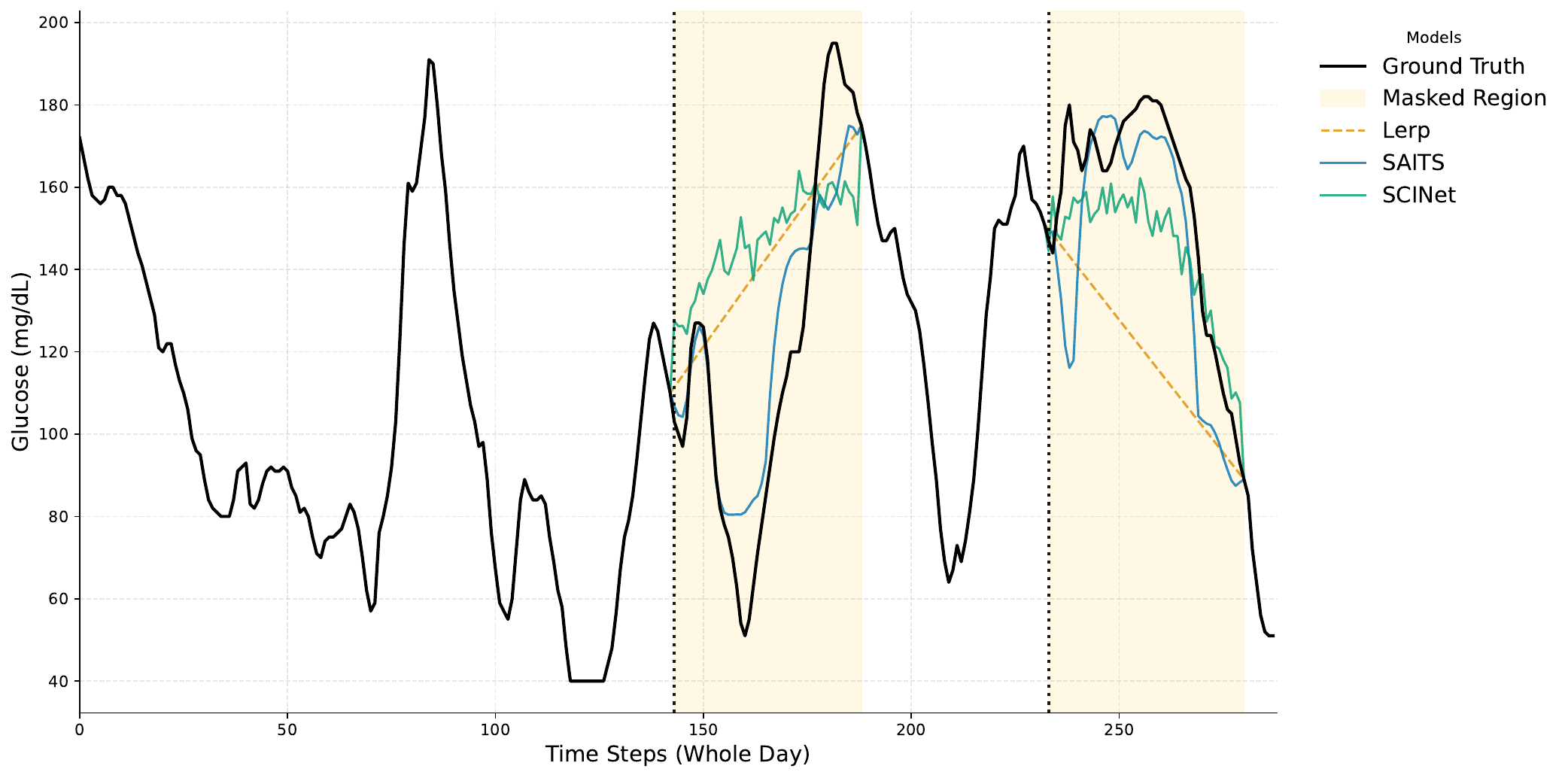}
    \end{subfigure}\hfill
    \begin{subfigure}[b]{0.32\linewidth}
        \includegraphics[width=\linewidth]{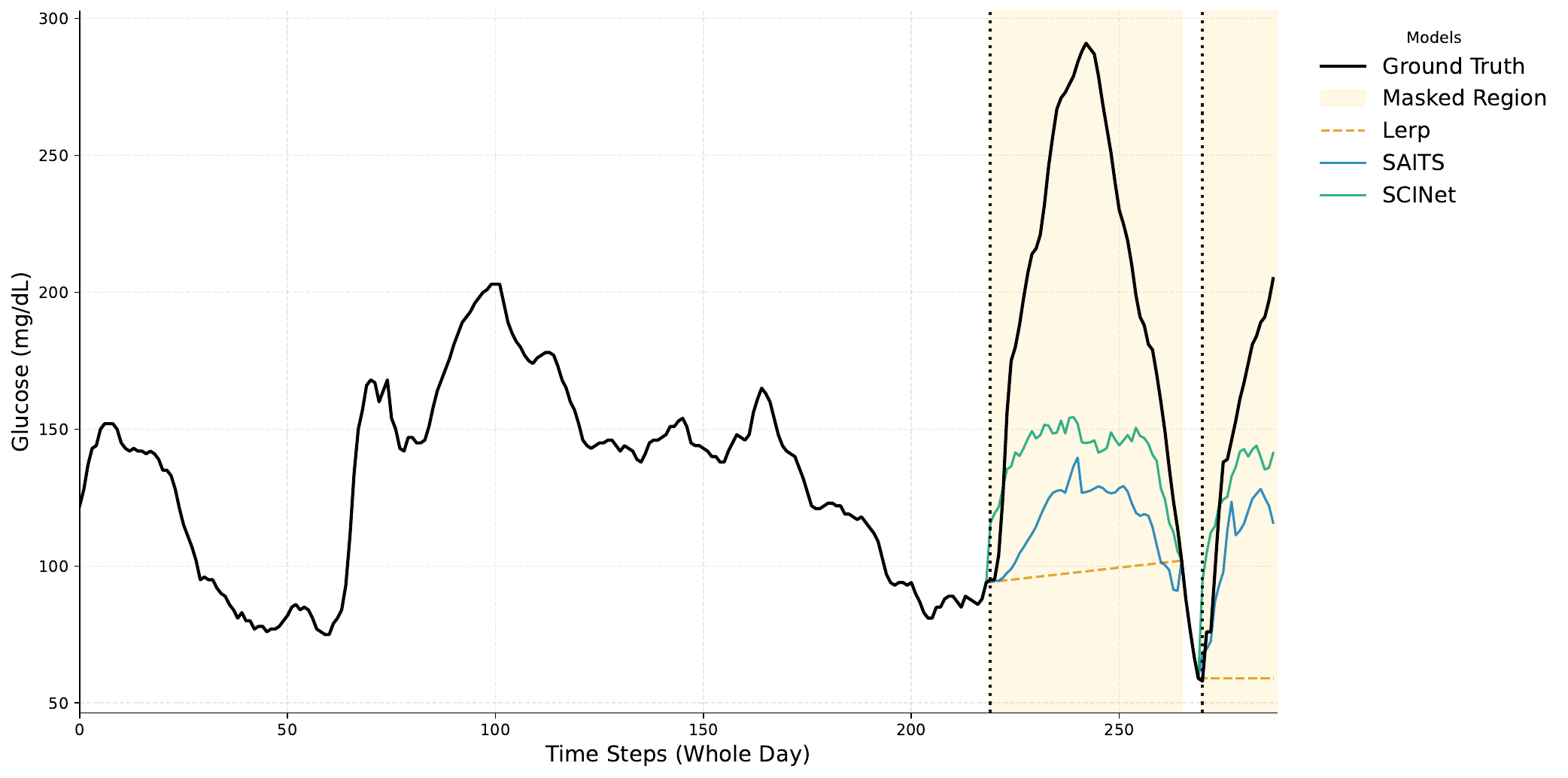}
    \end{subfigure}\hfill
    \begin{subfigure}[b]{0.32\linewidth}
        \includegraphics[width=\linewidth]{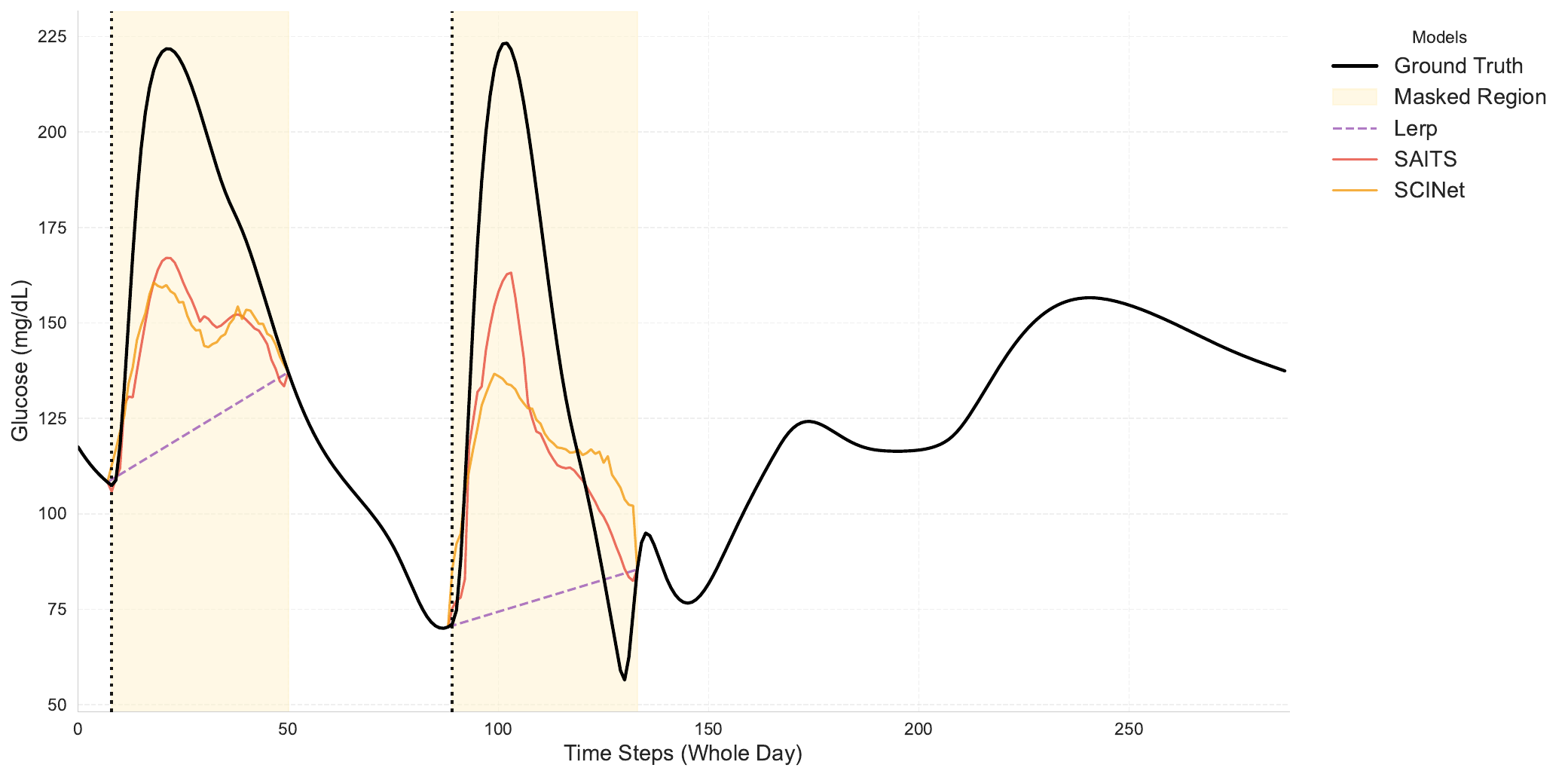}
    \end{subfigure}

    \smallskip

    \begin{subfigure}[b]{0.32\linewidth}
        \includegraphics[width=\linewidth]{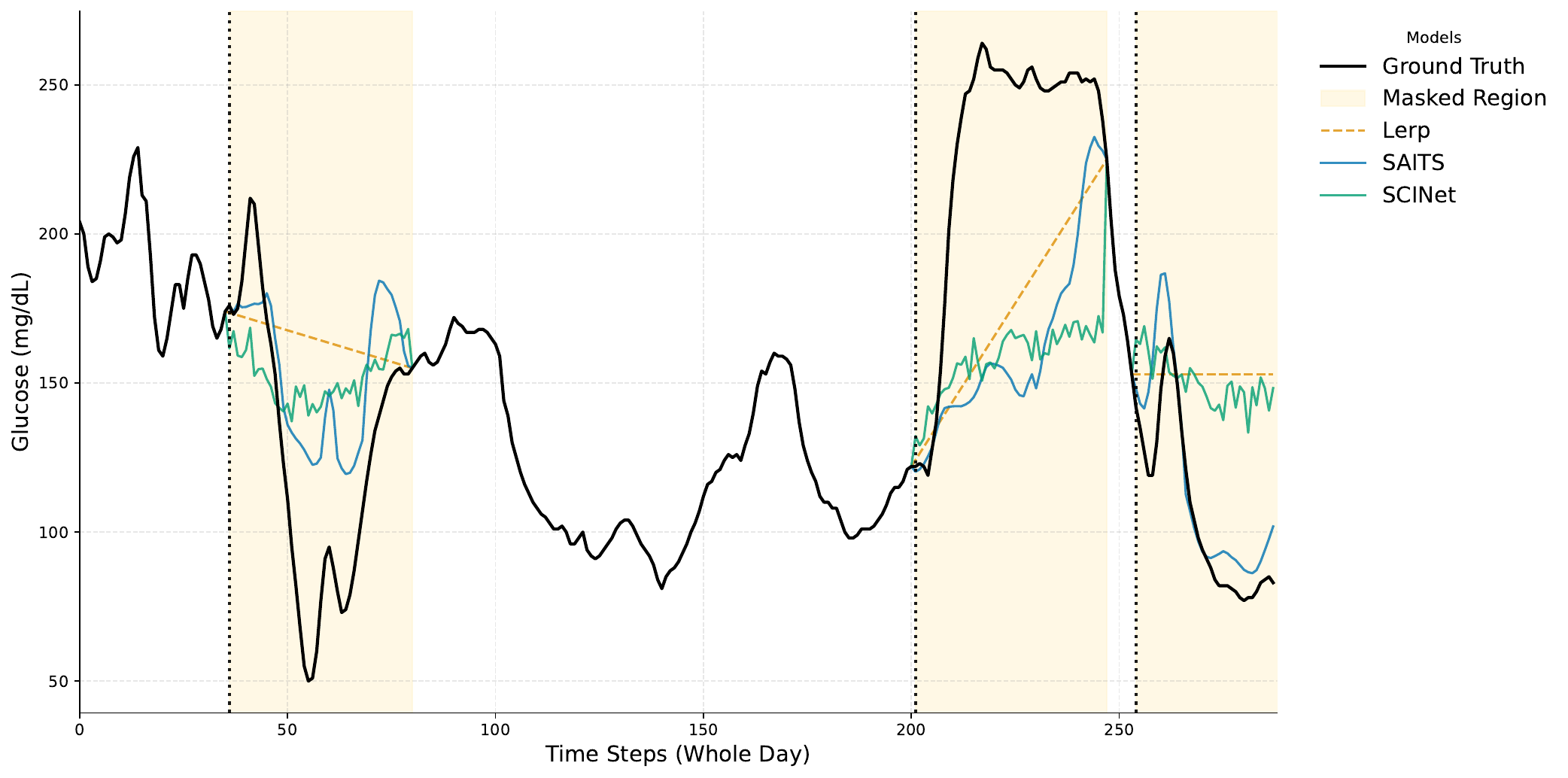}
    \end{subfigure}\hfill
    \begin{subfigure}[b]{0.32\linewidth}
        \includegraphics[width=\linewidth]{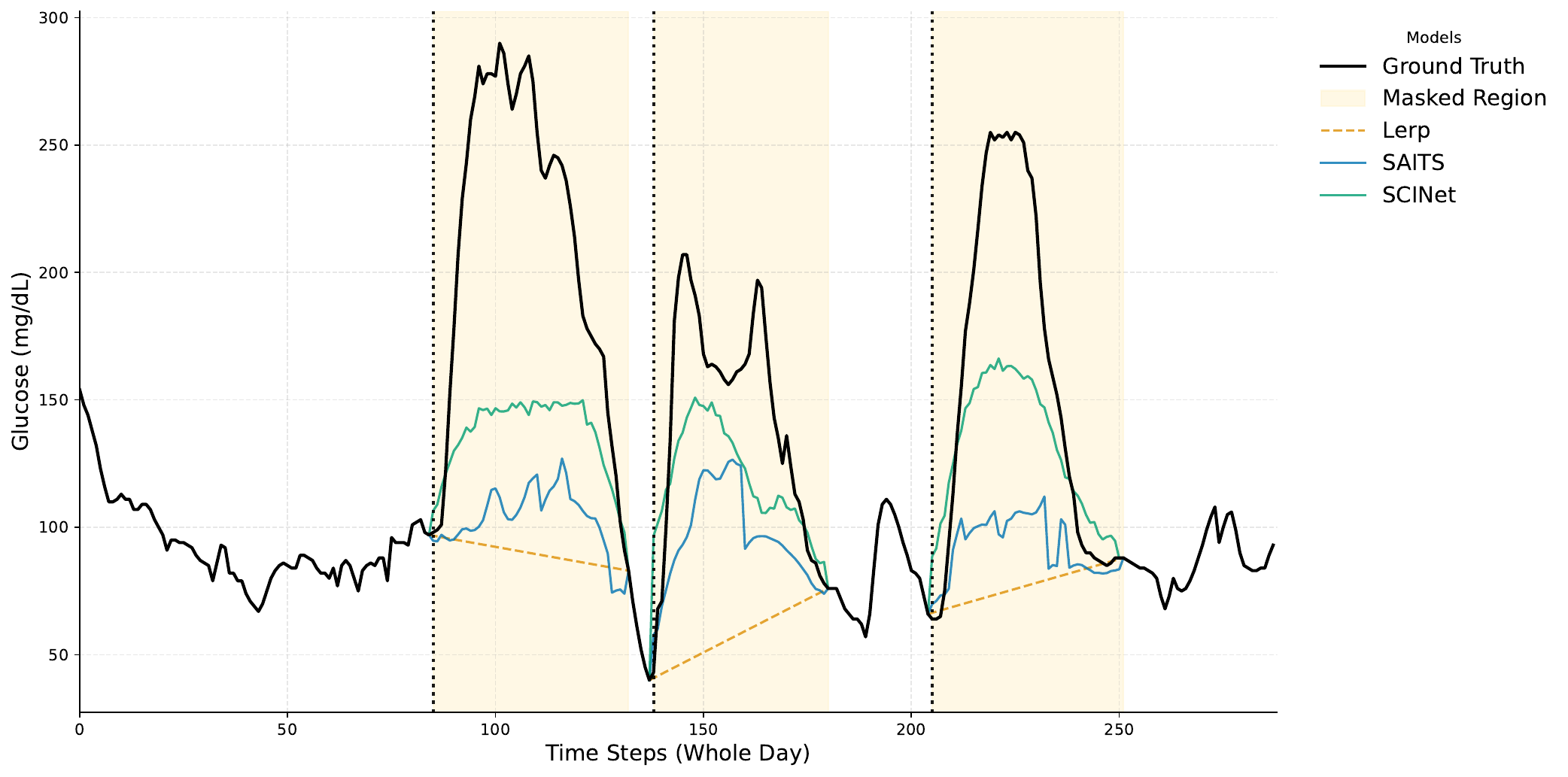}
    \end{subfigure}\hfill
    \begin{subfigure}[b]{0.32\linewidth}
        \includegraphics[width=\linewidth]{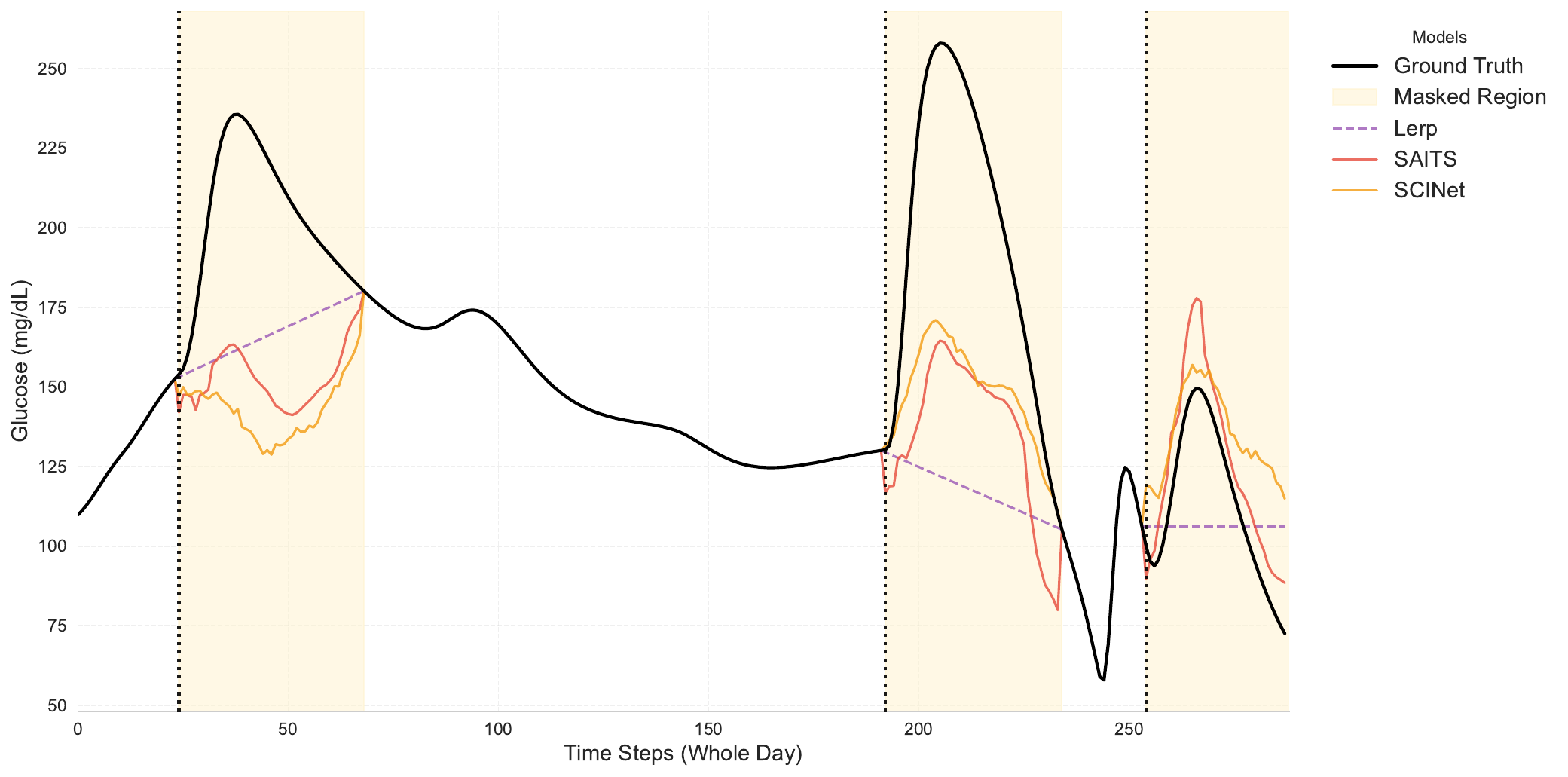}
    \end{subfigure}

    \caption{Scenario B imputation examples. Rows indicate 1, 2, and 3 masked peaks; columns correspond to datasets.}
    \label{fig:scenario_B}
\end{figure*}

\begin{figure*}[htbp]
    \centering
    \begin{subfigure}[b]{0.32\linewidth}
        \centering
        \includegraphics[width=\linewidth]{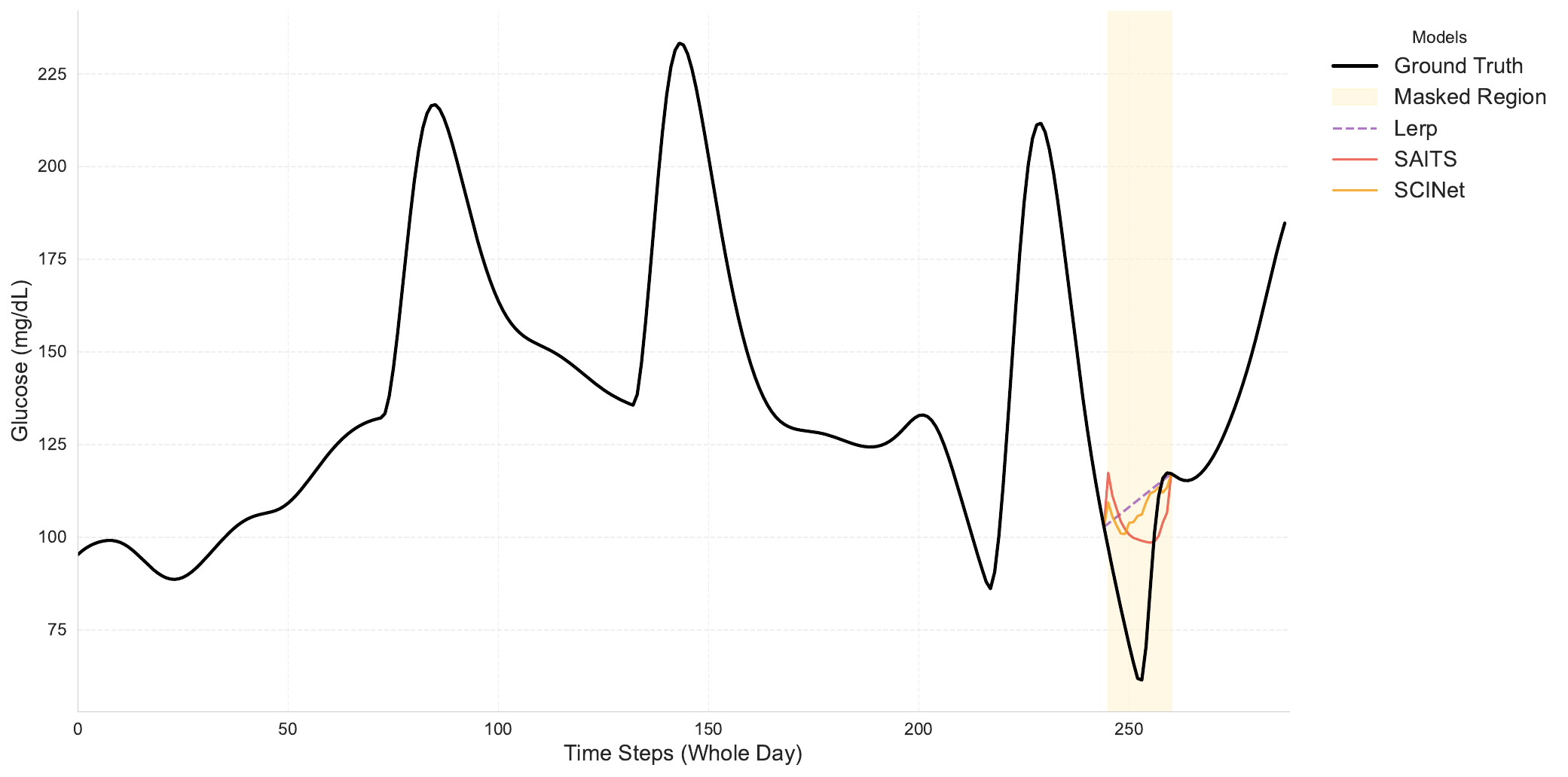}
    \end{subfigure}\hfill
    \begin{subfigure}[b]{0.32\linewidth}
        \centering
        \includegraphics[width=\linewidth]{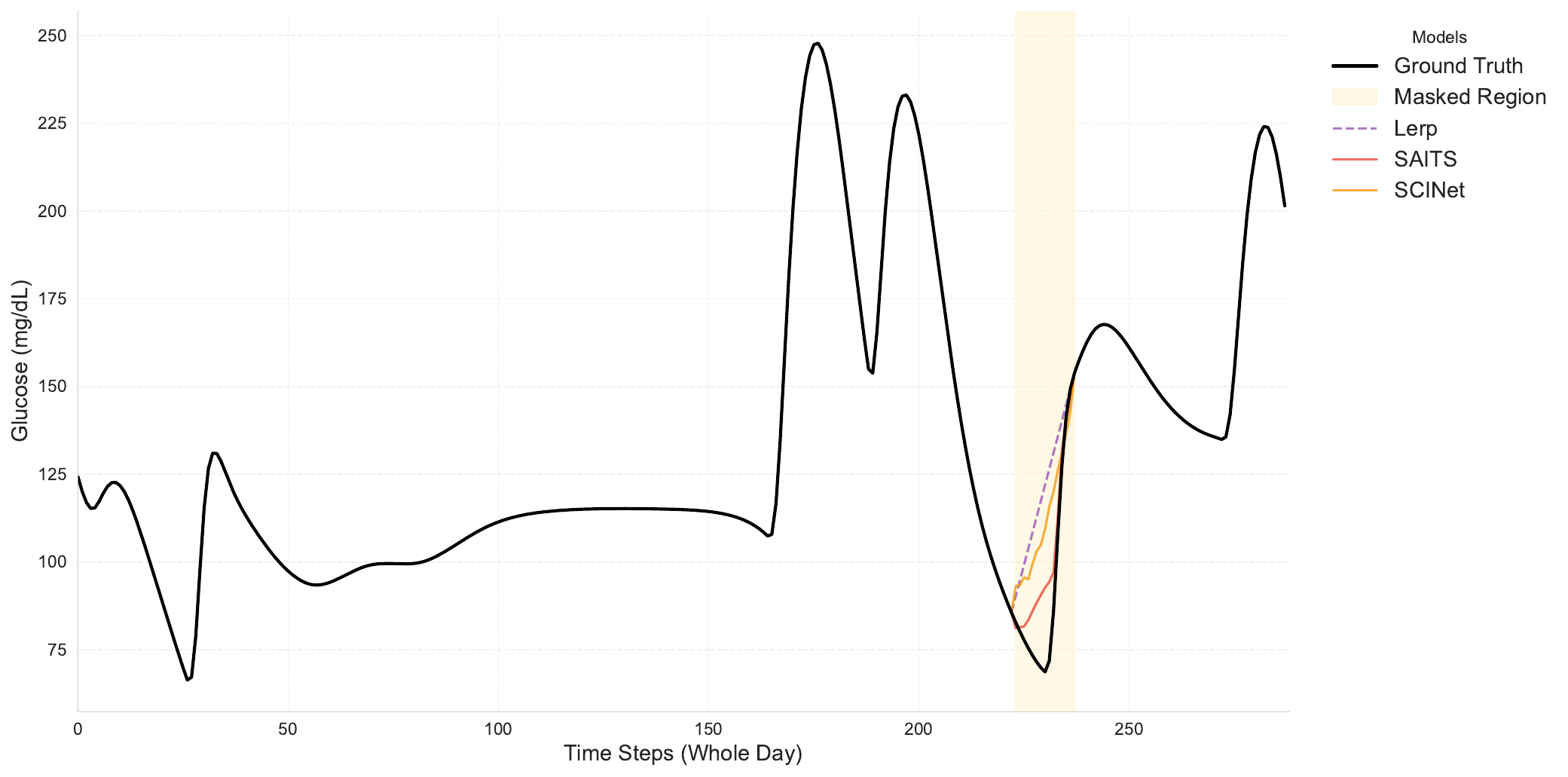}
    \end{subfigure}\hfill
    \begin{subfigure}[b]{0.32\linewidth}
        \centering
        \includegraphics[width=\linewidth]{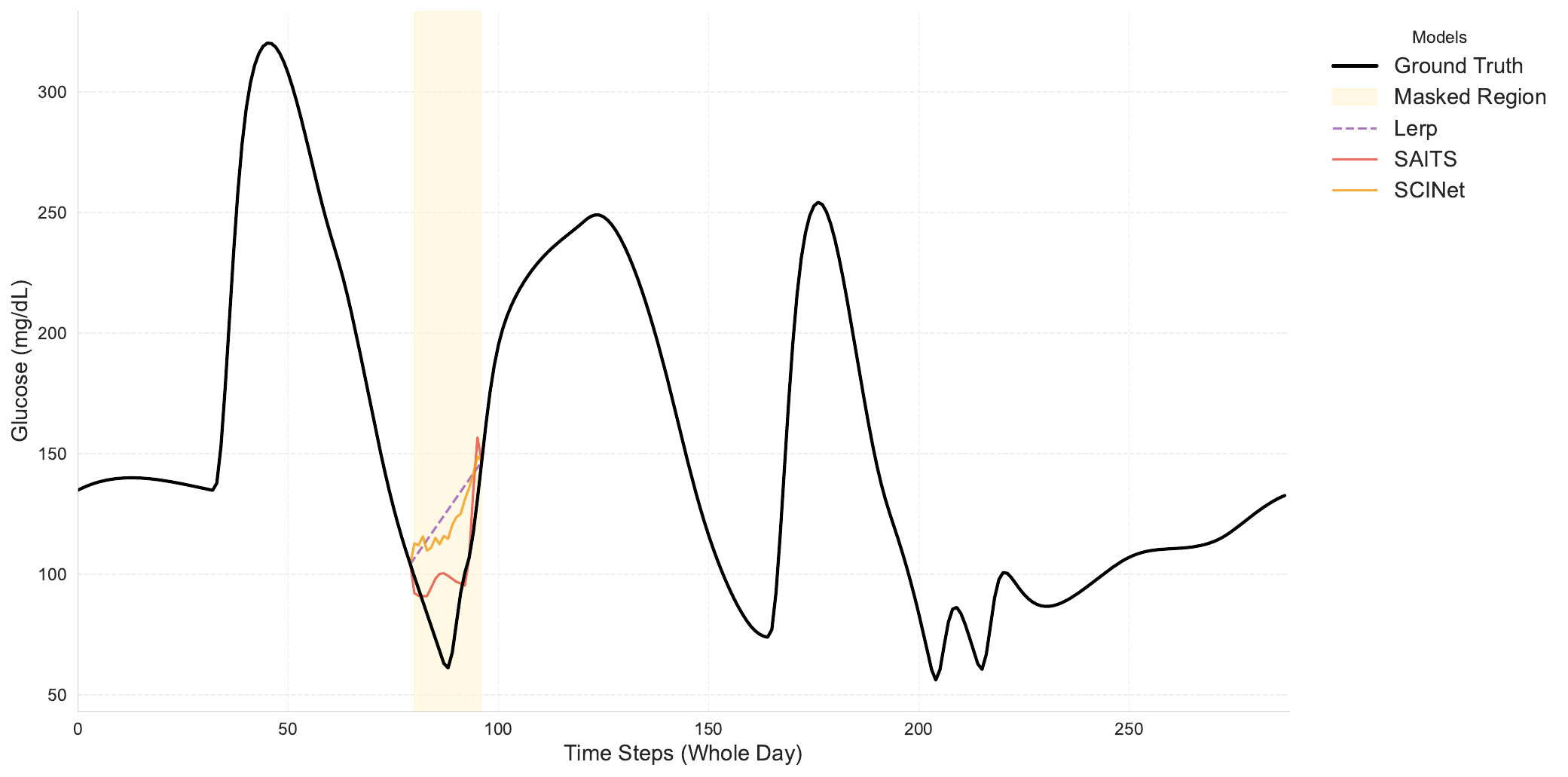}
    \end{subfigure}

    \caption{Scenario C imputation examples on the TCR simulation dataset.}
    \label{fig:scenario_C}
\end{figure*}

Figures~\ref{fig:scenario_A}, \ref{fig:scenario_B}, and \ref{fig:scenario_C} illustrate representative imputation samples across datasets for Scenarios A, B, and C, respectively.

\end{document}